\documentclass[lettersize,journal]{IEEEtran}
\usepackage{amsmath,amsfonts}
\usepackage{algorithmic}
\usepackage{array}
\usepackage[caption=false,font=normalsize,labelfont=sf,textfont=sf]{subfig}
\usepackage{textcomp}
\usepackage{stfloats}
\usepackage{cite}

\usepackage{url}
\usepackage{verbatim}
\usepackage{graphicx}
\usepackage{balance}
\usepackage{pgfplots}
\usepackage{pgfplotstable}
\usepackage{filecontents}
\usepackage{enumitem}
\usepackage{hyperref}
\usepackage{float}
\usepackage{makecell}
\usepackage{multirow}
\usepackage{graphicx}
\usepackage{tabularx}

\begin{document}
\title{SPPSFormer: High-quality Superpoint-based Transformer for Roof Plane Instance Segmentation from Point Clouds}

\author{Cheng Zeng, Xiatian Qi, Chi Chen, Kai Sun, Wangle Zhang, Yuxuan Liu, Yan Meng, Bisheng Yang
\thanks{This work was supported by the Major Program (JD) of Hubei Province under Grant No. 2023BAA018, the Key R \& D projects in Hubei Province under Grant No. 2021BAA188, the Knowledge Innovation Program of Wuhan-Shuguang Project under Grant No. 2023020201020414, and the Open Research Fund Program of LIESMARS under Grant No. 22S04. 
(Cheng Zeng, Xiatian Qi and Chi Chen contributed equally to this work.)
(Corresponding authors: Yan Meng and Bisheng Yang)}
\thanks{Cheng Zeng, Xiatian Qi, Kai Sun, and Yan Meng are with the School of Artificial Intelligence, Hubei University, Wuhan 430062, China, and also with the Key Laboratory of Intelligent Sensing System and Security (Hubei University), Ministry of Education, Wuhan 430062, China (email: zc@hubu.edu.cn; qixiatian@stu.hubu.edu.cn; sunkai@stu.hubu.edu.cn; mengyan@hubu.edu.cn).}
        
\thanks{Chi Chen and Bisheng Yang are with the State Key Laboratory of Information Engineering in Surveying, Mapping and Remote Sensing, Wuhan University, Wuhan 430079, China; the Engineering Research Center for Spatio-Temporal Data Smart Acquisition and Application, Ministry of Education of China, Wuhan 430079, China; and the Institute of Geospatial Intelligence, Wuhan University, Wuhan 430079, China (email: chichen@whu.edu.cn; bshyang@whu.edu.cn).}
        
\thanks{Wangle Zhang is with the Hubei Water Resources Research Institute, Wuhan 430205, China.}
\thanks{Yuxuan Liu is with the Institute of Photogrammetry and Remote Sensing, Chinese Academy of Surveying and Mapping (CASM), Beijing 100830, China.}
}

\markboth{Journal of \LaTeX\ Class Files,~Vol.~18, No.~9, September~2020}%
{How to Use the IEEEtran \LaTeX \ Templates}

\maketitle

\begin{abstract}
Transformers, especially superpoint Transformers, have demonstrated significant potential in point cloud segmentation tasks, owing to their powerful feature extraction and learning capabilities. 
However, Transformers have been seldom employed in point cloud roof plane instance segmentation, which is the focus of this study, and existing superpoint Transformers suffer from limited performance due to the use of low-quality superpoints.
To address this challenge, we establish two criteria that high-quality superpoints for Transformers should satisfy and introduce a corresponding two-stage superpoint generation process. 
The superpoints generated by our method not only have accurate boundaries, but also exhibit consistent geometric sizes and shapes, both of which greatly benefit the feature learning of superpoint Transformers. 
To compensate for the limitations of deep learning features when the training set size is limited, we incorporate multidimensional handcrafted features into the model. 
Additionally, we design a decoder that combines a Kolmogorov–Arnold Network with a Transformer module to improve instance prediction and mask extraction. 
Finally, our network’s predictions are refined using traditional algorithm-based postprocessing.
For evaluation, we annotated a real-world dataset and corrected annotation errors in the existing RoofN3D dataset. 
Experimental results show that our method achieves state-of-the-art performance on our dataset, as well as both the original and reannotated RoofN3D datasets. Moreover, our model is not sensitive to plane boundary annotations during training, significantly reducing the annotation burden. 
Through comprehensive experiments, we also identified key factors influencing roof plane segmentation performance: in addition to roof types, variations in point cloud density, density uniformity, and 3D point precision have a considerable impact. 
These findings underscore the importance of incorporating data augmentation strategies that account for point cloud quality to enhance model robustness under diverse and challenging conditions. 
We will release our code, trained models, and datasets.

\end{abstract}

%


\section{Introduction}
\label{sec:introduction}
\IEEEPARstart{S}{egmenting} roof plane instances from 3D point clouds is critical for 3D building reconstruction and rooftop photovoltaic installation planning.
In fact, 3D point cloud roof plane instance segmentation serves as a key step in many 3D building reconstruction algorithms \cite{Li_ISPRS_2022, Li_RS_2020, Yan_ISPRS_2016, Nan_ICCV_2017}. 
Consequently, accurate roof plane instance segmentation directly contributes to the quality of the resulting models. 
In addition, this segmentation technology plays a vital role in planning rooftop photovoltaic installations \cite{Zhang_PR_2022}, as accurate instance segmentation of roof planes helps to reduce the need for costly and time-consuming on-site surveys.

Traditional plane instance segmentation methods---such as region growing algorithms \cite{Poux_AIC_2022,Li_RS_2020}, feature clustering-based algorithms   \cite{Sampath_TGRS_2010,Chen_PR_2023}, and model fitting-based algorithms \cite{Zhang_JGGS_2019,Xu_AIC_2020,Fischler_ACM_1981,Li_RS_2017}---typically rely on prior knowledge of point cloud characteristics and require manual parameter tuning. 
These dependencies limit their generalization ability and automation, particularly in complex and diverse real-world building scenarios. 
In practice, the shortcomings of traditional algorithms become especially evident when facing intricate roof structures.

In contrast, deep learning methods, especially Transformer-based methods, have demonstrated strong capabilities in point cloud segmentation, owing to their powerful feature extraction and representation learning abilities \cite{Guo_PAMI_2021,Sun_AAAI_2023,Robert_ICCV_2023,Wu_CVPR_2024,Zhu_IROS_2023,Dai_TGRS_2024,Zhang_IJAEOG_2024,Huang_TGRS_2024}.
Nevertheless, current deep learning models specifically designed for roof plane instance segmentation in 3D point clouds primarily adopt PointNet++ \cite{Qi_ANIPS_2017} as the backbone network \cite{Li_ISPRS_2024,Zhang_PR_2022}, which struggles to fully exploit global contextual information.
The superpoint Transformer-based 3D point cloud instance segmentation network, SPFormer \cite{Sun_AAAI_2023}, was retrained on the RoofNTNU dataset in \cite{Kong_TGRS_2024} and achieved promising results.
However, the RoofNTNU dataset contains only 930 roof samples, and its limited scale makes it insufficient to fully validate the effectiveness of SPFormer in the domain of point cloud roof plane segmentation.

In fact, existing superpoint Transformers, including SPFormer \cite{Sun_AAAI_2023}, generally rely on simplistic geometric rules or clustering strategies to construct superpoints.
These methods often yield superpoints that include points from multiple instances when handling boundary regions between adjacent instances, resulting in blurred segmentation boundaries and inaccurate plane segmentation. 
In addition, superpoints generated by these simplistic methods often vary significantly in size and shape. 
Such irregularities introduce complex data distributions, which hinder the Transformer from learning stable and discriminative features. 
This challenge makes model training more difficult and demanding larger datasets for adequate performance. 
Unfortunately, annotating large-scale 3D point cloud datasets is both time-consuming and expensive. 
Therefore, generating high-quality superpoints has become a critical challenge that must be addressed to fully leverage the capabilities of superpoint Transformers for segmentation tasks.

In this paper, we develop a superpoint Transformer-based method for segmenting plane instances in point clouds. 
To address the issue of suboptimal superpoints in existing superpoint Transformer models, we propose two essential criteria that high-quality superpoints should meet:

\vspace{0.2cm}
\begin{center}
    \begin{minipage}{0.45\textwidth} 
        \textbf{They must have accurate boundaries, and different superpoints should be similar in both size and shape. }
    \end{minipage}
\end{center}
\vspace{0.2cm}

Based on the proposed criteria, we introduce a two-stage superpoint generation process that significantly enhances the performance of the superpoint Transformer in segmenting plane instances from 3D point clouds. 
To address the limited feature extraction capability of deep learning models---especially in scenarios with small training sets---we incorporate multidimensional handcrafted features and spatial position features at the input stage of our network. 
To further strengthen the decoder, we adopt the Kolmogorov–Arnold Network (KAN) \cite{Liu_ICLR_2024} for mask feature extraction and combine it with a Transformer module. 
This hybrid decoder architecture enables direct prediction of instance classes, confidence scores, and mask outputs.

Finally, traditional algorithm-based plane completion and boundary refinement modules are introduced to refine the predictions of our network.
Although this conflicts with the end-to-end paradigm pursued by many researchers, we will demonstrate later in this paper that our architecture---consisting of deep learning-based prediction followed by traditional postprocessing---offers several important advantages.

\textbf{Plane completion.}
Existing datasets for 3D roof plane instance segmentation, such as the RoofN3D dataset \cite{Wichmann_RS_2019}, may contain incomplete plane annotations, which can cause trained models to miss valid roof planes during inference. 
To mitigate this, we propose a self-supervised plane completion strategy that first infers the parameters of missing planes based on already segmented planes and uses this information to guide a region growing algorithm to recover missing plane instances from points that are initially labeled as ``nonplane".

\textbf{Boundary refinement.}
While deep learning models often produce inaccurate segmentation boundaries, traditional region growing methods are also prone to boundary inaccuracies due to their sequential nature---early-growing planes may absorb points from neighboring planes. 
To address this, we apply a plane boundary refinement algorithm to further enhance segmentation accuracy.

For evaluation, we constructed a 3D point cloud roof plane instance segmentation dataset by annotating 10,539 buildings based on wireframe annotations from the Building3D dataset \cite{Wang_ICCV_2023} and corrected annotation errors in the RoofN3D dataset \cite{Wichmann_RS_2019} for samples containing at least 700 points. 
By integrating our network design with traditional algorithm-based postprocessing, our method achieves state-of-the-art (SOTA) performance on the Building3D dataset, as well as on both the original and reannotated versions of RoofN3D.

We also disrupted the boundary annotations in the Building3D and RoofN3D training sets---while preserving accurate segmentation of plane main bodies---and retrained both our model and the current SOTA deep learning–based method DeepRoofPlane \cite{Li_ISPRS_2024} using these boundary-degraded training datasets.
Remarkably, our method showed almost no performance drop on the test sets, whereas the performance of DeepRoofPlane declined significantly. 
In general, neural network–based segmentation models rely heavily on fine annotations of the training set. 
However, accurately annotating the boundaries between different categories or instances is particularly challenging. 
Our model, nonetheless, demonstrates strong robustness to inaccurate boundary annotations, thanks to its hybrid structure: a deep neural network predicts plane instances, followed by traditional postprocessing algorithms. 
This design allows for accurate segmentation even when only the main bodies of planes are accurately annotated in the training set, significantly reducing the complexity and cost of creating point cloud datasets for plane instance segmentation.

Additionally, we applied controlled degradations to our annotated Building3D dataset by separately reducing point cloud density, introducing greater density variations, and decreasing point precision---resulting in three distinct degraded datasets.
We trained and evaluated models on these variants, and our algorithm consistently outperformed the existing SOTA method DeepRoofPlane \cite{Li_ISPRS_2024} across all conditions. 
More critically, we observed that the segmentation performance of both DeepRoofPlane \cite{Li_ISPRS_2024} and our model dropped significantly under these degradations. 
This highlights that data quality factors---especially point cloud density, density variation, and point precision---have a profound impact on the effectiveness of point cloud roof plane instance segmentation. 
These findings underscore the necessity of incorporating data augmentation strategies that simulate these degradations during training to improve model robustness in real-world and unpredictable scenarios.

In summary, the key contributions of this paper are as follows:

\begin{itemize}[leftmargin=*, label=\scalebox{1.2}\textbullet, itemsep=0em]  
\item We establish criteria for generating superpoints that are well-suited for Transformer feature learning, and propose a corresponding superpoint generation procedure. 
Our experiments demonstrate that applying these criteria significantly enhances the performance of the superpoint Transformer.

\item By employing high-quality superpoints, incorporating handcrafted features into our superpoint Transformer, designing a decoder that integrates KAN and Transformer, and applying traditional algorithm-based postprocessing, we realize accurate point cloud roof plane instance segmentation.
Our model achieves new SOTA results on both the RoofN3D and Building3D datasets. 
Notably, our model relies primarily on the accurate annotation of the plane main bodies, while being less sensitive to the accuracy of boundary annotations. Therefore, our hybrid architecture---combining neural network–based prediction with traditional algorithm-based postprocessing---substantially reduces the annotation burden for point cloud plane instance segmentation training sets.

\item Through experiments, we identify the critical factors influencing point cloud plane segmentation performance: in addition to roof structure complexity, point cloud density, density variation, and point precision all significantly affect segmentation results. 
As such, future research should not only focus on the complexity of roof structures but also consider the impact of data quality on segmentation performance. 
For instance, incorporating data augmentation techniques targeting point cloud quality---including point cloud density, density variation, and point precision---during training can improve the model’s robustness when handling complex and unseen data.

\item We have annotated a real-world 3D point cloud roof plane instance segmentation dataset and corrected labeling errors in the existing RoofN3D dataset for samples containing at least 700 points.
We will publicly release both our newly annotated dataset and the corrected RoofN3D samples.

\end{itemize}

Additionally, we will open-source our code and trained models.

\section{Related work}
\label{sec:relatedwork}

In this section, we review the advancements in research areas closely related to our method, including techniques for segmenting plane instances from 3D point clouds, superpoints in superpoint Transformers, and the KAN along with its recent applications.

\subsection{Plane instance segmentation from 3D point clouds}
\label{sec::rgmethods}

Over the past decades, substantial progress has been made in 3D point cloud plane instance segmentation. 
Existing methods can be broadly grouped into three categories: traditional sequential algorithms, traditional global algorithms, and deep learning-based algorithms.

\subsubsection{Traditional sequential algorithms}
\label{sec::type1}
Sequential algorithms segment plane instances in 3D point clouds in a step-by-step manner, where the results depend on the order in which points are processed. Traditional sequential methods primarily include region growing, RANdom SAmple Consensus (RANSAC), and Hough Transform algorithms.

Region growing, initially proposed for image segmentation \cite{Adams_TPAM_1994}, was later adapted for 3D point cloud segmentation. 
The core idea is to begin with one or more seed points and, by comparing its (or their) similarity with neighboring pixels or other 3D units, merge similar elements into the same region \cite{Jagannathan_TPAMI_2007,Yu_PAMI_2008,Vo_ISPRS_2015,Wang_JGS_2021,Poux_AIC_2022}. 
In the context of plane instance segmentation from 3D point clouds, region growing typically operates on either individual 3D points or voxels. 
For example, Poux et al.~\cite{Poux_AIC_2022} and Yan et al.~\cite{Yan_ISPRS_2014} used individual 3D points as basic units, while Wang and Wang~\cite{Wang_JGS_2021} and Li et al.~\cite{Li_RS_2020} adopted voxels. 
Dong et al.~\cite{Dong_ISPRS_2018} proposed a hybrid representation: voxels are used in areas likely to contain planes, while individual points are used elsewhere. 
Using individual points provides higher precision but at a cost to computational efficiency. 
In contrast, using voxels improves efficiency---especially at lower resolution---but often compromises segmentation accuracy. 
The performance of region growing algorithms is heavily sensitive to the selection of seed points and growth criteria. 
Despite this, they are relatively efficient and are often used as a preprocessing step in multistage algorithms \cite{Yan_ISPRS_2014,Dong_ISPRS_2018,Li_RS_2020,Wang_STAEORS_2021}.

RANSAC and Hough Transform algorithms can segment plane instances from 3D point clouds through model fitting. 
In RANSAC-based methods \cite{Schnabel_CGF_2007,Fujiwara_ICIEA_2013,Xu_RS_2016,Liu_JGS_2021,Xu_GISWU_2023}, a small number of 3D points (more than two points are required) are sampled to fit a plane, and the number of inliers---points near the fitted plane---is counted. 
The plane with the most inliers is selected, and the inliers are removed from the point cloud. 
The process repeats on the remaining points to extract additional planes. 
The final segmentation consists of the planes with the largest inlier sets. 
In contrast, Hough Transform methods \cite{Leng_TPR_2016,Maltezos_ISPRS_2016,Vera_PRL_2018,Xu_AIC_2020,Xu_JMS_2023} operate by mapping points from 3D space into a discretized parameter space, where planes are detected through a voting mechanism to identify parameter combinations that best represent planar surfaces.

Both RANSAC and Hough Transform have been used in plane segmentation since the early days of 3D point cloud processing ~\cite{Wahl_CGA_2005,Vosselman_ISPRS_2004} and remain widely adopted in modern applications~\cite{Xu_AIC_2020,Liu_JGS_2021,Xu_GISWU_2023,Xu_JMS_2023}. 
Typically, RANSAC offers higher efficiency than Hough Transform, particularly when processing large-scale point clouds. 
However, RANSAC also faces efficiency challenges. 
The introduction of local sampling by Schnabel et al.~\cite{Schnabel_CGF_2007} significantly improves RANSAC’s runtime. 
Still, both RANSAC and Hough Transform are susceptible to producing false positives.

\subsubsection{Traditional global algorithms}
\label{sec::type1}
Traditional global algorithms refer to methods that segment all plane instances in a 3D point cloud simultaneously using conventional, nonlearning-based techniques. 
These approaches can be categorized into three main types: energy optimization–based algorithms, feature clustering–based algorithms, and regularity-based algorithms.

Energy optimization–based algorithms generally begin by generating an initial plane segmentation and then refine plane boundaries through the minimization of an energy function. 
Due to their multistage nature, efficient techniques such as region growing algorithms are often employed for the initial segmentation. 
For example, Yan et al.~\cite{Yan_ISPRS_2014} and Dong et al.~\cite{Dong_ISPRS_2018} first applied region growing and subsequently refined the segmentation using graph cut optimization. 
However, graph cut methods are computationally expensive when applied to large-scale point clouds. 
To improve efficiency, Li et al.~\cite{Li_RS_2020} introduced a local optimization strategy to refine boundary points after the initial region growing step. 
This method achieves effective boundary refinement with significantly reduced computational cost and has been validated in subsequent work~\cite{Li_arXiv_2023}. 
We also adopt the local optimization method from~\cite{Li_RS_2020} in our superpoint generation process.

Feature clustering–based algorithms group 3D points into plane instances by clustering those with similar features. Common clustering methods include K-means, mean shift, and Density-Based Spatial Clustering of Applications with NOISE (DBSCAN). Research in this area has primarily focused on designing or selecting effective features, as the quality of input features directly impacts segmentation accuracy. 
For instance, Sampath and Shan~\cite{Sampath_TGRS_2010} computed normal vectors via eigenvalue analysis and applied fuzzy K-means for segmentation. 
Huang et al.~\cite{Huang_PSA_2006} constructed a five-dimensional feature combining spatial coordinates and normal vectors, and apply the mean shift algorithm, while Wang et al.~\cite{Wang_PR_2013} used mean shift clustering on Gaussian maps. 
Czerniawski et al.~\cite{Czerniawski_AIC_2018} performed DBSCAN clustering using a six-dimensional feature space, while Chen et al.~\cite{Chen_PR_2023} designed a local tangent plane distance metric that combines spatial and normal vector features for DBSCAN clustering. Importantly, DBSCAN determines core, border, and noise points on a global scale, and operates independently of point processing order, qualifying it as a global algorithm.

Modern buildings often exhibit geometric regularities---such as parallelism, coplanarity, orthogonality, and symmetry---due to artificial and construction factors. Leveraging these regularities during plane instance segmentation can significantly improve a model’s robustness to noise, leading to the development of regularity-based segmentation algorithms. 
Given that such regularities typically manifest at a global or semiglobal scale in building point clouds, existing regularity-based plane segmentation algorithms are typically global approaches, such as those in references~\cite{Monszpart_ACM_2015} and \cite{Lin_ISPRS_2020}. 
By definition, regularity-based algorithms cannot be purely local, as the regularities they exploit extend beyond localized regions. 
For instance, the orthogonality of mutually perpendicular planes extracted in references~\cite{Sommer_RAL_2020} and \cite{Wu_CG_2021} represents a semiglobal regularity.

\subsubsection{Deep learning-based algorithms}
\label{sec::type1}
Traditional algorithms often rely on manual tuning of input parameters for different datasets, resulting in inefficiencies and limited effectiveness. 
With the growing success of deep learning across various domains, many point cloud processing tasks have increasingly adopted deep learning–based approaches.
These methods, known for their automation and efficiency, have emerged as a promising direction for plane instance segmentation.
Zhang and Fan~\cite{Zhang_PR_2022}, for example, employed PointNet++ \cite{Qi_ANIPS_2017} as the backbone of a multitask point cloud processing network capable of segmenting roof plane instances while also recognizing groups of roof planes with similar geometric properties. 
Similarly, Li et al.~\cite{Li_ISPRS_2024} used the PointNet++ backbone within a multitask framework to separately extract features in both Euclidean and embedding spaces, which are then combined to cluster plane instances.
Kong and Fan \cite{Kong_TGRS_2024} retrained the superpoint Transformer-based 3D point cloud instance segmentation network, SPFormer \cite{Sun_AAAI_2023}, on the RoofNTNU dataset~\cite{Zhang_PR_2022} and achieved promising results.
However, the RoofNTNU dataset is too small, containing only 930 roof samples, which makes it insufficient to fully validate the effectiveness of SPFormer in the field of point cloud roof plane segmentation.

Deep learning-based algorithms, particularly Transformer-based approaches, represent the most promising future research direction for plane instance segmentation in 3D point clouds.
However, the current SOTA models still utilize PointNet++ as the backbone network~\cite{Zhang_PR_2022,Li_ISPRS_2024}, underutilizing global contextual information, which may limit segmentation performance.
Moreover, unlike the experiments in \cite{Kong_TGRS_2024}, we tested SPFormer \cite{Sun_AAAI_2023} on two larger-scale plane segmentation datasets and found that its performance was far from satisfactory.
Therefore, in this paper, we propose a superpoint Transformer tailored for point cloud plane instance segmentation by generating high-quality superpoints, incorporating handcrafted features, and integrating the Kolmogorov-Arnold Network. 
Additionally, we further enhance segmentation results through postprocessing involving plane completion and boundary refinement.

\subsection{Superpoints in superpoint Transformers}
\label{sec::mfmethods}
In recent years, superpoint Transformers have demonstrated significant advantages over point Transformers in 3D point cloud processing \cite{Landrieu_CVPR_2019,Liang_ICCV_2021,Sun_AAAI_2023,Robert_ICCV_2023,Zhu_arXiv_2024}. 
Superpoints, analogous to superpixels in two-dimensional imagery, are generated by over-segmenting a 3D point cloud into spatially coherent regions. 
Replacing raw 3D points with superpoints as the fundamental processing units greatly reduces computational costs in dense prediction tasks, offering a new paradigm for a wide range of 3D point cloud applications. 
As a result, effectively generating superpoints and incorporating them into Transformer architectures has become a key challenge in the field.

However, most existing superpoint Transformers primarily focus on how superpoints are utilized, often overlooking the critical issues of how they are generated. 
Landrieu and Simonovsky~\cite{Landrieu_CVPR_2018} developed a global energy optimization-based method for unsupervised over-segmentation, aiming to produce superpoints that are both geometrically simple and semantically homogeneous. 
Building on this method, Landrieu and Boussaha~\cite{Landrieu_CVPR_2019} introduced a supervised learning framework called Supervised SuperPoint (SSP), which computes local geometric and radiometric embeddings for each point and constructs an adjacency graph. 
Superpoints are then defined as piecewise constant approximations of these embeddings on the adjacency graph. 
This approach has been widely applied in subsequent superpoint Transformers \cite{Liang_ICCV_2021,Hui_ANIPS_2022}. 
Sun et al.~\cite{Sun_AAAI_2023} proposed an end-to-end 3D instance segmentation method, SPFormer, which uses superpoint generation method in reference~\cite{Landrieu_CVPR_2018}. 
In contrast, Robert et al.~\cite{Robert_ICCV_2023} used K-nearest neighbor clustering to split point clouds into a hierarchical superpoint structure.

While these methods have demonstrated the utility of superpoints in various applications, they often fall short in enhancing feature representation of superpoints or optimizing them for deep learning. 
Graph-based segmentation can result in discontinuities or irregularities within superpoints and is often computationally inefficient. 
Conversely, clustering–based approaches may produce superpoints with imprecise boundaries, compromising segmentation accuracy.

Given that real-world 3D scans may contain hundreds of millions of points, improving the efficiency of 3D analysis remains a critical research goal. 
Using superpoints as the processing units can greatly enhance the efficiency of Transformer models; however, the effectiveness is highly dependent on the quality of the superpoints. 
Poorly generated superpoints---especially near semantic or instance boundaries---can result in segmentation errors. 
Conversely, high-quality superpoints can significantly boost model performance. 
Therefore, identifying the characteristics of high-quality superpoints for Transformer-based architectures and thus developing effective superpoint generation methods is the primary objective of this study.

\subsection{KAN and its application}
\label{sec::enermethods}
KAN \cite{Liu_ICLR_2024} introduces a novel neural network architecture grounded in the Kolmogorov–Arnold representation theorem \cite{kolmogorov_AMS_1961}. 
It fundamentally differs from traditional multilayer perceptrons (MLPs) \cite{Rosenblatt_PR_1958,Rumelhart_NATURE_1986} in both design philosophy and implementation. 
Instead of utilizing fixed nonlinear activation functions, KAN learns nonlinear relationships directly. 
Each KAN layer consists of multiple continuous univariate function mappings applied element-wise to the input vector. 
These function mappings act as learnable activation functions, combined through summation, allowing the network to perform nonlinear transformations via parameterization---ultimately enhancing its expressive power. 
Functionally, KAN resembles a two-layer MLP: the first layer performs an internal summation, while the second layer applies the learnable activation functions and aggregates the activation results. 
This architecture not only amplifies the network’s capacity for modeling complex nonlinearities but also introduces greater adaptability through its flexible activation functions.

KAN’s design leverages B-splines for local control and adjustable grid resolution, enabling a form of dynamic network structure and supporting continual learning. Variants of KAN have explored alternative basis functions such as Chebyshev polynomials \cite{Sidharth_arXiv_2024}, wavelet functions \cite{Bozorgasl_arXiv_2024}, and other orthogonal polynomials, each offering distinct modeling properties. 
A notable variant, FourierKAN \cite{Imran_arXiv_2024,Xu_arXiv_2024}, replaces B-splines with Fourier series, which excel at modeling continuous, and especially smooth and periodic functions. 
This makes FourierKAN particularly effective across various domains and a strong candidate for replacing MLPs in certain applications. 
In the FourierKAN-GCF model \cite{Xu_arXiv_2024}, FourierKAN is employed in place of the traditional MLP in the feature transformation stage of graph convolutional networks. 
This substitution enhances the representational power of graph collaborative filtering (GCF) \cite{Wang_ACM_2019}, simplifies training, and yields notable performance improvements on graph-structured data. 
Furthermore, FourierKAN’s ease of training and strong representational capabilities enable its integration with architectures such as Transformer, further improving model performance.

Despite these strengths, KAN also has limitations. 
Yu et al.~\cite{Yu_arXiv_2024} conducted a comprehensive comparison between KAN and MLP across multiple tasks, revealing that KAN is more prone to catastrophic forgetting and that MLP still outperforms it on most datasets. 
Similarly, Cang et al.~\cite{Cang_arXiv_2024} demonstrated KAN’s heightened sensitivity to noise. Nevertheless, KAN remains a promising innovation in neural network research, offering a fresh perspective and the potential to inspire new algorithmic developments. 
With continued investigation, improved variants of KAN may emerge, refining its architecture and training strategies to fully realize its potential in deep learning.

In our proposed superpoint Transformer-based point cloud plane instance segmentation method, KAN plays a central role, leveraging its strengths in nonlinear modeling. 
By utilizing parameterized and learnable activation functions in place of traditional linear weight matrices, KAN matches or exceeds the performance of standard MLPs while maintaining a compact model size. 
This contributes to improved feature extraction and instance prediction, thereby enhancing the accuracy of segmentation. 
Our experimental results further validate KAN’s effectiveness in the 3D point cloud plane instance segmentation task.

\section{Our approach}
\label{sec:approach}
\begin{figure*}[!h]
	\centering
	{\includegraphics[width= 0.99 \linewidth]{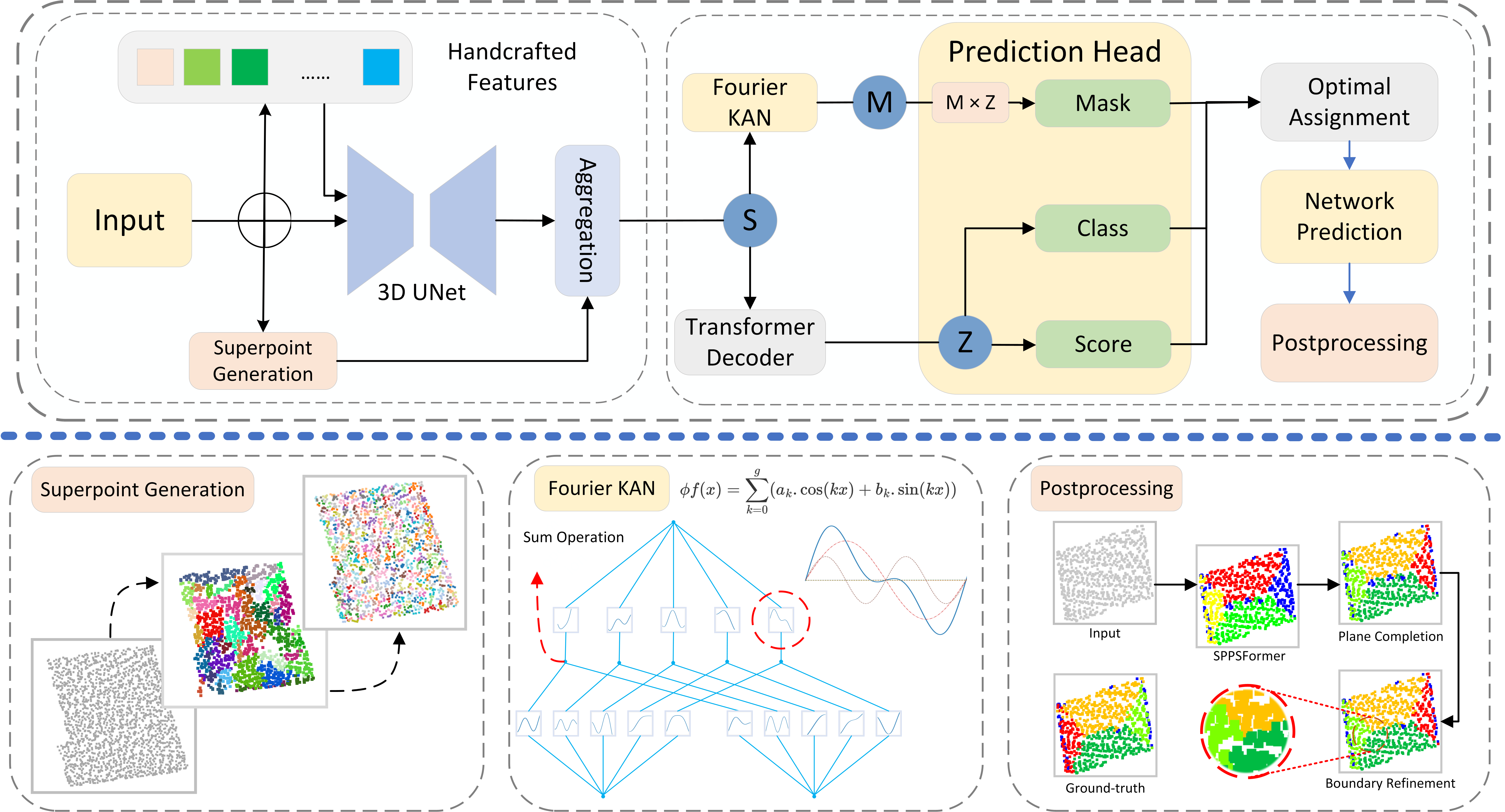}} 
	\caption{Overall architecture of our SPPSFormer. \textbf{S} denotes the superpoint features obtained by the superpoint pooling layer, \textbf{Z} represents the query vector features produced by the Transformer decoder, and \textbf{M} indicates the mask features extracted using FourierKAN.}
	\vspace{0.0em}
	\label{Fig:1}
\end{figure*}

%
Fig.~\ref{Fig:1} illustrates the architecture of the proposed superpoint Transformer. 
The framework begins with a two-stage superpoint generation strategy to produce high-quality superpoints from the input point cloud. 
Subsequently, the point cloud is voxelized, and multidimensional handcrafted features along with spatial position information are integrated into the initial representation. 
A 3D U-Net is then employed to extract per-point features from this voxelized input. 
Using the precomputed superpoints, the per-point features are passed through a superpoint pooling layer, which performs average pooling within each superpoint to produce aggregated superpoint features. These features are then fed into a Transformer decoder, which captures plane instance information from the superpoints. 
To enhance our model’s discriminative capability, mask-aware features are extracted using FourierKAN \cite{Imran_arXiv_2024,Xu_arXiv_2024}. 
Finally, the label assignment is formulated as an optimal matching problem. 
By applying bipartite matching to the predicted superpoint masks, our model supports end-to-end training.

At the prediction stage, our model directly outputs superpoint masks for K plane instances from an input point cloud. 
Each instance is associated with an IoU-aware score $\left \{ {S}_{i}\right \}$ and a corresponding mask confidence score ${\left \{ m{S}_{i}  \in \left [ 0,1\right ]\right \}}^{K}$. 
To derive the final prediction, we compute a combined score $\tilde{{S}_{i}}=\sqrt[2]{{S}_{i}\times{mS}_{i}}$, which is then used for ranking and filtering, eliminating the need for non-maximum suppression or other postprocessing steps, and thus ensuring fast inference.
Given that our network performs plane instance segmentation based on a superpoint Transformer, we name it SPPSFormer (SuperPoint Plane Segmentation transFormer).

While our SPPSFormer adopts a similar architecture to SPFormer \cite{Sun_AAAI_2023}, it incorporates several key improvements: 

(1) We establish criteria for generating superpoints tailored to the superpoint Transformer, and propose a corresponding generation method. 

(2) We introduce essential handcrafted features, effectively enhancing deep learning features and thus providing the decoder with stronger input features. 

(3) In the decoder, we replace the standard MLP with FourierKAN, further enhancing the decoder's performance. 

Among these, (1) has the most pronounced impact on performance. In addition to architecture enhancements, we introduce two novel postprocessing techniques to further refine predictions of our network:

(4) To mitigate incomplete plane segmentation due to partially annotated training data, we propose a self-supervised plane completion strategy. 

(5) To refine segmentation at plane boundaries, we design an efficient boundary refinement algorithm.

The details of these contributions are structured as follows.
Section~\ref{sec:3_1} defines the criteria of high-quality superpoints for the superpoint Transformer under limited training data and presents a workflow for generating such high-quality superpoints. 
Section~\ref{sec:3_2} describes the handcrafted features integrated into our neural network. 
Section~\ref{sec:3_3} details the Transformer–FourierKAN decoder architecture. 
Section~\ref{sec:postprocessing} presents the plane completion and boundary refinement methods.

\subsection{Generation of high-quality superpoints for the superpoint Transformer}
\label{sec:3_1}
In this section, we establish the criteria that superpoints well-suited for Transformer-based feature learning should meet, and we introduce a corresponding generation workflow. 
Superpoints constructed according to these criteria enhance the ability of the superpoint Transformer to learn generalizable features---even under limited training data conditions. 
As a result, the network’s performance in plane instance segmentation is significantly improved.


\subsubsection{Superpoint generation criteria for Transformer feature learning}
\label{sec:3_1_1}
Before generating superpoints suitable for Transformer-based processing, we first examine the key characteristics that make superpoints effective as fundamental units in a superpoint Transformer. 
Based on our analysis, ideal superpoints should satisfy the following two conditions:

\begin{itemize}
\item \textbf{Accurate boundaries:} 
Each superpoint should be spatially coherent and contain points from only a single semantic category or instance. 
This requirement is essential because, once superpoints are formed, the Transformer assumes them as fixed processing units. 
If a superpoint contains mixed-category or multi-instance points, this error cannot be corrected by the network and will propagate throughout the segmentation process.

\item \textbf{Uniform size and shape:} 
Superpoints should exhibit consistent size and geometric shape. 
Neural networks typically rely on the assumption that similar inputs lead to similar outputs. 
When superpoints corresponding to the same category or instance vary greatly in size and shape, this assumption breaks down. 
Consequently, the network requires a significantly larger training set to generalize well---an issue that cannot be easily addressed through standard data augmentation, thereby complicating the training process.
\end{itemize}

Our experimental results confirm that superpoints adhering to both of these criteria lead to marked improvements in segmentation performance of the superpoint Transformer.

\subsubsection{Our superpoint generation process}
\label{sec:architecture}
While generating superpoints that meet either of the two criteria outlined in Section~\ref{sec:3_1_1} is relatively straightforward, producing superpoints that simultaneously satisfy both requirements is considerably more challenging. 
For example, current energy optimization–based methods \cite{Xu_AIC_2020,Li_RS_2020,Dong_ISPRS_2018,Yan_ISPRS_2014} are effective at producing superpoints with accurate boundaries. 
However, the resulting superpoints often exhibit significant variability in size, which, as noted in Section~\ref{sec:3_1_1}, undermines the consistency required for effective feature learning in a superpoint Transformer. 
Conversely, enforcing uniform superpoint size tends to compromise boundary accuracy, as it typically necessitates the generation of small superpoints. 
This poses a problem because accurate boundary delineation often depends on reliable plane parameter fitting, which, however, becomes unreliable when performed on overly small superpoints.

To address this trade-off, we propose a two-stage superpoint generation process. 
In the first stage, we generate superpoints with accurate boundaries. 
In the second stage, we perform size adjustments to enforce greater uniformity in the superpoint scale. 
This approach allows us to generate superpoints that simultaneously satisfy both criteria, thereby enhancing the performance and generalization ability of the superpoint Transformer, as discussed in Section~\ref{sec:3_1_1}.

\begin{itemize}
\item \textbf{Stage 1: Generating coarse-grained superpoints with accurate boundaries} 
\end{itemize}

Accurate boundaries are essential for reliably distinguishing between different categories or instances, thereby preventing the creation of superpoints that span multiple categories or instances. 
This accuracy lays a solid foundation for downstream processing. 
However, in real-world point cloud data, complex and ambiguous boundary regions often challenge traditional superpoint generation methods, which tend to blur or misclassify instance boundaries. 
Therefore, the first stage of our pipeline focuses specifically on generating superpoints with high boundary accuracy.

We begin by applying a region growing algorithm to produce an initial coarse plane segmentation. 
To preserve boundary integrity, we use strict growth parameters, which help prevent points of different planes from being grouped into a single superpoint. However, due to the sequential nature of region growing, early-segmented planes may inadvertently claim points from adjacent, yet distinct, planes---resulting in inaccurate boundary delineation. 
To correct such errors, we incorporate the local boundary optimization method proposed by Li et al.~\cite{Li_RS_2020}, which selectively adjusts point assignments near plane boundaries. 
Unlike global optimization techniques, this localized refinement is computationally efficient and sufficient for our purpose. 
Given that our goal is to prevent misassignments at boundaries rather than perform full-scene optimization, a global algorithm is unnecessary.

\begin{figure}[!htb]
    \centering
    \begin{tabular}{cccc}
        \includegraphics[width=0.235\linewidth]{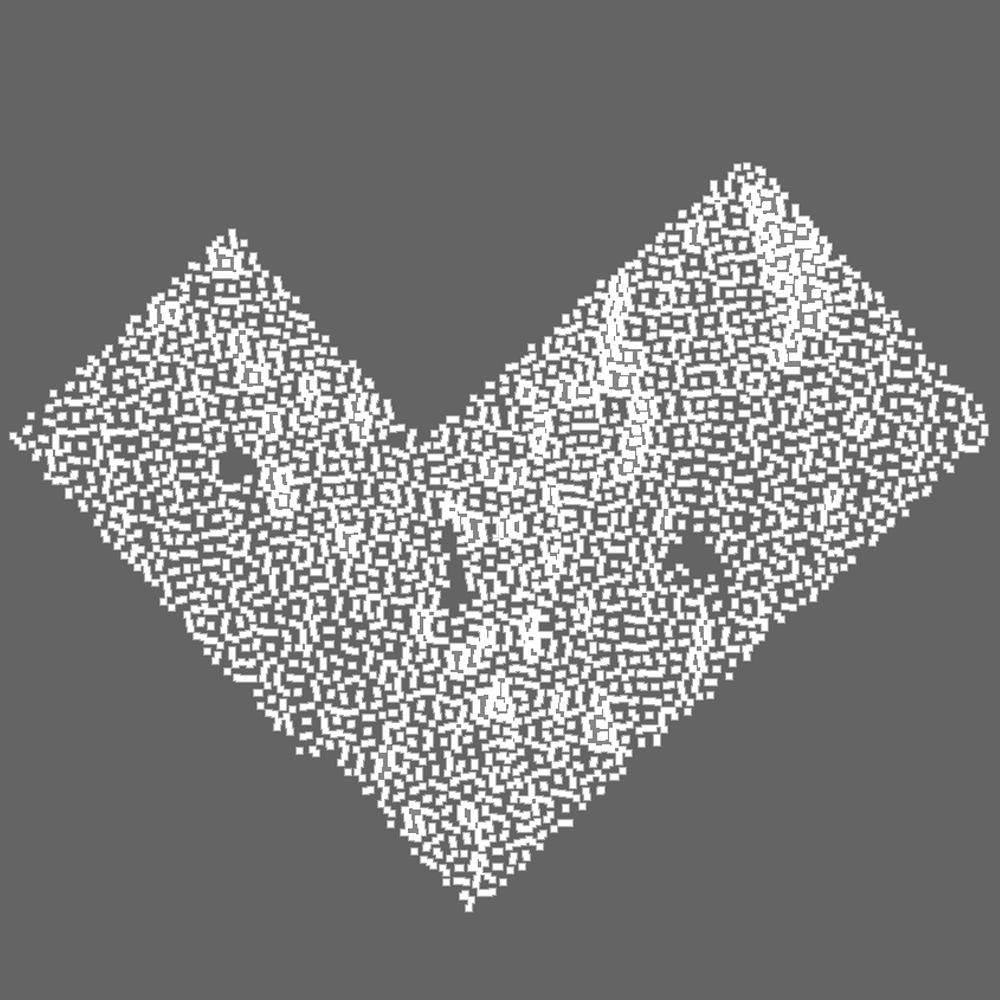}\hspace{1em}&
        \hspace{-1em}\includegraphics[width=0.235\linewidth]{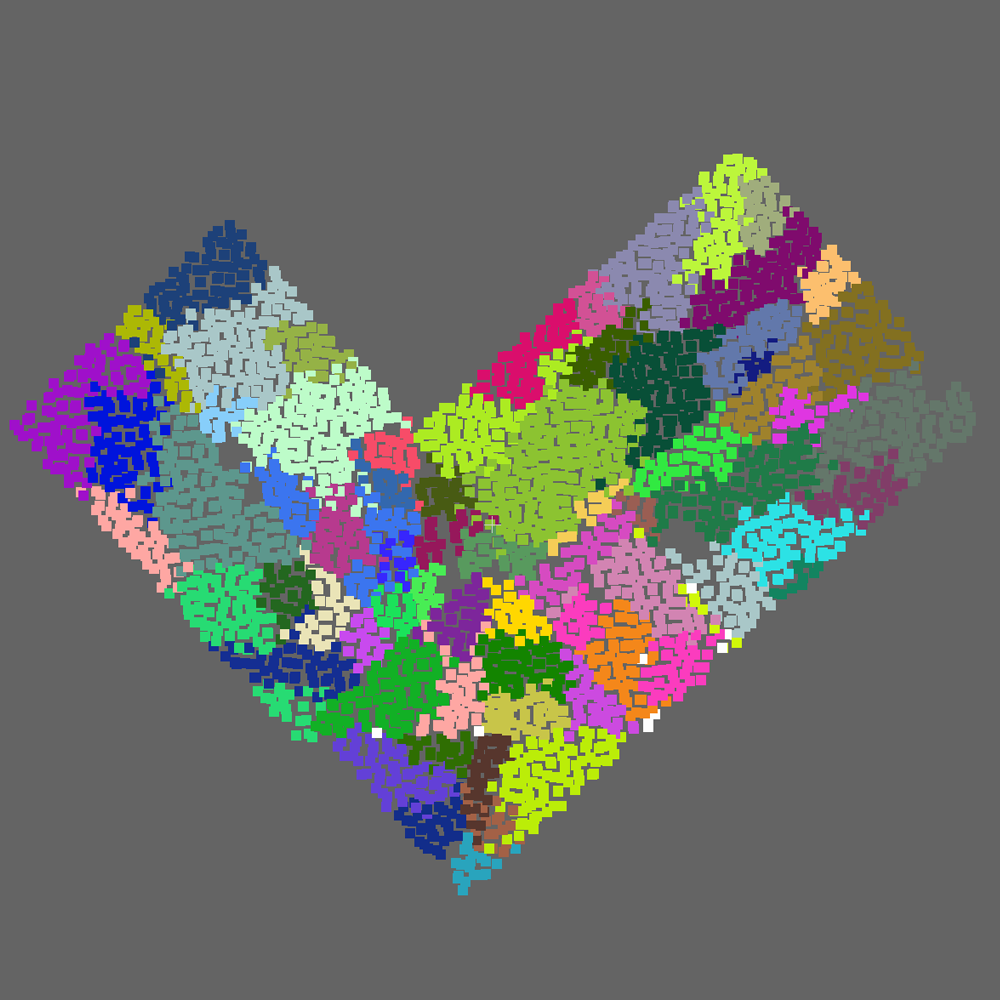}\hspace{-1em} &
        \includegraphics[width=0.235\linewidth]{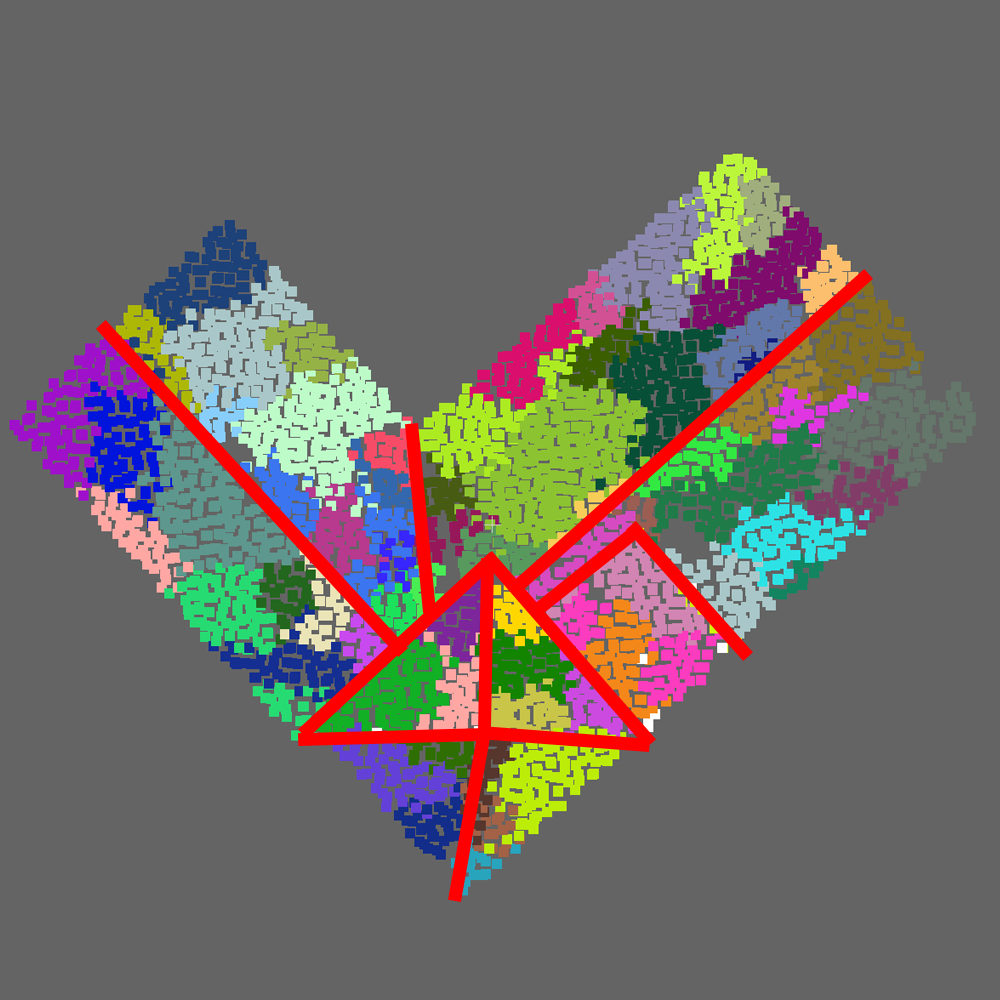}\hspace{-1em} &
        \includegraphics[width=0.235\linewidth]{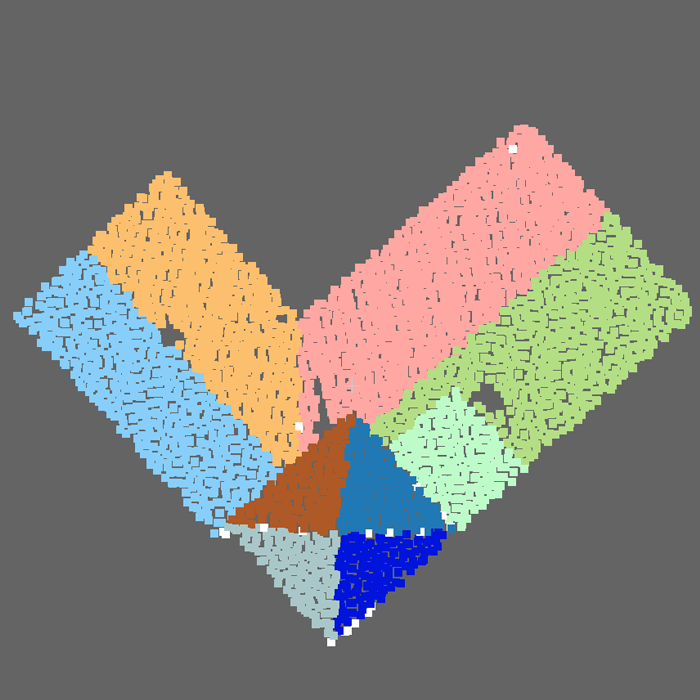} \hspace{-1em}\\
        (a) & (b) & (c) & (d)
    \end{tabular}
    \caption{Illustration of the generation process of our coarse-grained superpoints: (a) shows the point cloud; (b) and (c) depict the generated superpoints, with (c) overlaid with the ground-truth boundaries between planes. 
	A comparison with the ground truth (d) demonstrates that our superpoints align well with true boundaries. 
	Note that noise points in (b), (c), and (d) are marked in white.}
    \label{Fig:2}
\end{figure}

As illustrated in Fig.~\ref{Fig:2}, this two-step process---initial region growing followed by boundary refinement---produces superpoints with well-defined boundaries. However, these superpoints still exhibit considerable variation in size and shape, making them unsuitable at this stage as the basic units for the superpoint Transformer.

It is important to highlight that while our implementation uses region growing for the initial segmentation and the local boundary refinement method from~\cite{Li_RS_2020} for the second step, this design is not exclusive. 
In principle, other algorithms---such as RANSAC, the Hough Transform or watershed-based algorithms---could also be applied during the over-segmentation phase. Similarly, graph cut optimization or other refinement techniques could be used in place of our current boundary refinement method. 
Nevertheless, we select region growing and local boundary refinement~\cite{Li_RS_2020} primarily due to their computational efficiency.

\begin{itemize}
\item \textbf{Stage 2: Generating fine-grained superpoints with consistent sizes and shapes} 

\end{itemize}

To ensure that the superpoints fed into the Transformer possess uniform sizes and shapes, we apply the K-means clustering algorithm to partition the coarse-grained superpoints ${SP}_{\mathrm{coarse}}$ and the unfitted points. 
The desired number of clusters is estimated based on Equation (1).

\begin{equation}
	k= \left \lceil \frac{\left | {P}_{i}\right |}{n}\right \rceil
\end{equation}

\noindent where $n$ denotes the desired average number of points per superpoint and  $\left | {P}_{i}\right |$ represents the total number of 3D points within a given coarse superpoint or the noise set being processed.

K-means clustering is applied to each coarse-grained superpoint based on the 3D spatial coordinates of its constituent points:
\begin{equation}
F_{i} =
\bigg\{
\begin{matrix}
Kmeans(P_{i},k), & \left| P_{i}\right| > n \\ 
Kmeans(P_{i},P_{i}/2), & \left| P_{i}\right|\le n
\end{matrix}
\bigg.
\end{equation}

\noindent where $Kmeans\left ( \right )$ denotes the clustering algorithm, with the first argument being the input point cloud and the second indicating the number of desired clusters. 
In practice, situations may arise where $\left | {P}_{i}\right |<=n$, especially in sparse regions where some coarse-grained superpoints contain few points while covering large spatial space. 
To address this, we increase the number of clusters in such cases to produce superpoints with a more uniform spatial distribution.

Unfitted points, referred to as NOISE, are treated as a single large superpoint and subjected to K-means clustering. 
Because the NOISE may contain significant real noise, we use a smaller average superpoint size during the partition process of the NOISE to ensure that each superpoint predominantly contains points from either the same plane or noise. 
This setting helps reduce the likelihood of mixing plane points with noise points within the same superpoint, thereby improving the network’s performance during feature extraction.

The final set of fine-grained superpoints is formed by dividing the original large-sized superpoints and the unfitted NOISE points, as shown in Equation (3):
\begin{equation}
    {SP}_{\mathrm{fine}} = \Bigl( {\mathop{\scalebox{1.3}{$\bigcup$}}}_{i}{F}_{i} \Bigr) \hspace{1mm} {\bigcup } \hspace{1mm} \Bigl\{ Kmeans\left(\mathrm{NOISE}, 2k\right) \Bigr\}
\end{equation}

Because K-means uses only the spatial coordinates of 3D points and the cluster size is kept relatively small, the resulting superpoints exhibit similar sizes and shapes.

\begin{figure}[!htb]
    \centering
    \begin{tabular}{@{\hspace{0em}}c@{\hspace{0.4em}}c@{\hspace{0.4em}}c@{\hspace{0em}}}
        \includegraphics[width=0.32\linewidth]{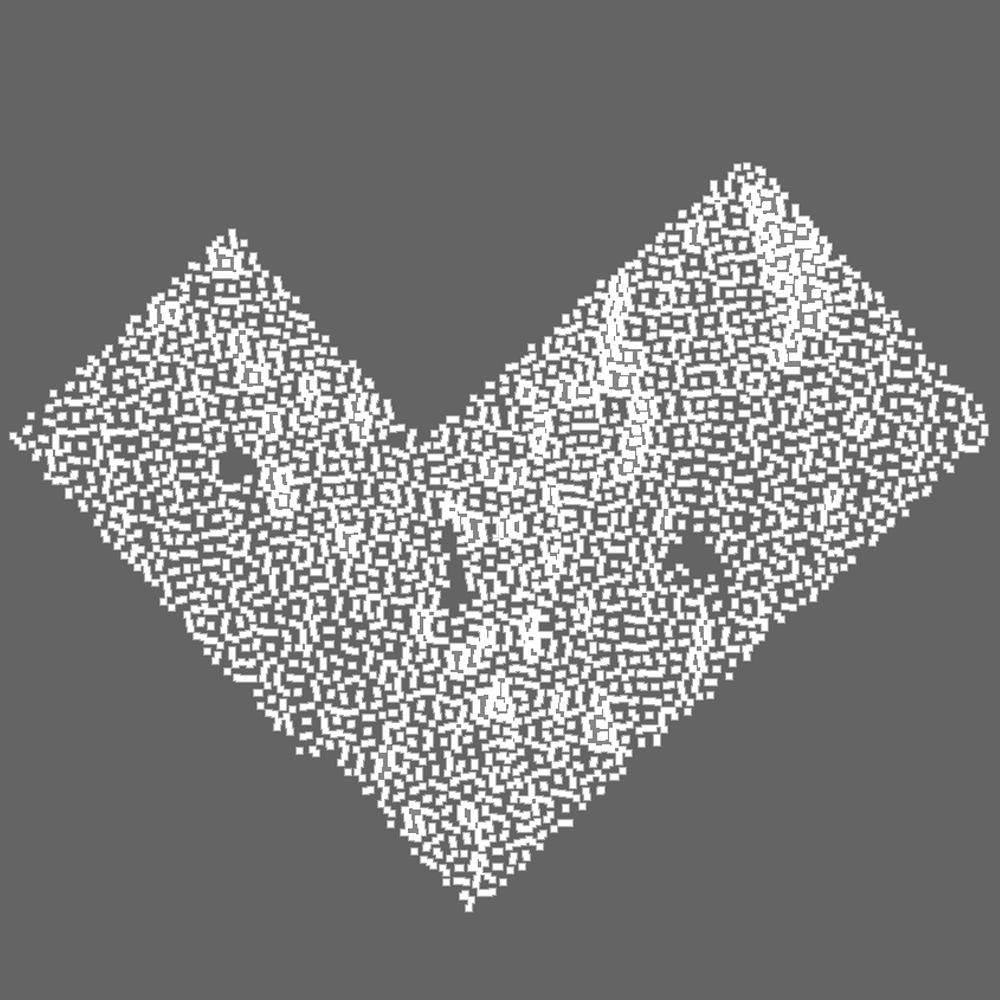} &
        \includegraphics[width=0.32\linewidth]{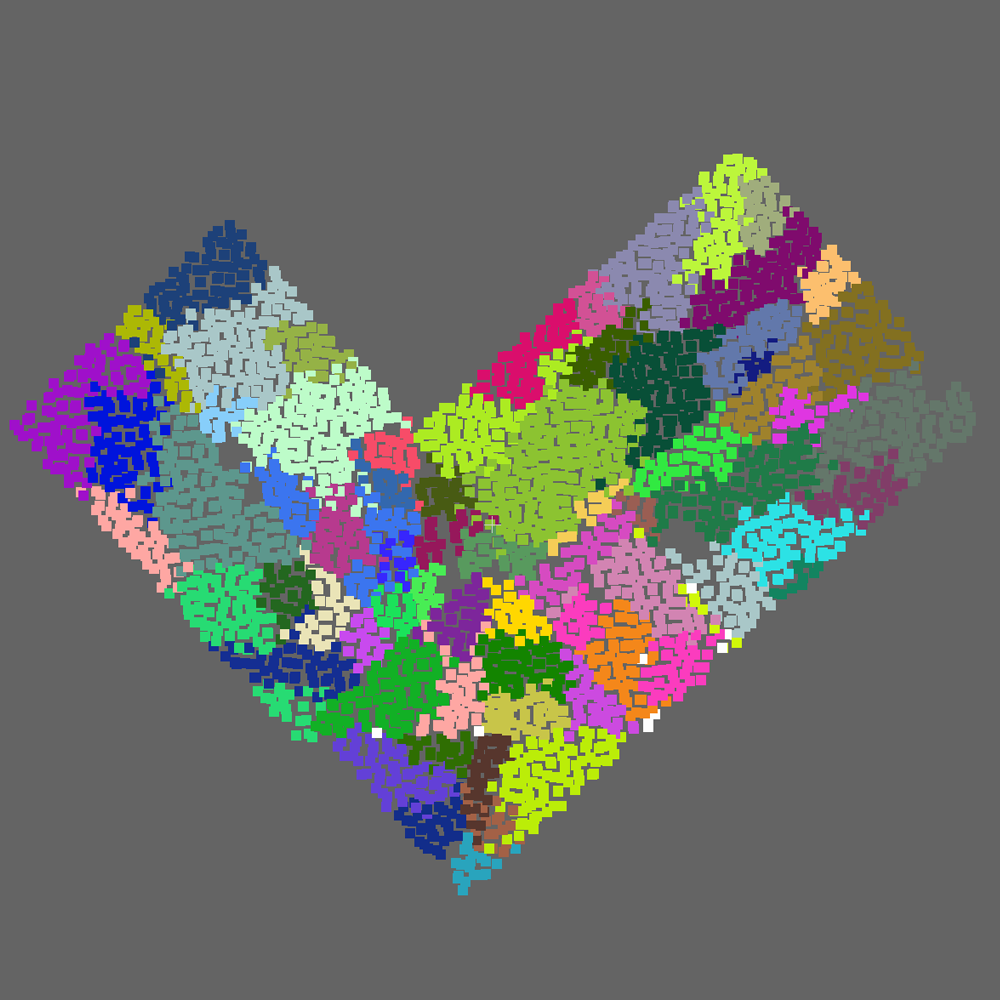} &
        \includegraphics[width=0.32\linewidth]{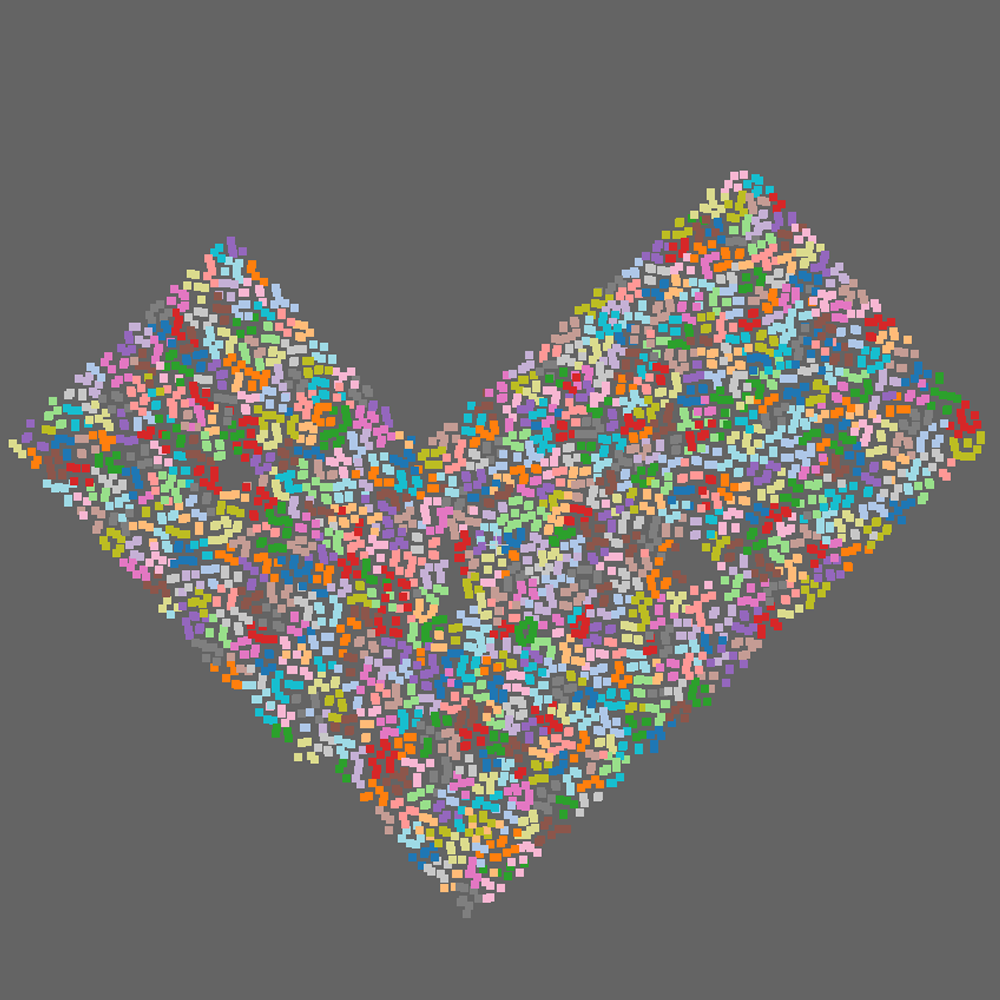} \\
        (a) & (b) & (c)
    \end{tabular}
    \caption{Illustration of our two-stage superpoint generation process. (a) shows the point cloud; (b) presents the superpoints with accurate boundaries produced in the first stage; and (c) displays the fine-grained superpoints with consistent sizes and shapes generated in the second stage.}

    \label{Fig:3}
\end{figure}

Ultimately, we generate superpoints that satisfy both criteria introduced in Section~\ref{sec:3_1_1}. 
Fig.~\ref{Fig:3} presents an illustration of our two-stage superpoint generation strategy. 
In the experimental section, we will evaluate how superpoint quality affects segmentation performance.

Compared to existing approaches, our method yields superpoints with accurate boundaries, uniform sizes, and geometrically consistent shapes. 
These characteristics make the superpoints particularly effective for feature extraction of the superpoint Transformer, enabling generalizable feature learning even under limited training data.

\subsection{Handcrafted features}
\label{sec:3_2}

Deep learning algorithms typically require large training datasets to extract robust and generalizable features \cite{Radford_ICML_2021,Meng_RS_2022,Kirillov_ICCV_2023,Ao_TGRS_2023,Wang_CPGIS_2023,Zhang_IJAEOG_2024}. 
When the training dataset size is limited, the features extracted by deep learning models often have inherent limitations. 
To address this issue, we provide the network with a set of manually designed features as input, in addition to the raw 3D point coordinates, to enhance the expressiveness of the deep learning features. 
These handcrafted features include geometric attributes derived from principal component analysis \cite{Hotelling_JEP_1933}, such as linearity, planarity, scattering, and verticality. Additionally, we introduce a roof contour feature based on normal vectors (contour). 
These handcrafted features serve as complementary inputs to the raw deep learning features, enhancing the model’s ability to capture more nuanced characteristics of the 3D point cloud, particularly in the context of 3D plane instance segmentation. 
Our experiments thoroughly validate the effectiveness of these handcrafted features, demonstrating their critical role in improving model performance.

Below is a brief description of the handcrafted features we employ.

\textbf{Linearity:} Regions exhibiting high linearity~\cite{Demantk_ISPRS_2011} typically correspond to edges or straight-line features, which are crucial for identifying and delineating boundaries and linear structures, such as building edges, during the segmentation process. 
By recognizing these high linearity regions, we can optimize boundary detection between planes, improving the accuracy of the segmentation of linear structures.

\textbf{Planarity:} High planarity regions~\cite{Demantk_ISPRS_2011} are indicative of planar structures, which are often associated with building surfaces. 
Using the planarity feature allows us to enhance the accuracy of identifying and segmenting planar surfaces.
In the context of roof plane segmentation, this feature is critical for accurately recognizing roof planes.

\textbf{Scattering:} The scattering~\cite{Demantk_ISPRS_2011} feature helps distinguish between dense areas (e.g., roofs) and dispersed areas (e.g., noise or railing structures). 
By evaluating the scattering of points, we can better differentiate between valid planar features and noise, thus improving the robustness of segmentation and preventing erroneous classification of sparse noise points as part of the planes.

\textbf{Verticality:} Many important building structures, such as walls, are typically vertical. 
The verticality~\cite{Guinard_ISPRS_2017} feature enables the model to accurately identify vertical planes and distinguish them from horizontal and oblique surfaces. 
By leveraging this feature, we can improve segmentation performance for roof planes.

For a detailed explanation of linearity, planarity, and scattering, please refer to reference~\cite{Demantk_ISPRS_2011}. 
For verticality, please see reference~\cite{Guinard_ISPRS_2017}.

\textbf{Contour:} The contour feature aids in distinguishing roof planes that are not actually connected in 3D space, such as multilayered roofs. 
Incorporating this feature into the network is equivalent to performing a connectivity analysis after the network predicts plane instances, but our approach is more efficient.

The contour feature is computed as follows. 
For each point $P_i$ in the point cloud, the tangent plane is calculated based on its neighboring points. 
For each neighboring point $P_j$ of $P_i$, it is projected onto the tangent plane, and this projection is denoted as $P_j'$. 
Next, for each neighboring point $P_j$ of $P_i$, we compute the vector
\begin{equation}
\mathbf{v}_{ij} = P_j' - P_i.
\end{equation}

These vectors are then sorted in a counterclockwise order by calculating the angle between each vector and a reference vector within the tangent plane of $P_i$ (e.g., the X-axis of $P_i$'s tangent plane), and the vectors are sorted based on these angles. 
The maximum angle $\alpha_i$ between adjacent vectors in the sorted list is computed as follows:
\begin{equation}
\alpha_i = \max \big( \theta(\mathbf{v}_{i1}, \mathbf{v}_{i2}), \theta(\mathbf{v}_{i2}, \mathbf{v}_{i3}), \dots, \theta(\mathbf{v}_{ik}, \mathbf{v}_{i1}) \big),
\end{equation}
where $\theta(\mathbf{v}_a, \mathbf{v}_b)$ represents the angle between vectors $\mathbf{v}_a$ and $\mathbf{v}_b$.

Finally, we classify $P_i$ as a contour point if $\alpha_i$ exceeds a user-defined angle threshold.

\subsection{Decoder integrating Transformer and KAN}
\label{sec:3_3}
In the original SPFormer \cite{Sun_AAAI_2023} architecture, the query decoder consists of two branches: an instance branch and a mask branch. 
The instance branch uses Transformer blocks, while the mask branch extracts mask-aware features via an MLP. 
However, MLPs rely on stacking nonlinear activation functions to approximate complicated mappings. 
For highly intricate target functions, this requires many network layers. 
In the original SPFormer, the mask branch employs a two-layer MLP, which is overly simplistic and lacks sufficient representational power. 
Although a deeper MLP could increase capacity, it would also introduce a substantial increase in parameters, many of which may contribute little to model performance, further increasing both the model complexity and training difficulty.

To address this issue, we replace the MLP with FourierKAN \cite{Imran_arXiv_2024,Xu_arXiv_2024} while keeping the parameter count roughly unchanged, because FourierKAN offers a more parameter-efficient solution for approximating nonlinear functions. 
Under comparable representational capacity, it has fewer parameters, and offers stronger representational capacity at comparable complexity. 
Additionally, FourierKAN substitutes the B-spline coefficients with a one-dimensional Fourier series, using sine and cosine functions, to represent continuous functions that are naturally smooth and periodic. 
In contrast, B-splines, while effective locally, lack the same level of global smoothness. 
As a result, Fourier series are easier to optimize than B-spline designs, and consequently, FourierKAN exhibits better stability.

Another reason for adopting KAN is to assess its effectiveness in processing 3D point clouds, providing insights for future research. 
In some computer vision tasks, KANs have demonstrated advantages over MLPs \cite{Li_arXiv_2024,Xu_arXiv_2024,Imran_arXiv_2024,cheon_arXiv_2024}, whereas in other tasks, the opposite has been observed \cite{Yu_arXiv_2024,Cang_arXiv_2024}. 
As the effectiveness of KANs in computer vision remains an active area of exploration, our investigation into its application for 3D point cloud plane instance segmentation aims to contribute valuable insights, potentially guiding future research in this domain.

\subsection{Postprocessing}
\label{sec:postprocessing}
After obtaining the network's predictions, we further refine the segmentation results by applying two traditional postprocessing modules: plane completion and boundary refinement.

\subsubsection{Self-supervised plane completion}
\label{sec:plane_completion}

\begin{figure}[!htb]
    \centering
      \begin{tabular}{@{\hspace{0em}}c@{\hspace{0.4em}}c@{\hspace{0.4em}}c@{\hspace{0em}}}
        \includegraphics[width=0.32\linewidth]{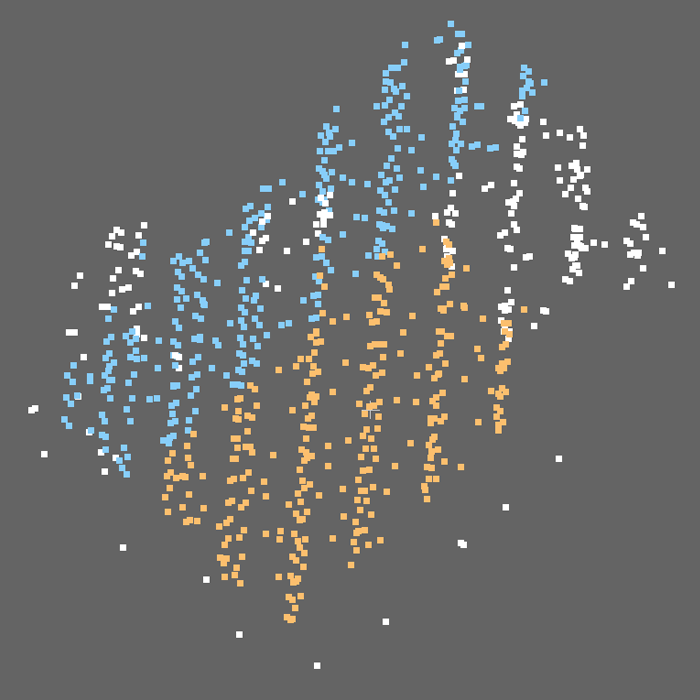} &
        \includegraphics[width=0.32\linewidth]{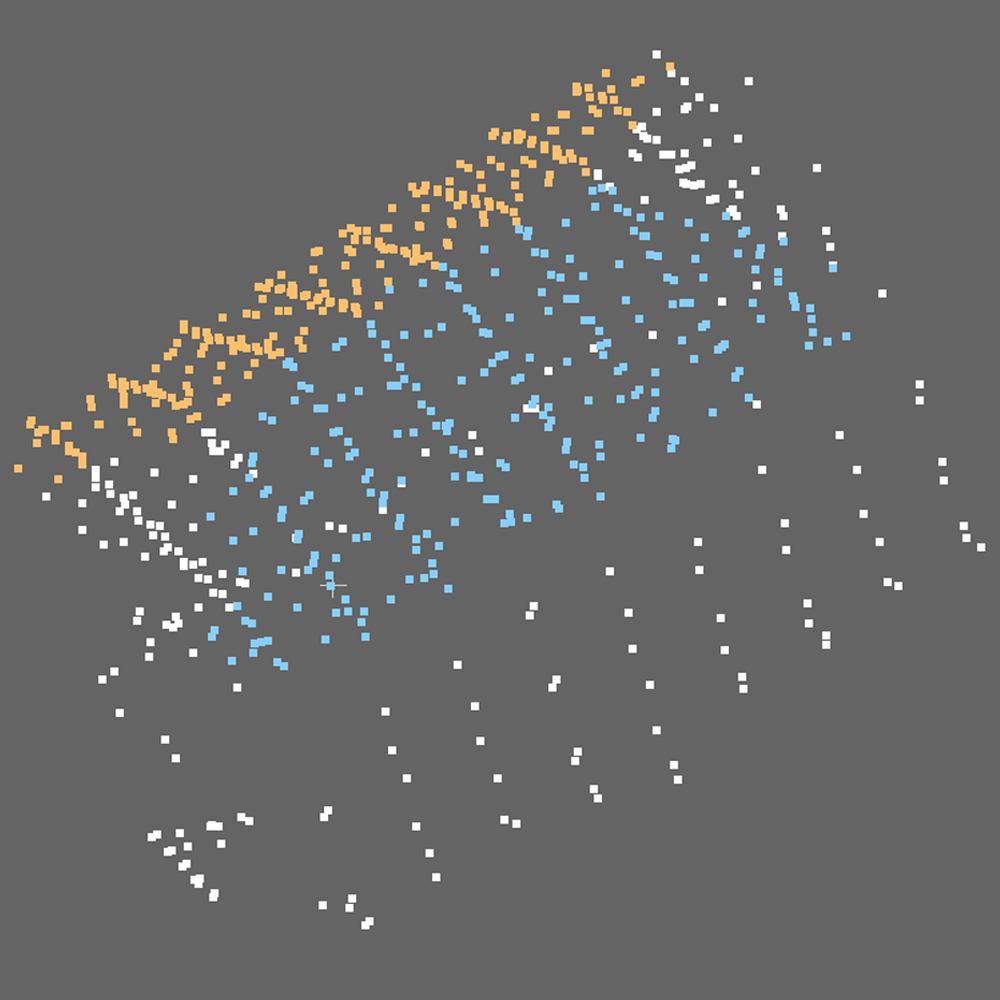} &
        \includegraphics[width=0.32\linewidth]{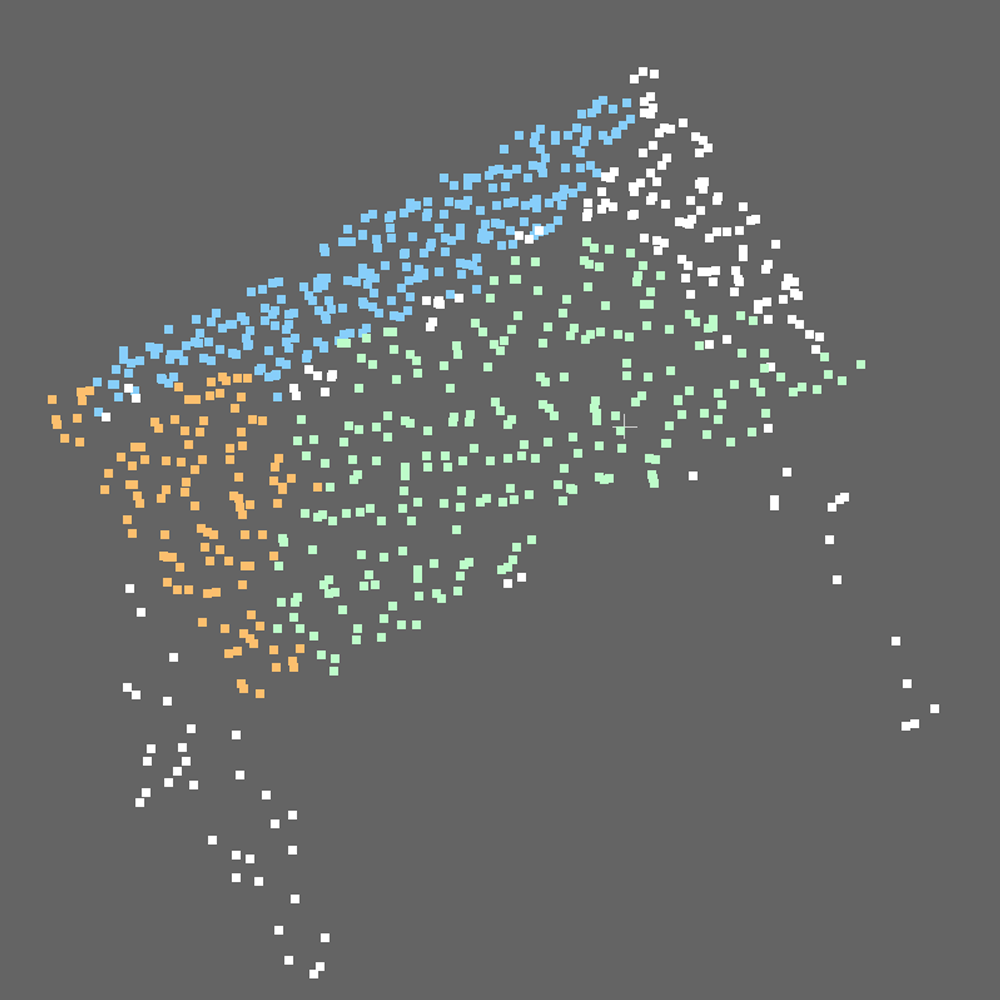} \\
        (a) & (b) & (c)
    \end{tabular}
\caption{Examples of incomplete plane segmentation in the RoofN3D training set and predictions by our SPPSFormer. 
	(a) and (b) show incomplete annotations in the RoofN3D training set, while (c) illustrates a prediction made by our network on the test set. 
	White points indicate nonplane points.}

    \label{Fig:4}
\end{figure}

Real-world plane instance segmentation datasets may suffer from incomplete annotations. 
For example, in the RoofN3D dataset \cite{Wichmann_RS_2019}, some planes are mistakenly labeled as noise points (see Fig.~\ref{Fig:4}). 
These erroneous annotations in the training set may result in incomplete plane segmentation by neural networks (see Fig.~\ref{Fig:4}). 
Therefore, we introduce a plane completion module. 
However, if the annotation quality of the dataset is sufficiently high, this postprocessing step becomes unnecessary. 
For instance, when processing the Building3D dataset created by us, this module is redundant.

Research across various fields has shown that traditional algorithms often struggle in complex scenarios because these algorithms usually need different parameters for different data \cite{Meng_GRSS_2014,Georganos_RS_2018,Meng_TGRS_2018,Wang_JGS_2021}, or even different parts of the same data sample \cite{Felzenszwalb_IJCV_2004,Rao_ECCV_2010,Yang_RSE_2017,Su_ISPRS_2019}. 
However, in real-world applications, it is challenging to fine-tune parameters for each individual data sample or even for different parts of the same data sample. 
Fortunately, the plane completion task we address here involves relatively simple scenarios. 
This is because our neural network has already segmented the majority of the planes from the point cloud, and the remaining nonplane data are minimal. 
Thus, a single set of global parameters is sufficient to achieve satisfactory results for an individual point cloud. 
Therefore, with reasonable global parameters, region growing is capable of producing satisfactory segmentation for these simple point cloud data; considering also that region growing is highly efficient, we employ it to segment any real planes missed by our network. 
However, as mentioned earlier, the challenge remains that different point clouds may require distinct global parameters.

To address the issue of global parameters for region growing in nonplane points across different point cloud data, we propose inferring the necessary parameters from the segmented plane instances. 
In our region growing approach, we rely on the perpendicular distance from a point to the plane and the cosine distance between the point’s normal and the plane’s normal. 
Thus, the global parameters that need to be estimated are related to these two features. 
Given that the precision of plane instances may vary across different data, we infer the required parameters from the existing plane instances in the respective point cloud. 
First, we assume that the existing plane instances were obtained using our region growing algorithm. 
Based on this assumption, we can retrospectively infer the optimal segmentation parameters for each individual plane instance with respect to the two  features used.
Then, we set both needed thresholds for initially missed planes to the maximum values of the corresponding thresholds across the existing plane instances within the point cloud, because the planes missed by the network typically have lower precision. 
It is also important to note that we discard planes containing very few points before performing plane completion, as they are likely false detections.

\subsubsection{Efficient boundary refinement}
\label{sec:boundary_refinement}

When using only neural networks, the resulting segmentation often suffers from inaccurate boundaries, including in our network. 
Additionally, our region growing algorithm for plane completion, being sequential, struggles to effectively segment 3D points at plane boundaries. 
As a result, after plane completion, we apply a boundary refinement step. 
To improve efficiency, we focus solely on the 3D points at plane boundaries and avoid using energy-based optimization. 
Instead, we utilize a composite distance---defined by both the perpendicular distance from a point to the plane and the cosine distance between the point’s normal and the plane’s normal---to assign boundary points to the closest plane. 
The composite distance is formulated as follows:
\begin{equation}
	{dis}_{\mathrm{composite}} = \lambda \cdot {dis}_{\mathrm{P2P}}+ {dis}_{\mathrm{N2N}}
\end{equation}

\noindent where ${dis}_{\mathrm{P2P}}$ represents the perpendicular distance from the point to the plane, ${dis}_{\mathrm{N2N}}$ is the cosine distance between the point's normal and the plane's normal, and $\lambda$ is the weight of ${dis}_{\mathrm{P2P}}$. 
We recommend setting $\lambda > 1$ because  ${dis}_{\mathrm{P2P}}$ is generally more robust than ${dis}_{\mathrm{N2N}}$.

In our model, we utilize two types of boundary refinement algorithms: the local boundary refinement algorithm from reference~\cite{Li_RS_2020} used in Section~\ref{sec:architecture} and our proposed algorithm used in this section. 
The key difference between the two lies in the trade-off between efficiency and applicability: the algorithm from reference~\cite{Li_RS_2020} is less efficient but applicable to a wide range of scenarios, while our own algorithm is more efficient but is specifically suited for cases where the main bodies of the roof planes have been well segmented. 
During the superpoint generation phase, the segmentation quality of the roof planes’ main bodies is relatively poor, so we apply the local boundary refinement algorithm from reference~\cite{Li_RS_2020}. 
In contrast, when refining the predictions of our network---where the main bodies of the roof planes are already well segmented---we opt for our own more efficient boundary refinement algorithm.

\section{Experiments}
\label{sec:experiments}

\subsection{Datasets}
\label{sec:dataset}

We evaluated our model on two real-world datasets to assess its generalization capability in practical applications.

The first dataset we used is the RoofN3D dataset \cite{Wichmann_RS_2019}, which contains 118,074 buildings from New York City. 
However, a majority of the data  are sparse, and some of the annotations are of low quality \cite{Wang_ICCV_2023}. 
For our experiments, we selected data samples with at least 700 points, resulting in a total of 11,189 buildings. 
Among these, 8,189 buildings were used for training, 2,000 for validation, and 1,000 for testing. 
It is worth noting that 457 samples in the original RoofN3D dataset contain annotation errors. 
Therefore, we corrected these errors and conducted experiments on both the original and corrected versions of the dataset.

The second dataset we used is the Building3D dataset \cite{Wang_ICCV_2023}, which spans over 160,000 buildings across 16 cities in Estonia, covering an area of approximately 998 km$^2$. 
To date, approximately 40,000 annotated buildings from this dataset have been released. 
However, the annotations provided only include meshes and wireframes, lacking plane instance segmentation labels. 
We annotated the roof plane instances for 10,539 buildings based on the available wireframe annotations. 
Of these, 7,539 buildings were used for training, 2,000 for validation, and 1,000 for testing.

\begin{figure*}[!htb]
    \centering
    \renewcommand{\arraystretch}{1.2}  
    \resizebox{\textwidth}{!}{%
        \begin{tabular}{
            @{\hspace{0cm}} c  
            @{\hspace{0.15cm}} c
            @{\hspace{0.15cm}} c
            @{\hspace{0.15cm}} c
            @{\hspace{0.15cm}} c
            @{\hspace{0.15cm}} c
        }
            & \makecell{\footnotesize \textbf{PC-1} \\ (RoofN3D)} &
            \makecell{\footnotesize \textbf{PC-2} \\ (RoofN3D)} &
            \makecell{\footnotesize \textbf{PC-3} \\ (Building3D)} &
            \makecell{\footnotesize \textbf{PC-4} \\ (Building3D)} &
            \makecell{\footnotesize \textbf{PC-5} \\ (Building3D)} \\
            
            \multirow{2}{*}[4.8em]{\rotatebox[origin=c]{90}{\textbf{Top-view}}} &
            \includegraphics[width=2.5cm]{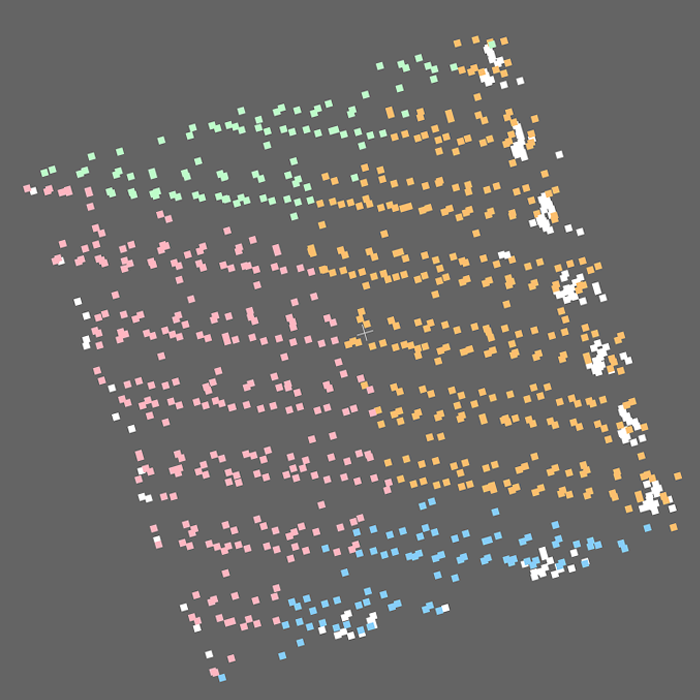} &
            \includegraphics[width=2.5cm]{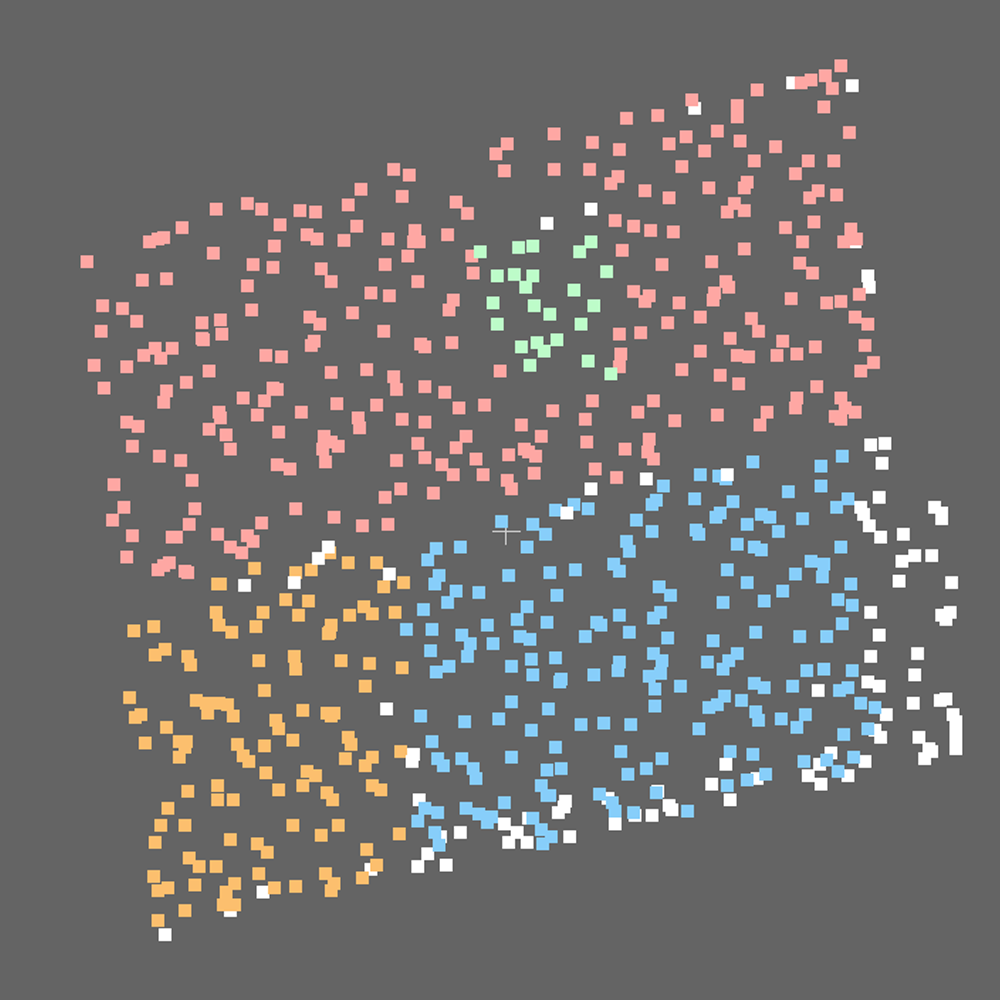} &
            \includegraphics[width=2.5cm]{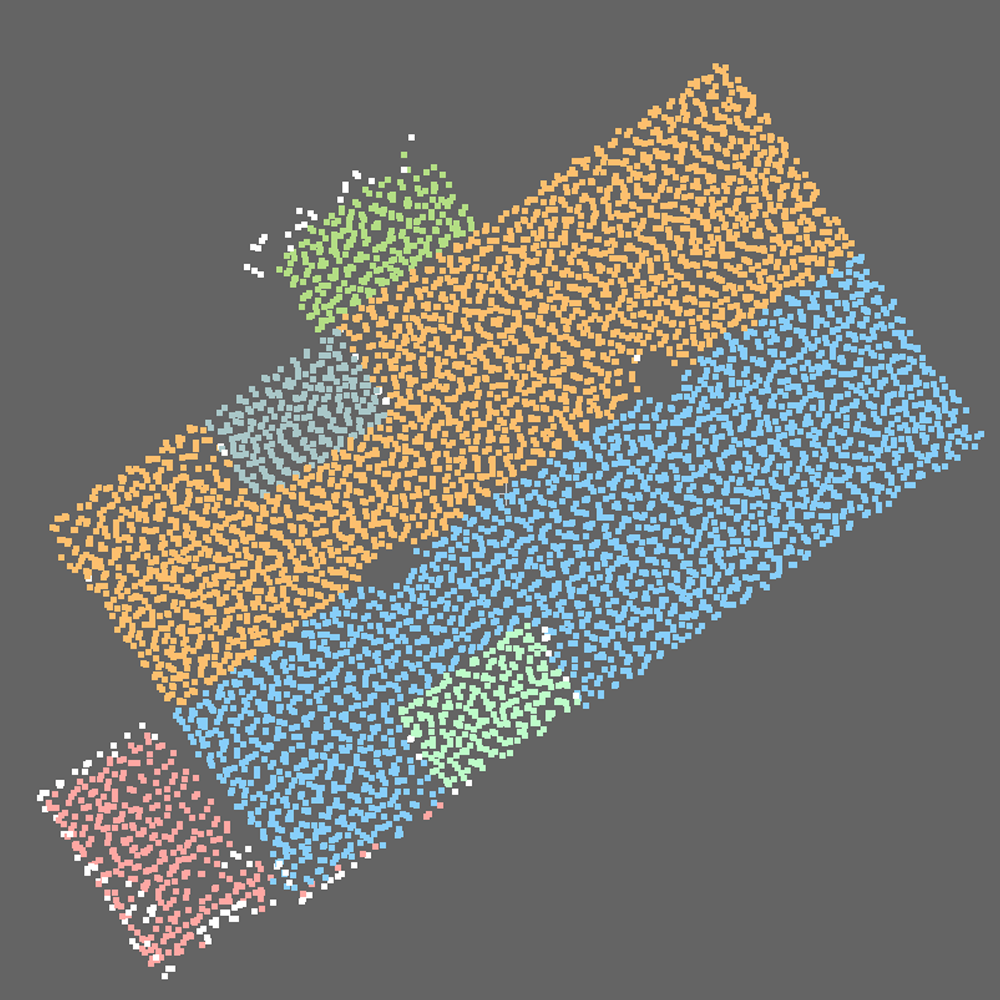} &
            \includegraphics[width=2.5cm]{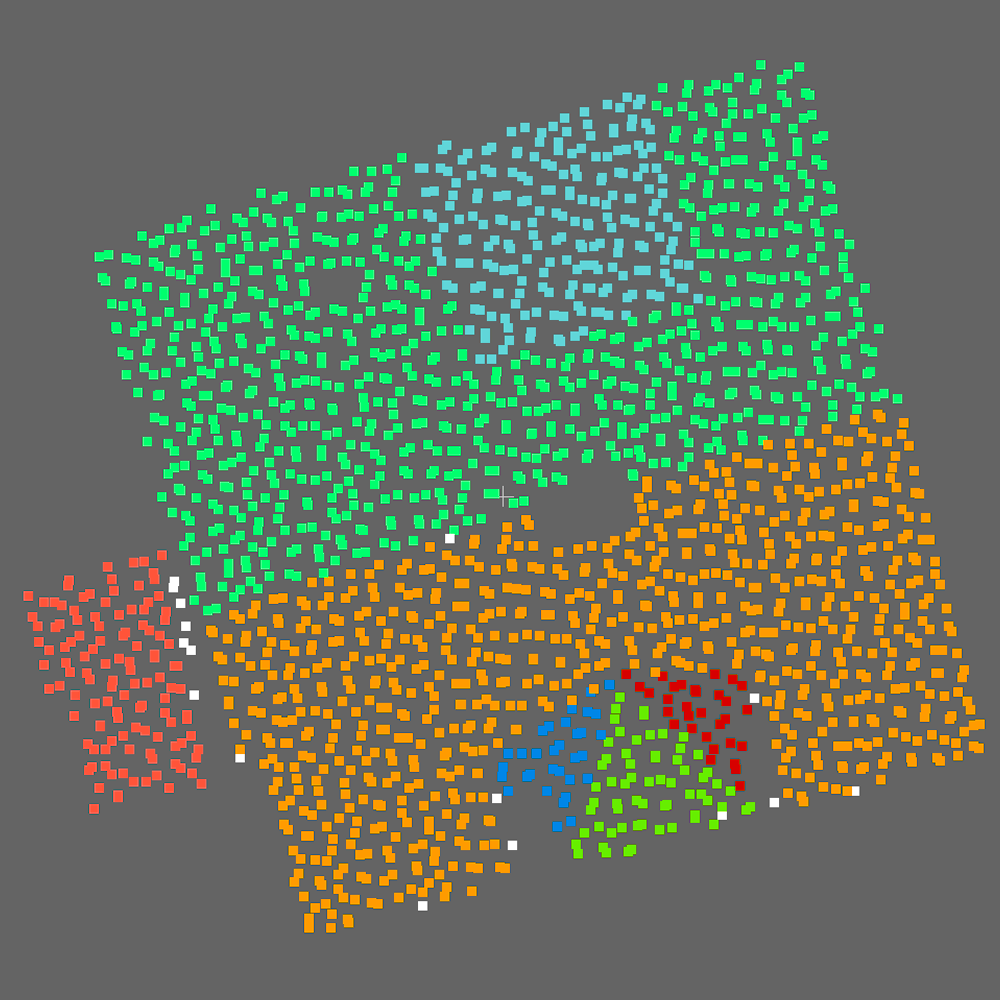} &
            \includegraphics[width=2.5cm]{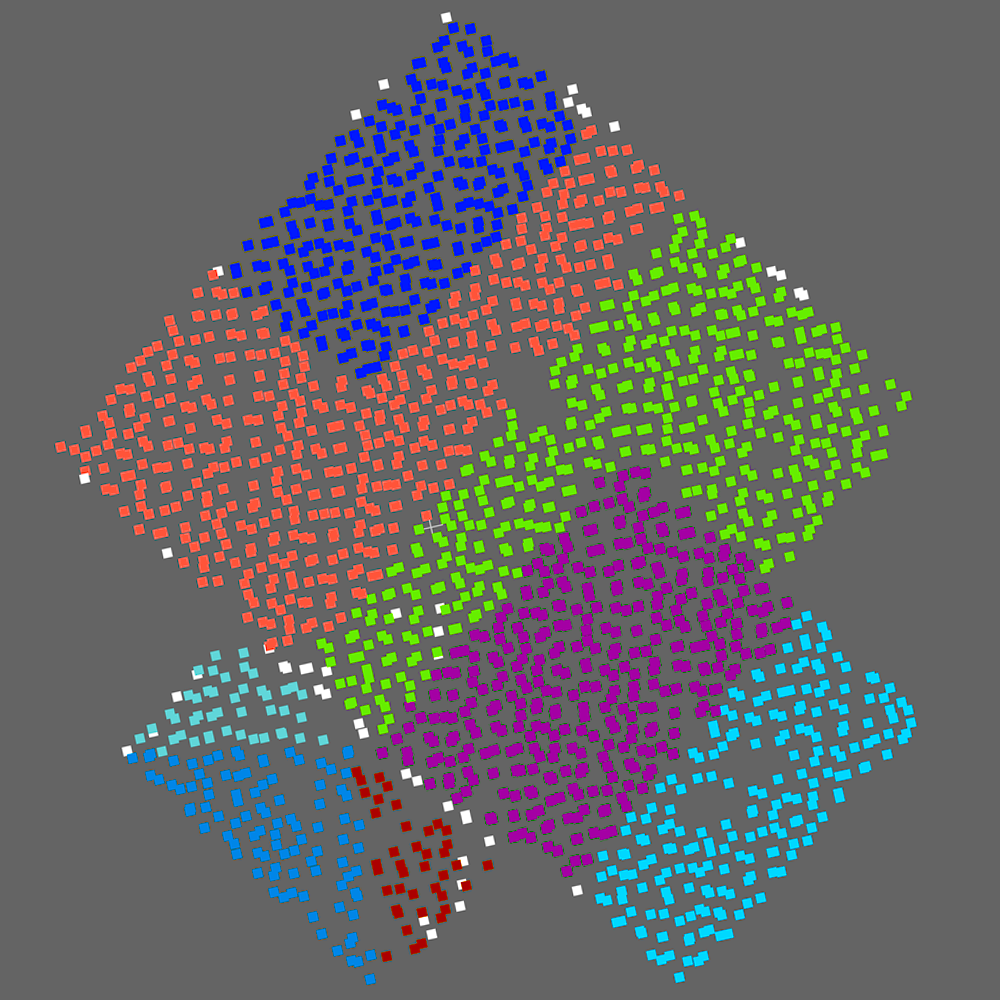} \\
            
            \multirow{2}{*}[4.8em]{\rotatebox[origin=c]{90}{\textbf{Side-view}}} &
            \includegraphics[width=2.5cm]{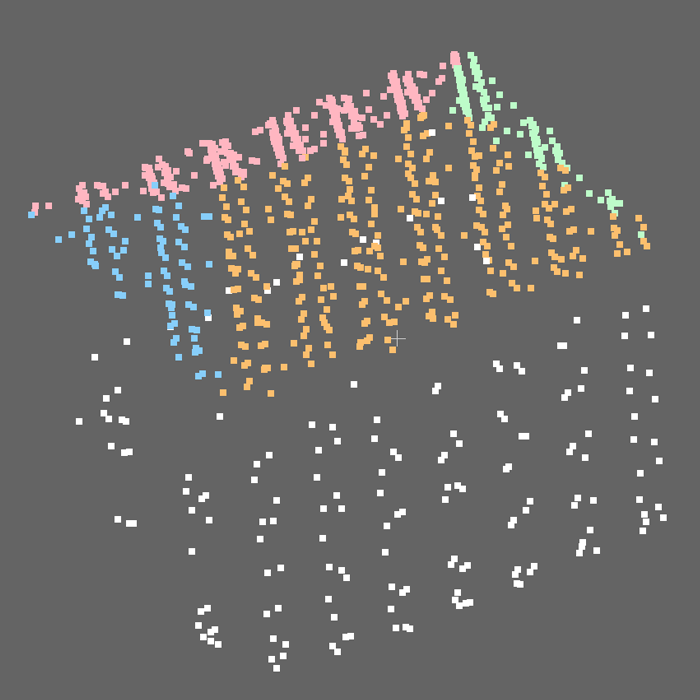} &
            \includegraphics[width=2.5cm]{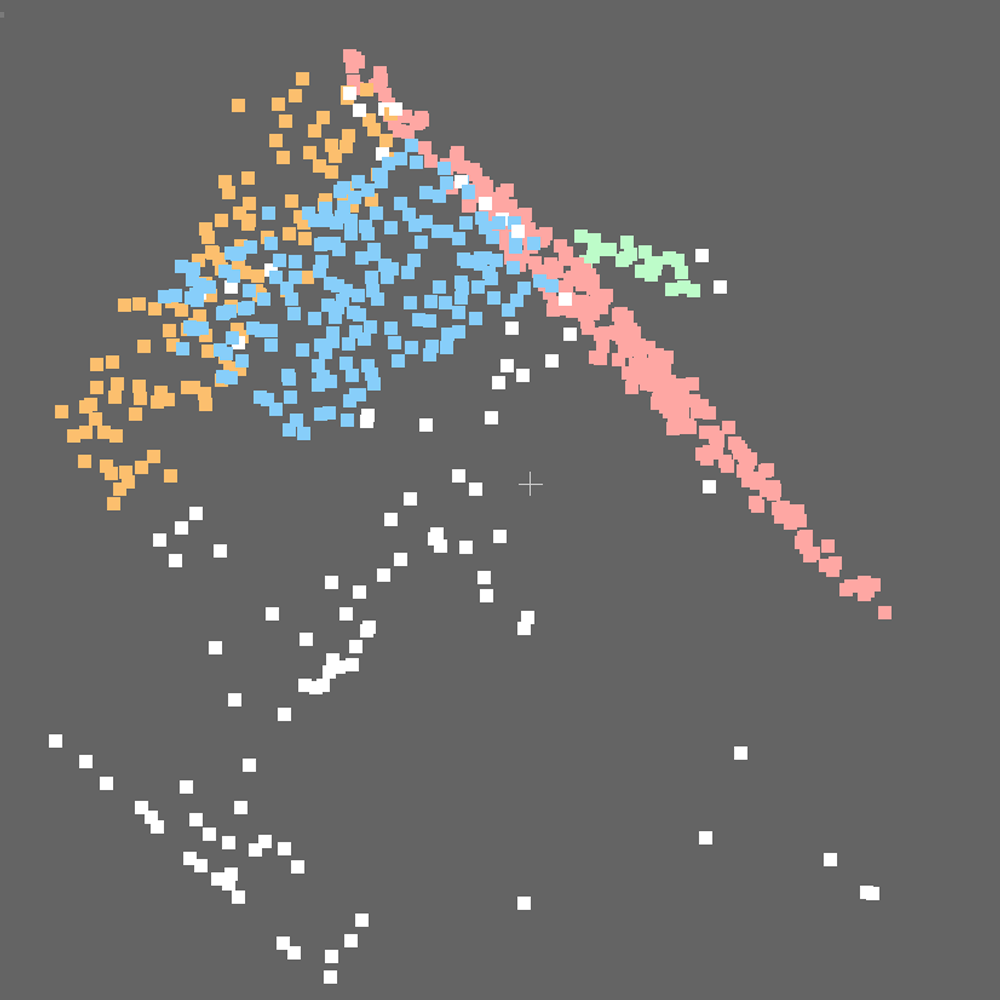} &
            \includegraphics[width=2.5cm]{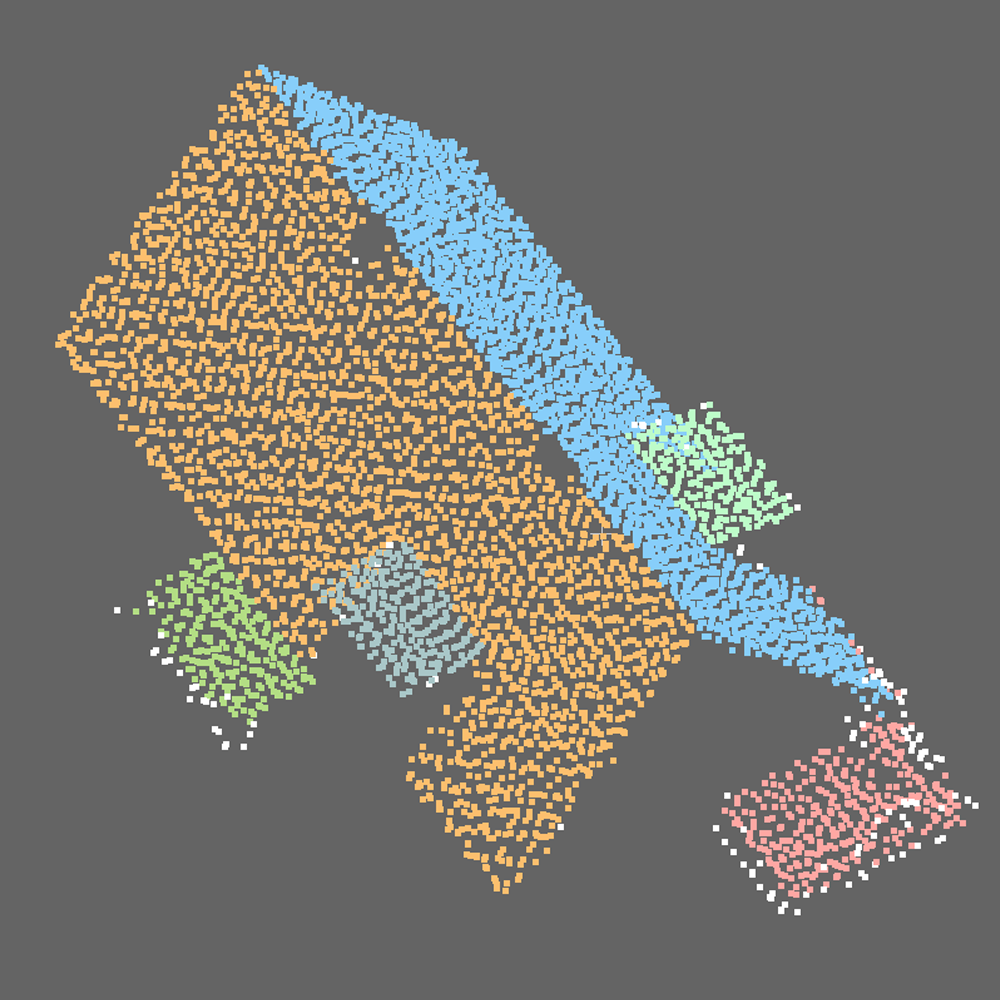} &
            \includegraphics[width=2.5cm]{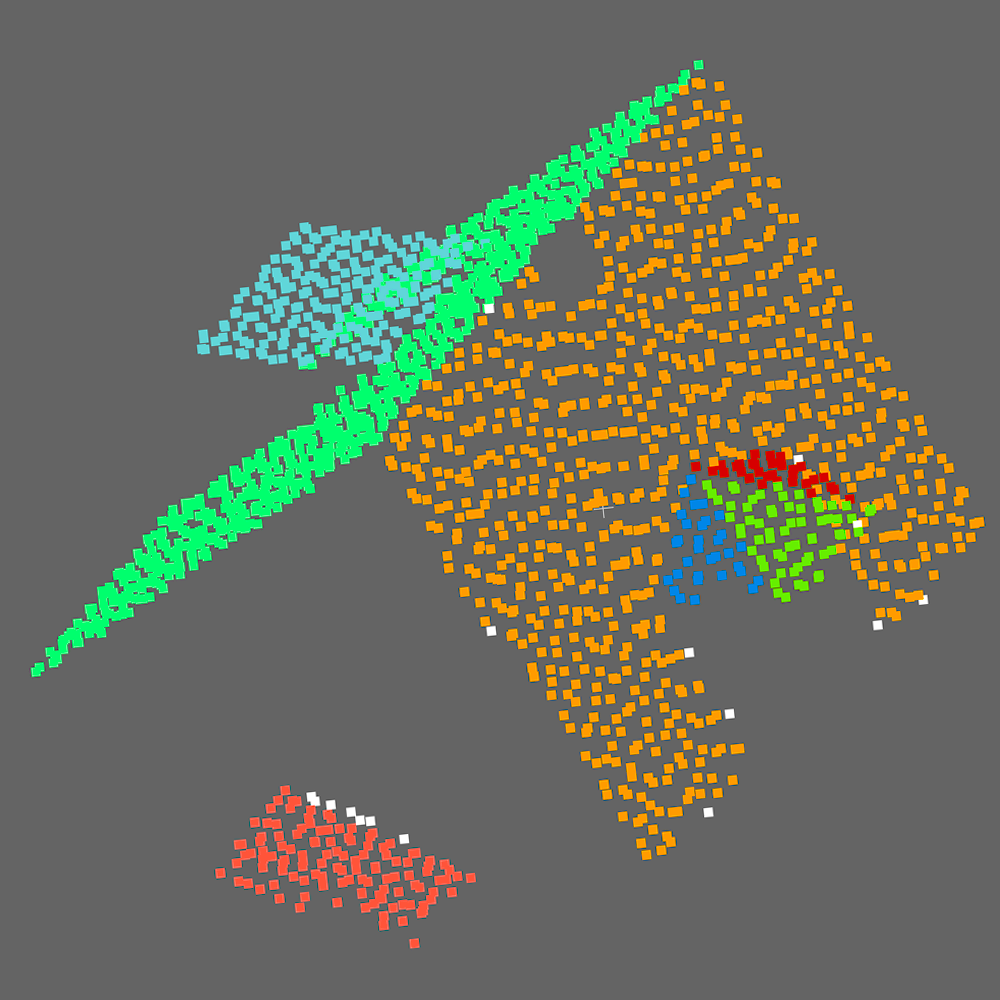} &
            \includegraphics[width=2.5cm]{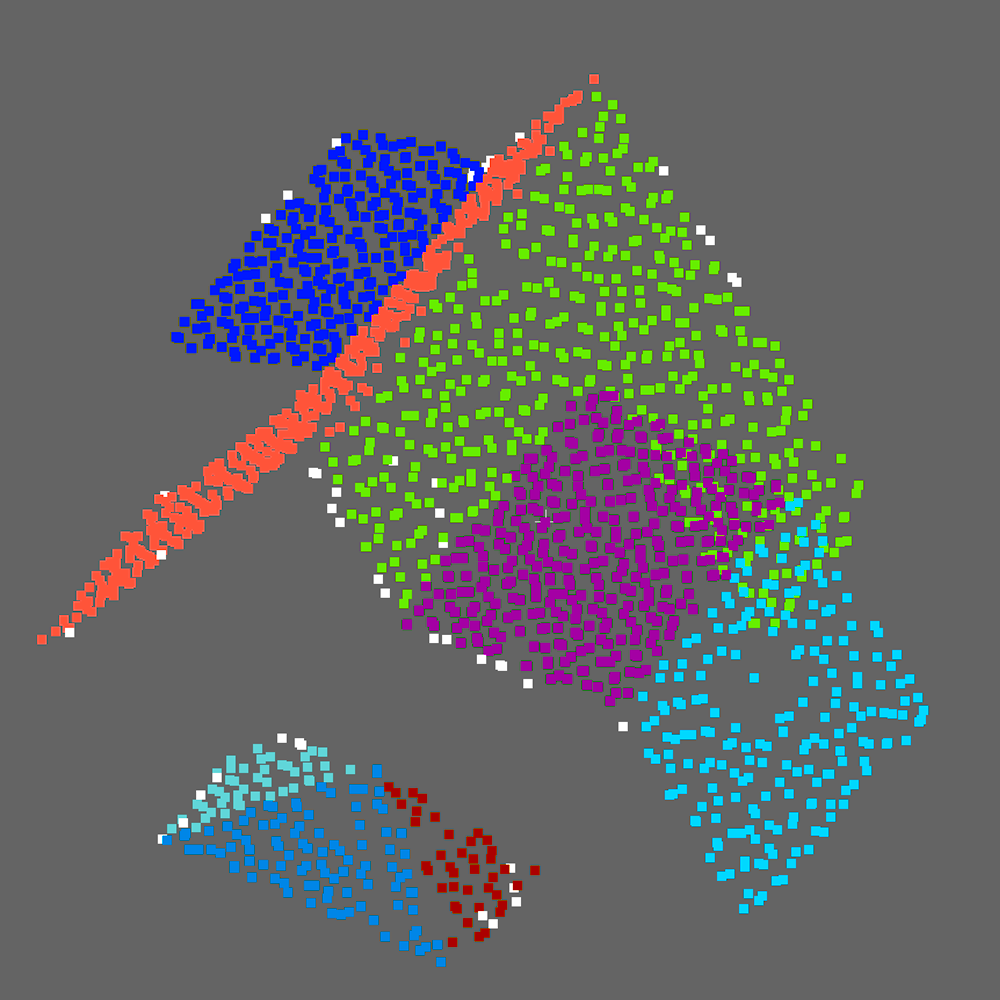}  \\
        \end{tabular}
    }
   \caption{
        Sample point clouds from the two datasets used in our experiments. 
        PC-1 and PC-2 are from the original RoofN3D dataset, while PC-3 to PC-5 are from the Building3D dataset. 
        Notable variations in point cloud density are observed in some samples,  particularly in PC-1 and PC-2.
    }
    \label{Fig:5}
\end{figure*}

Each of the two datasets presents unique challenges. 
The Building3D dataset features relatively high and uniform point cloud density, precise 3D points, and high-quality annotations, but it covers a wide variety of complex roof types. 
In contrast, the RoofN3D dataset primarily contains three roof types: gable, pyramid, and hip. However, the point cloud density is low and highly variable, with low 3D point precision, making it highly challenging to process. 
Additionally, as discussed in Section~\ref{sec:plane_completion}, this dataset suffers from incomplete plane instance annotations. 
In the experiments, we will observe that the original RoofN3D dataset is actually more difficult to handle. 
Fig.~\ref{Fig:5} provides some samples from both datasets.

\subsection{Evaluation metrics}
\label{sec:metric}

We use a range of evaluation metrics to comprehensively assess our model's performance. 
These metrics include coverage (Cov), weighted coverage (WCov)~\cite{Wang_CVPR_2019}, precision, recall, F1 score, and accuracy. 
Together, these metrics provide a holistic evaluation framework for our model. 
While Cov and WCov are primarily used for quantitative assessment, the other metrics offer a more detailed analysis of the model’s effectiveness in segmentation tasks.
Note that all quantitative improvements reported in this paper refer to the absolute increases in evaluation metrics, rather than relative gains.

\subsection{Implementation details}
\label{sec:Implementation_details}
Our SPPSFormer was trained using the AdamW optimizer with an initial learning rate of 0.0001, a batch size of 4, and a weight decay factor of 0.05. 
The PolyLR learning rate scheduling strategy was applied, with a decay power of 0.9. 
To ensure reproducibility, the random seed was fixed at 200 for all experiments. 
Training was performed on an NVIDIA Tesla V100 GPU, with 100 epochs on the RoofN3D dataset and 200 epochs on the Building3D dataset.
In all experiments, we set $\lambda$ to 20 in our boundary refinement module.

\subsection{Comparative experiments}
\label{sec:Comparative_experiments}

We compared the proposed SPPSFormer (including our two postprocessing steps) with traditional roof plane segmentation algorithms---including the hierarchical clustering and boundary relabeling (HCBR) algorithm \cite{Li_RS_2020} and the quasi-a-contrario theory-based plane segmentation (QTPS) algorithm \cite{Zhu_STAEORS_2021}, the superpoint Transformer-based 3D point cloud instance segmentation network SPFormer \cite{Sun_AAAI_2023}, and the  latest deep learning network for segmenting plane instances from 3D point clouds, DeepRoofPlane \cite{Li_ISPRS_2024}.

Table~\ref{Tab:1} presents detailed quantitative evaluation results using the original RoofN3D dataset.
Due to low point cloud density, significant density variations, and low point precision in this dataset, plane instance segmentation is theoretically highly challenging. 
As shown in Table~\ref{Tab:1}, our proposed SPPSFormer achieves excellent performance across all evaluation metrics, particularly excelling in the Cov metric.
Compared to the second-best method, DeepRoofPlane \cite{Li_ISPRS_2024}, our SPPSFormer improves Cov and WCov by 3.03\% and 1.85\%, respectively. 
These results indicate that our method offers high accuracy and robustness in plane instance extraction tasks. 
It is worth noting that the QTPS algorithm was only able to process 392 out of the 1,000 test samples, and the evaluation results for QTPS are based solely on these 392 samples. 
The qualitative comparison results, shown in Fig.~\ref{Fig:6}, further validate the above conclusions. 
It should also be noted that in Table~\ref{Tab:1}, the best-performing algorithms are highlighted in bold, a formatting choice that is consistently applied to the subsequent tables throughout the paper.

\begin{table} [H] \renewcommand{\tabcolsep}{4.0 pt}
    \scriptsize
    \renewcommand{\arraystretch}{1.5}
    \begin{center}
        \caption{Quantitative evaluation results on the RoofN3D dataset.}
        \label{Tab:1}
        \begin{tabular}{c|cccccc} 
            \hline
            \multirow{2}*{Different approaches} &   \multicolumn{6}{c}{Original RoofN3D test set}  \\ 
            \cline{2-7}  
            & Cov & WCov & Precision & Recall & F1 score & Accuracy \\
            \hline 
            SPFormer & 0.6525 & 0.7023 & 0.7864 & 0.4992 & 0.6044 & 0.4966 \\ 
            QTPS & 0.6304 & 0.6579 & 0.6962 & 0.6114 & 0.6512 & 0.6836 \\
            HCBR & 0.6755 & 0.7128 & 0.7017 & 0.9630 & 0.7905 & 0.7916 \\ 
            DeepRoofPlane & 0.9020 & 0.9213 & 0.8551 & \textbf{0.9965} & 0.9171 & 0.9513 \\
            Our SPPSFormer & \textbf{0.9323} & \textbf{0.9398} & \textbf{0.8724} & 0.9870 & \textbf{0.9238} & \textbf{0.9564} \\
            SPPSFormer nano & 0.9044 & 0.9152 & 0.8626 & 0.9812 & 0.9152 & 0.9386 \\
            \hline 
        \end{tabular}
    \end{center}
\end{table}

\begin{figure*}[!t]
    \centering
    \renewcommand{\arraystretch}{1.2}  
    \begin{tabularx}{\textwidth}{@{}c@{\hspace{0.00cm}}c@{\hspace{0.00cm}}c@{\hspace{0.00cm}}c@{\hspace{0.00cm}}c@{\hspace{0.00cm}}c@{\hspace{0.00cm}}c@{\hspace{0.00cm}}c@{}}
        & \makecell{\footnotesize \textbf{Ground-truth} \\ (Side-view)} &
        \makecell{\footnotesize \textbf{Ground-truth} \\ (Top-view)} &
        \makecell{\footnotesize \textbf{SPFormer}} &
        \makecell{\footnotesize \textbf{HCBR}} &
        \makecell{\footnotesize \textbf{QTPS}} &
        \makecell{\footnotesize \textbf{DeepRoofPlane}} &
        \makecell{\footnotesize \textbf{Our Approach}} \\
        
        \multirow{6}{*}[6.8em]{\rotatebox[origin=c]{90}{\makecell{\textbf{PC-1}}}} &
        \includegraphics[width=2.5cm]{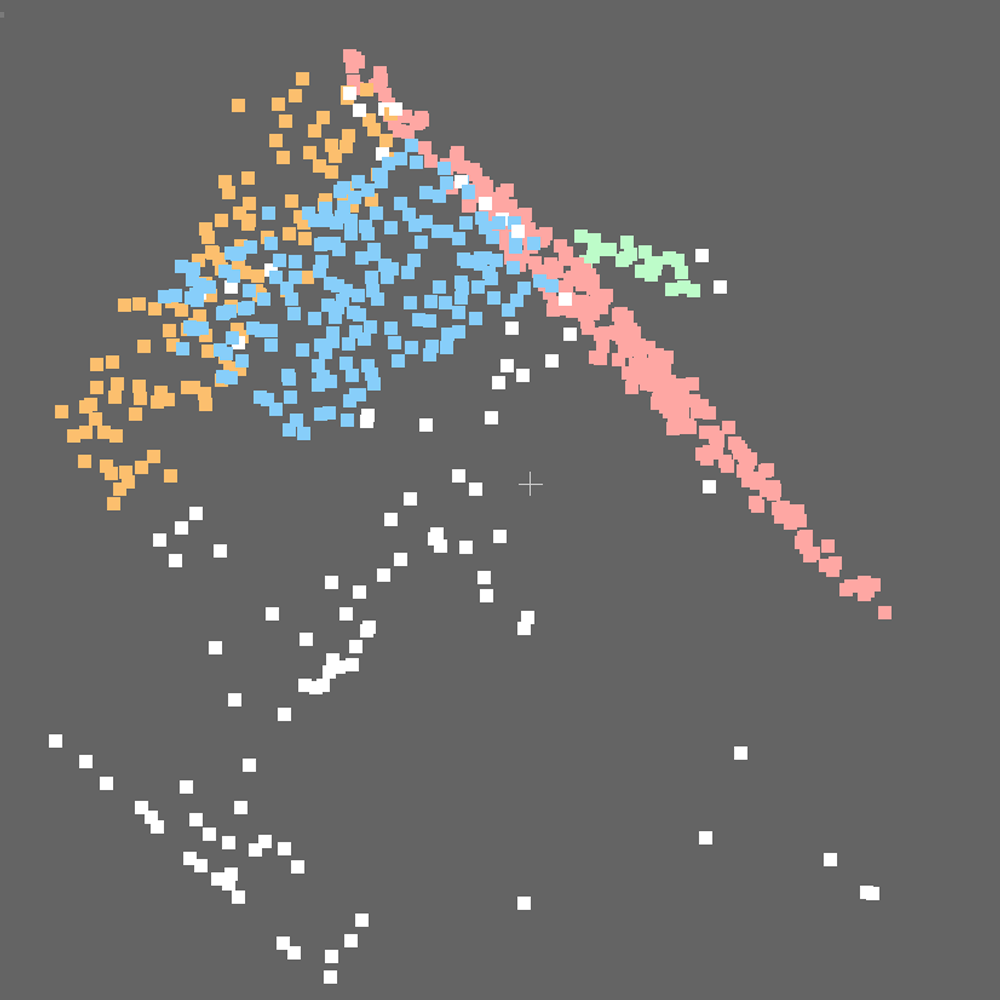} &
        \includegraphics[width=2.5cm]{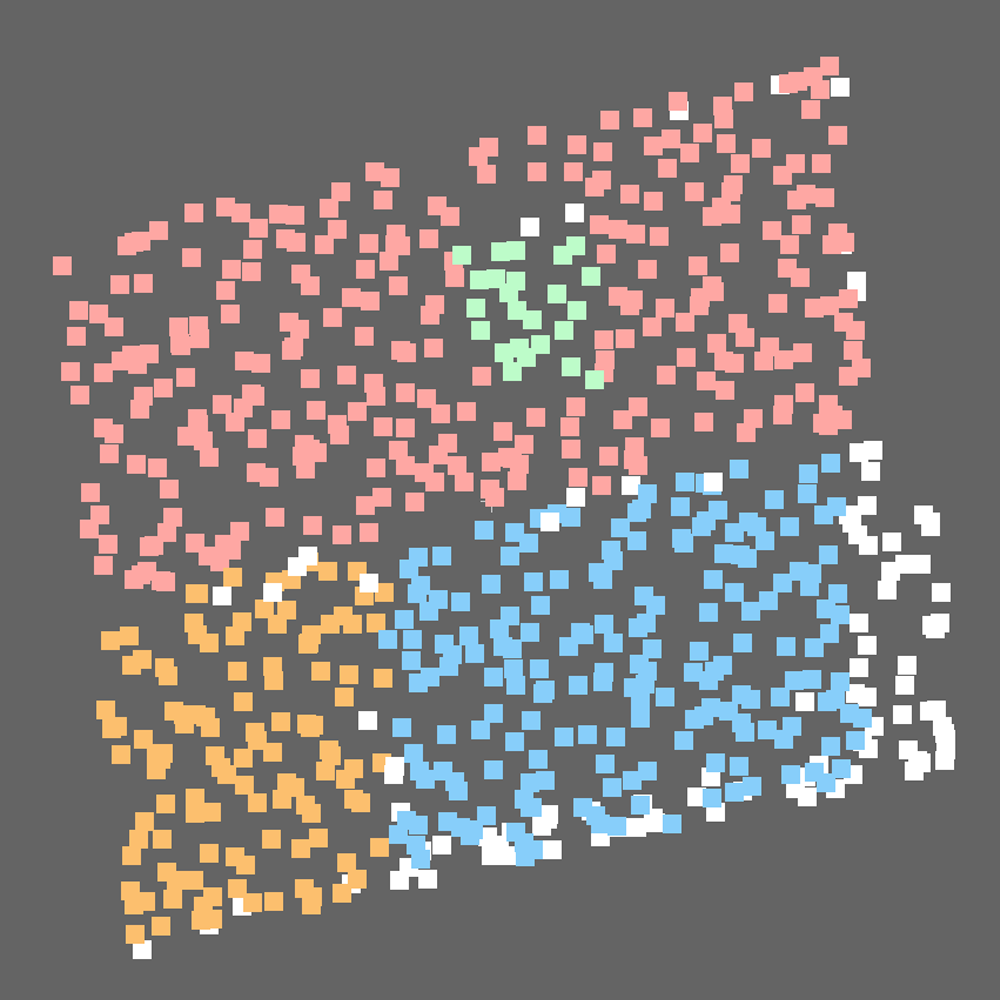} &
        \includegraphics[width=2.5cm]{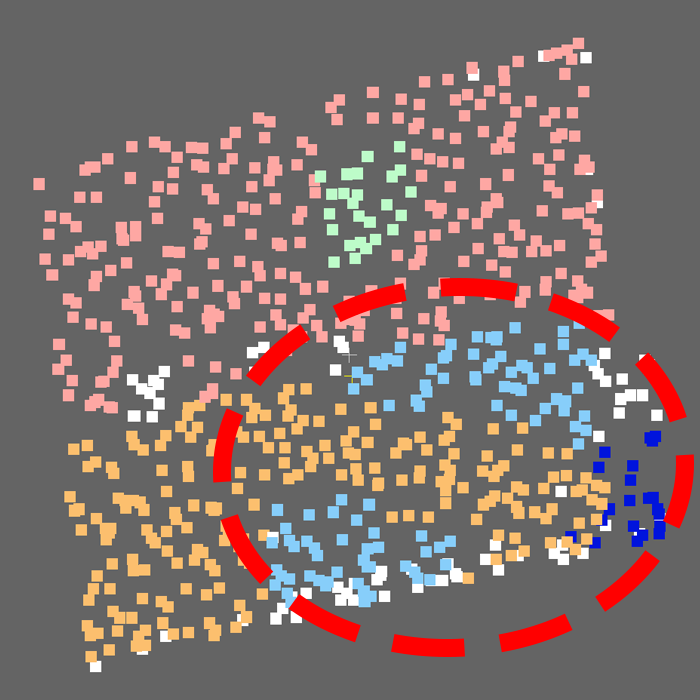} &
        \includegraphics[width=2.5cm]{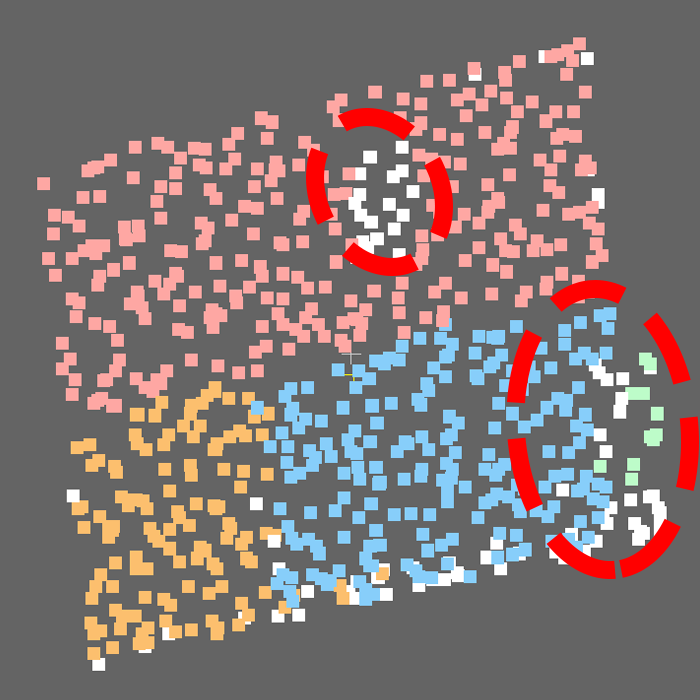} &
        \includegraphics[width=2.5cm]{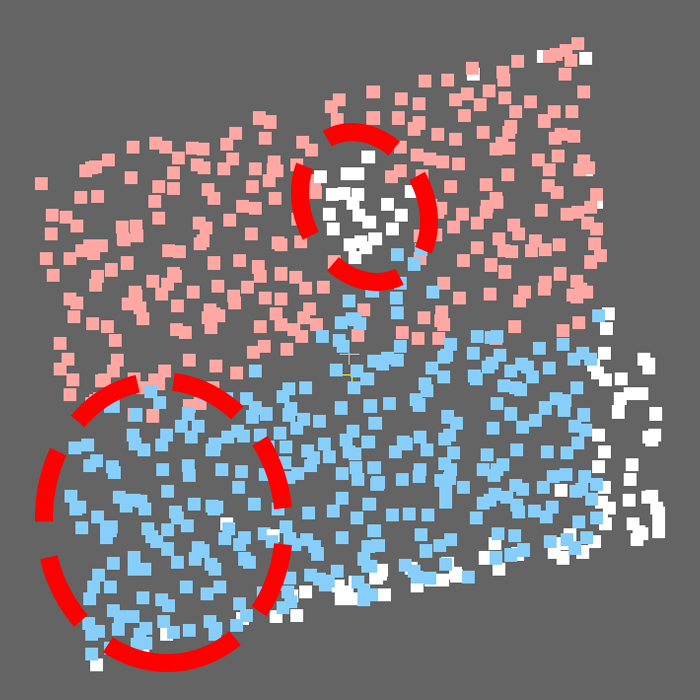} &
        \includegraphics[width=2.5cm]{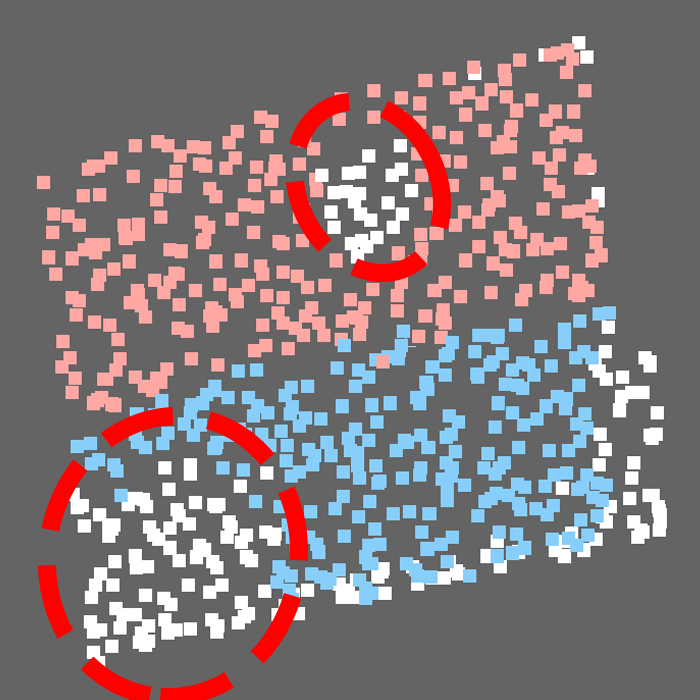} &
        \includegraphics[width=2.5cm]{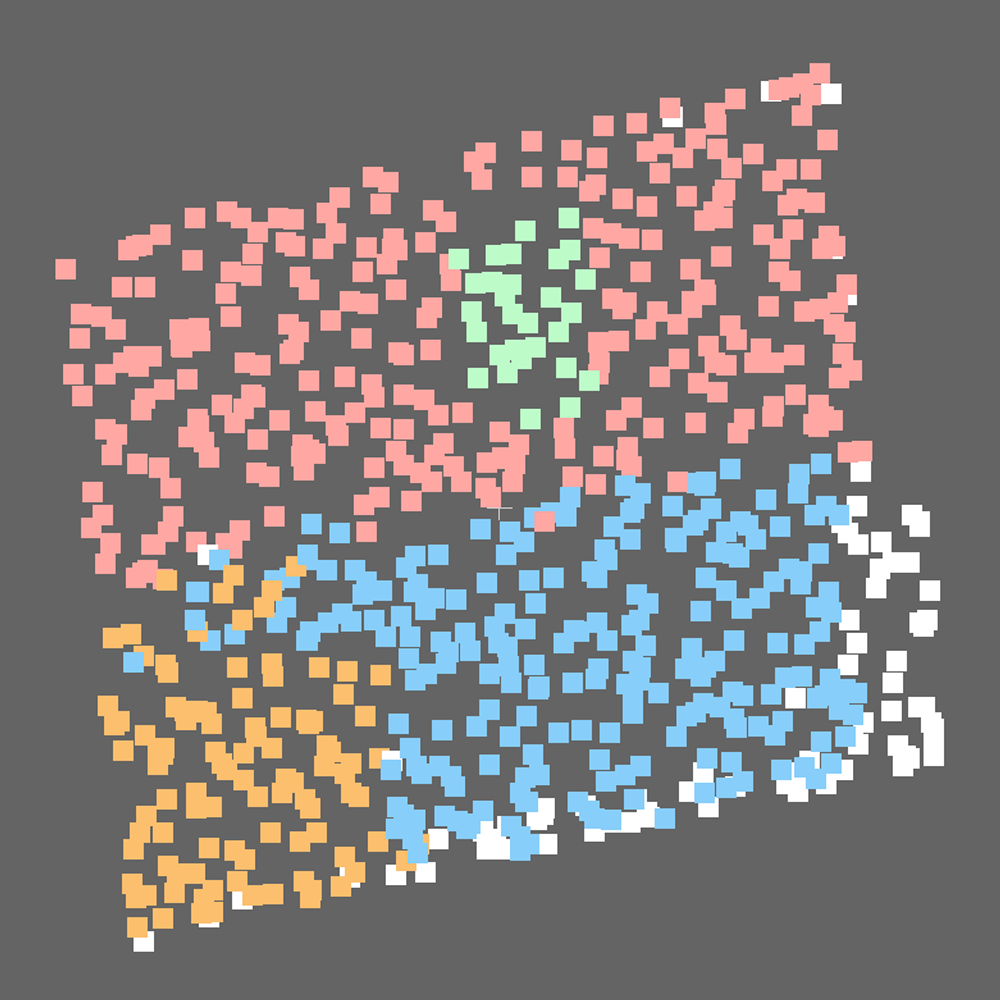} \\
        
        \multirow{6}{*}[6.8em]{\rotatebox[origin=c]{90}{\makecell{\textbf{PC-2}}}} &
        \includegraphics[width=2.5cm]{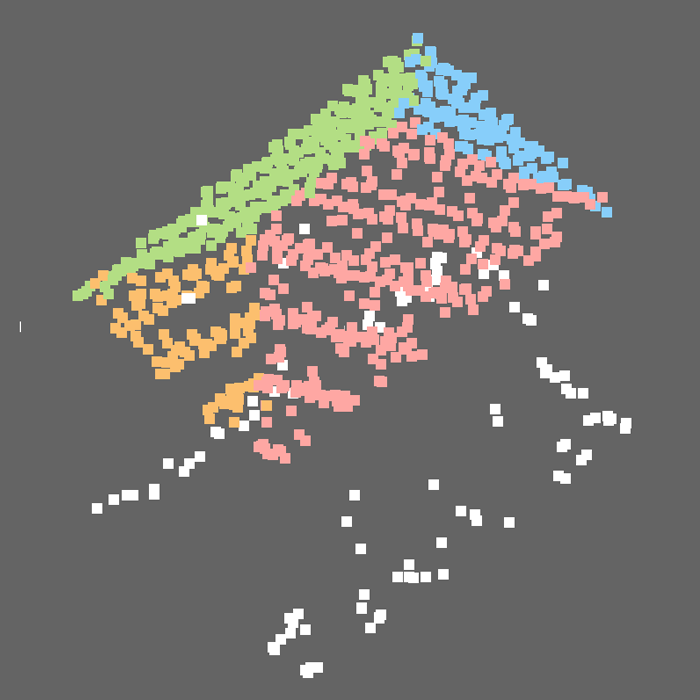} &
        \includegraphics[width=2.5cm]{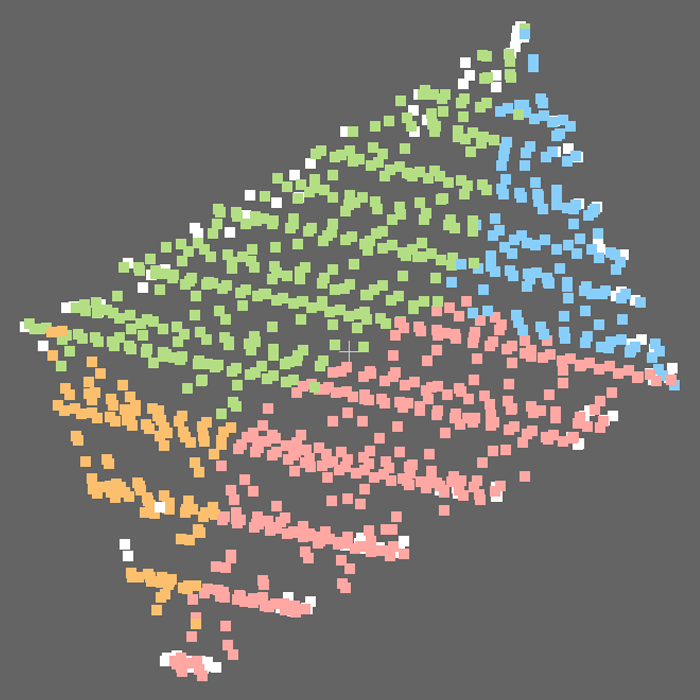} &
        \includegraphics[width=2.5cm]{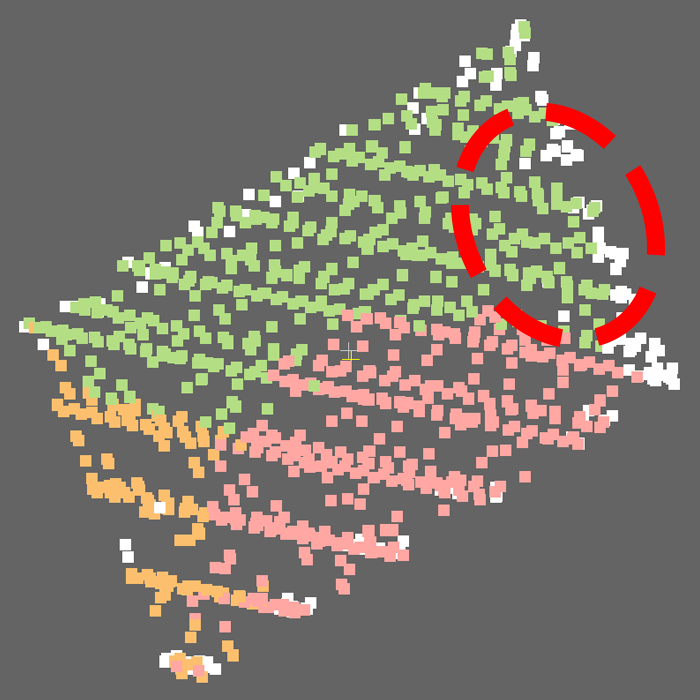} &
        \includegraphics[width=2.5cm]{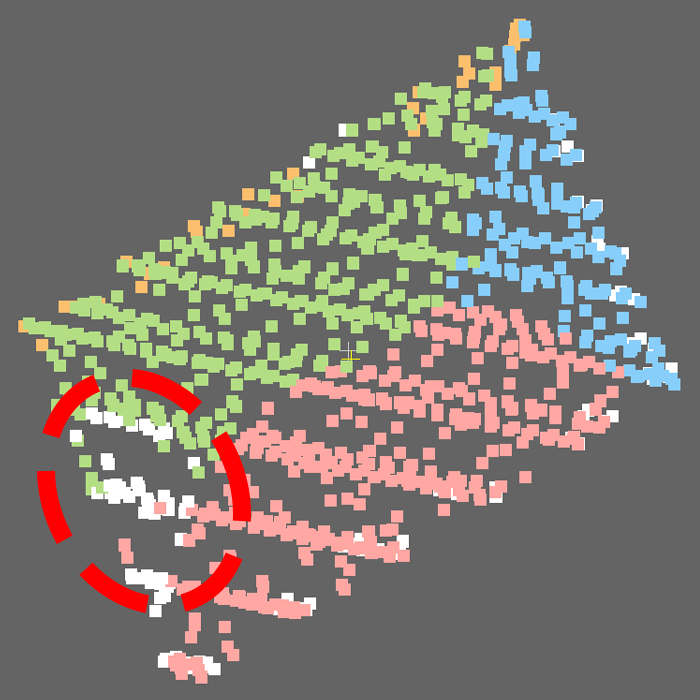} &
        \includegraphics[width=2.5cm]{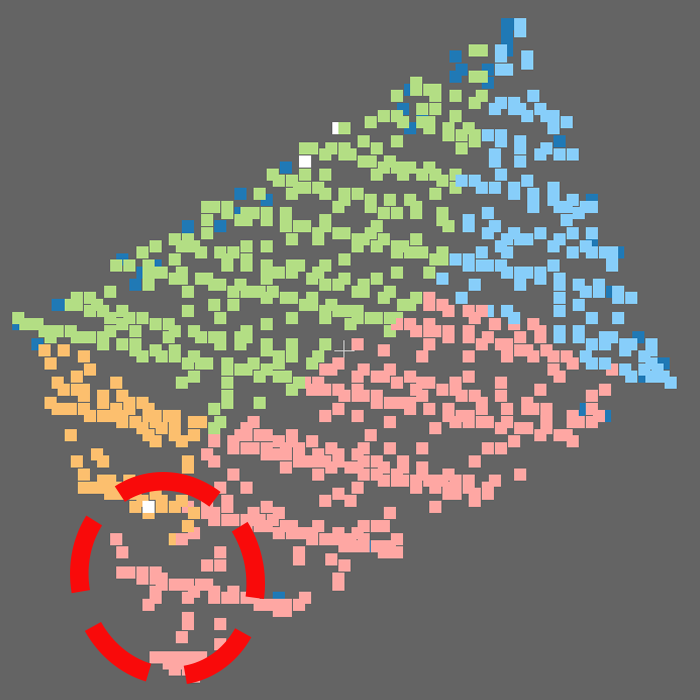} &
        \includegraphics[width=2.5cm]{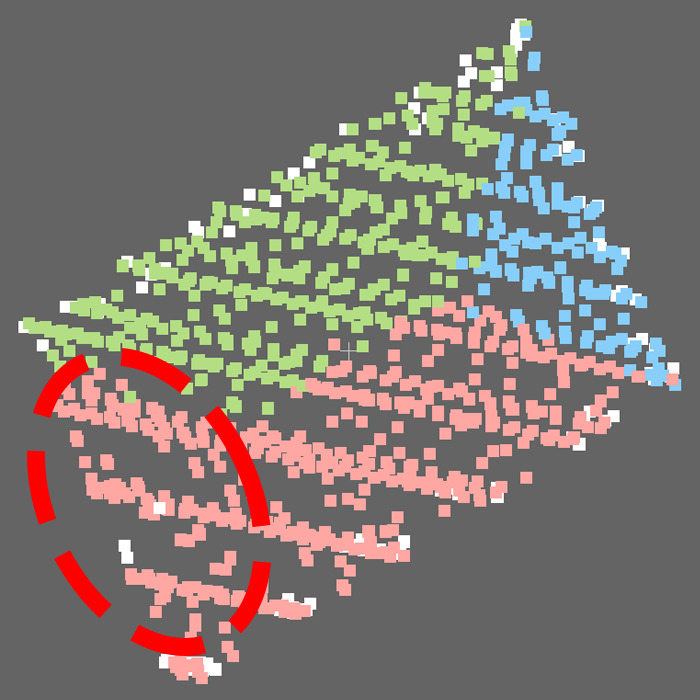} &
        \includegraphics[width=2.5cm]{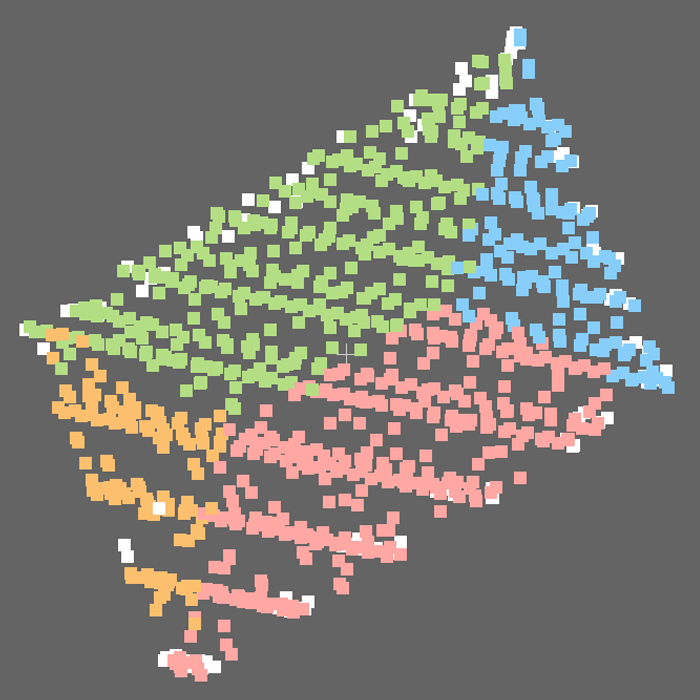} \\
        
        \multirow{6}{*}[6.8em]{\rotatebox[origin=c]{90}{\makecell{\textbf{PC-3}}}} &
        \includegraphics[width=2.5cm]{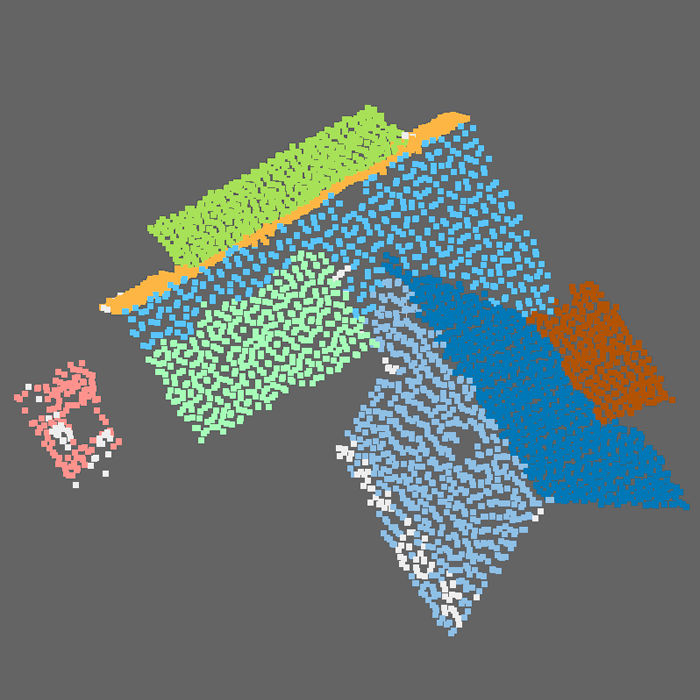} &
        \includegraphics[width=2.5cm]{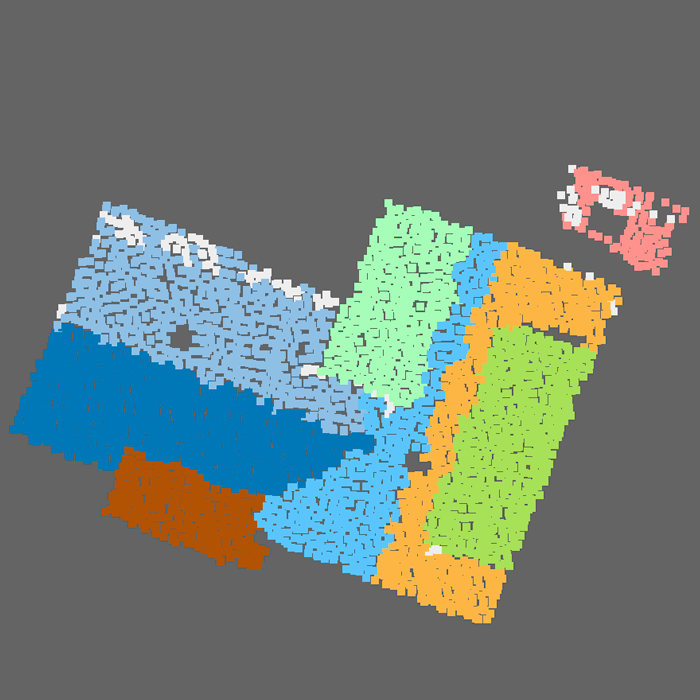} &
        \includegraphics[width=2.5cm]{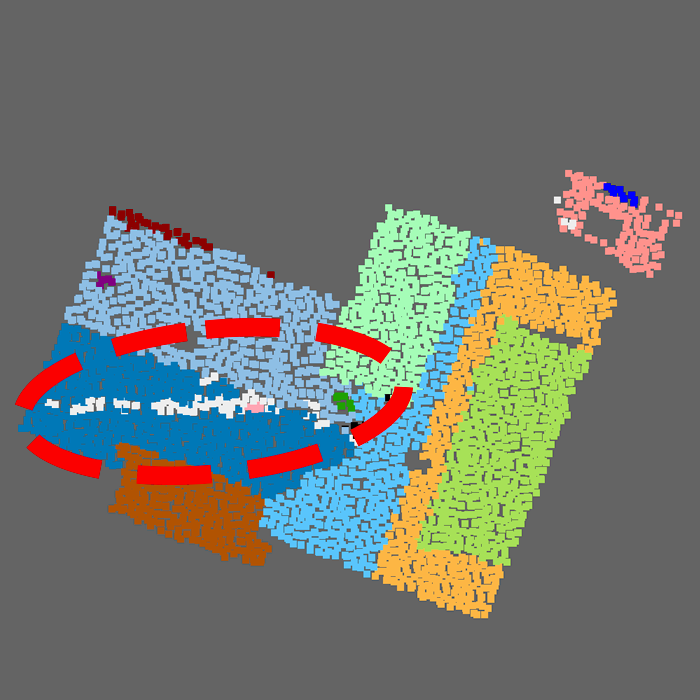} &
        \includegraphics[width=2.5cm]{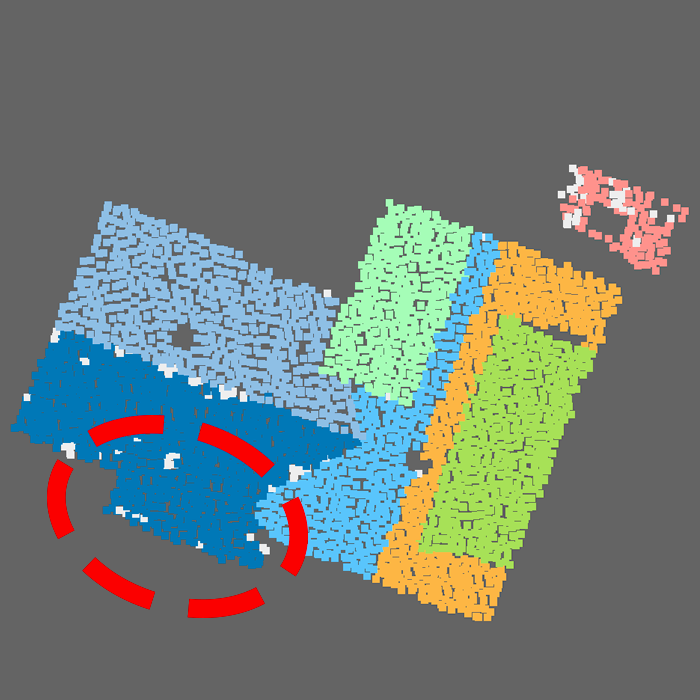} &
        \includegraphics[width=2.5cm]{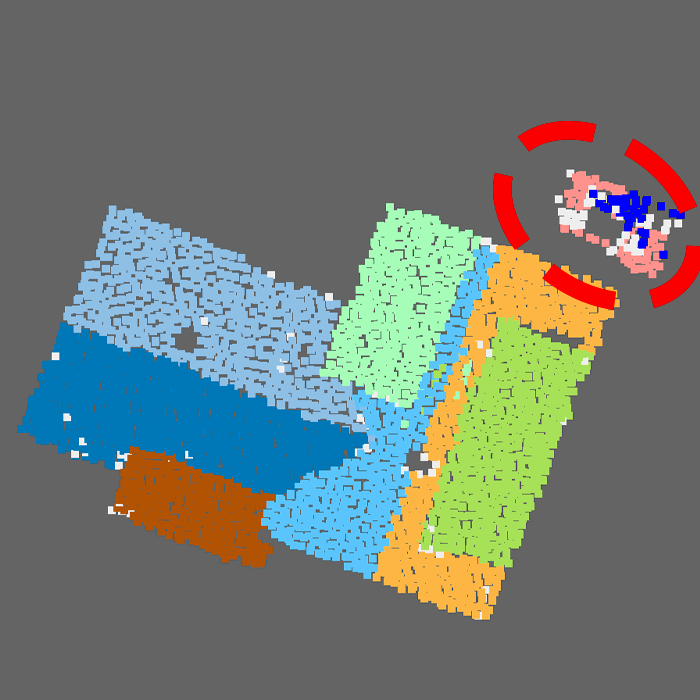} &
        \includegraphics[width=2.5cm]{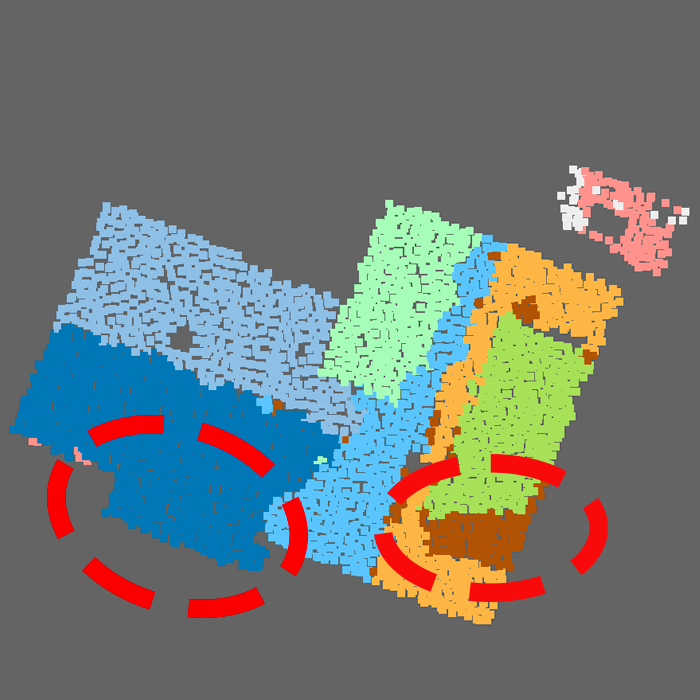} &
        \includegraphics[width=2.5cm]{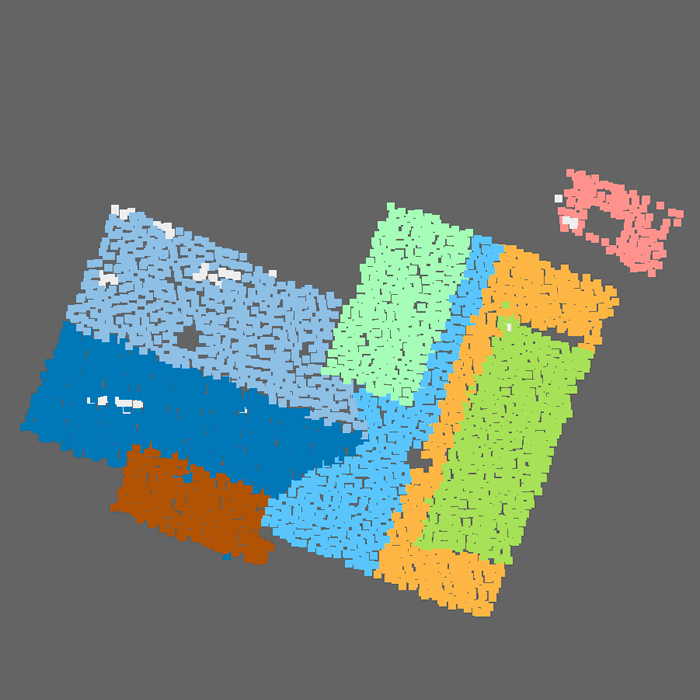} \\
        
        \multirow{6}{*}[6.8em]{\rotatebox[origin=c]{90}{\makecell{\textbf{PC-4}}}} &
        \includegraphics[width=2.5cm]{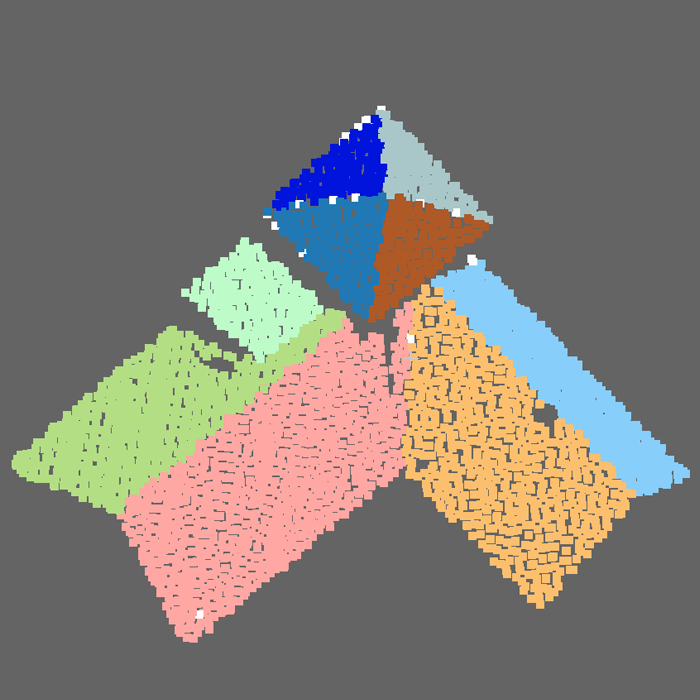} &
        \includegraphics[width=2.5cm]{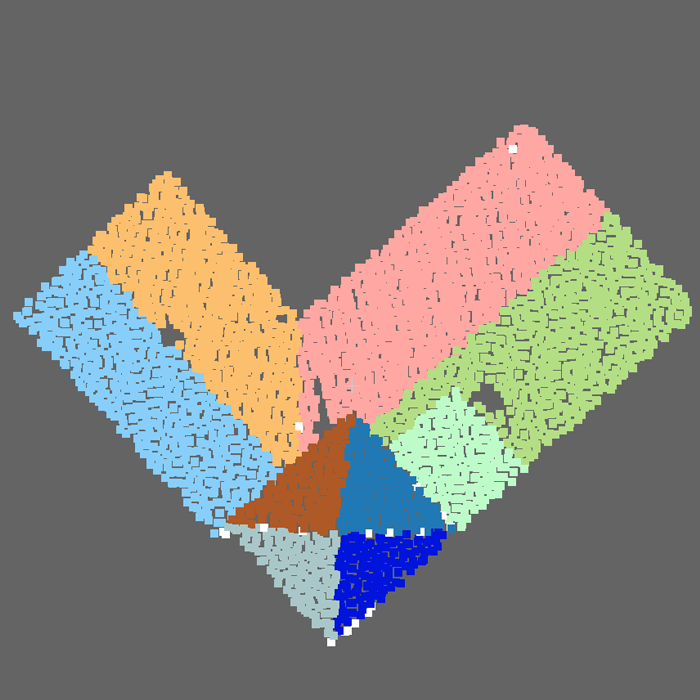} &
        \includegraphics[width=2.5cm]{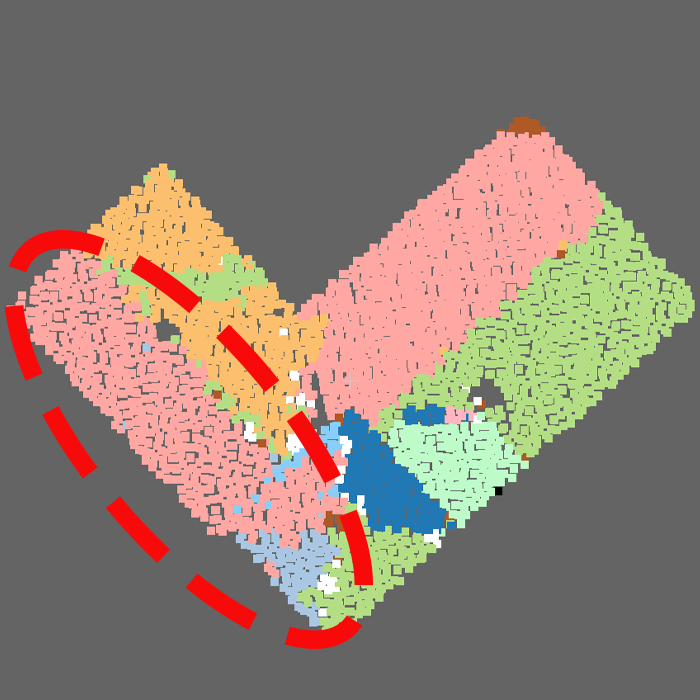} &
        \includegraphics[width=2.5cm]{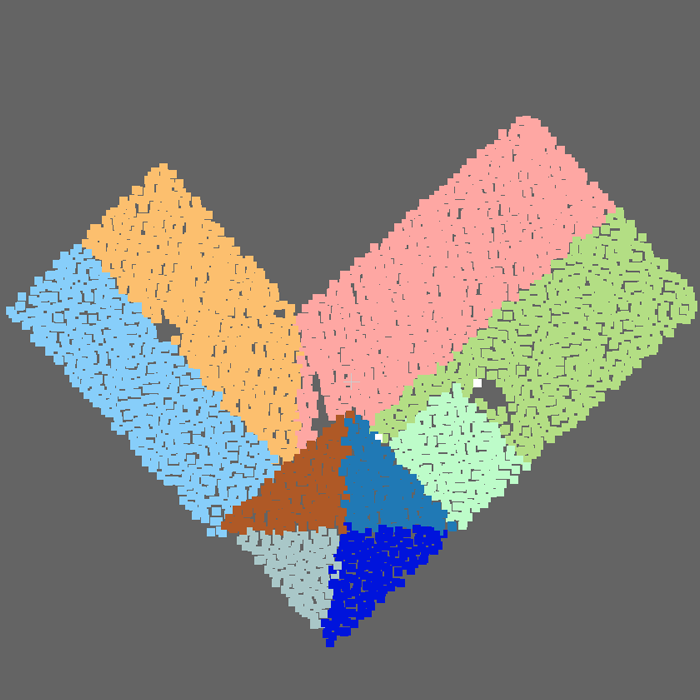} &
        \includegraphics[width=2.5cm]{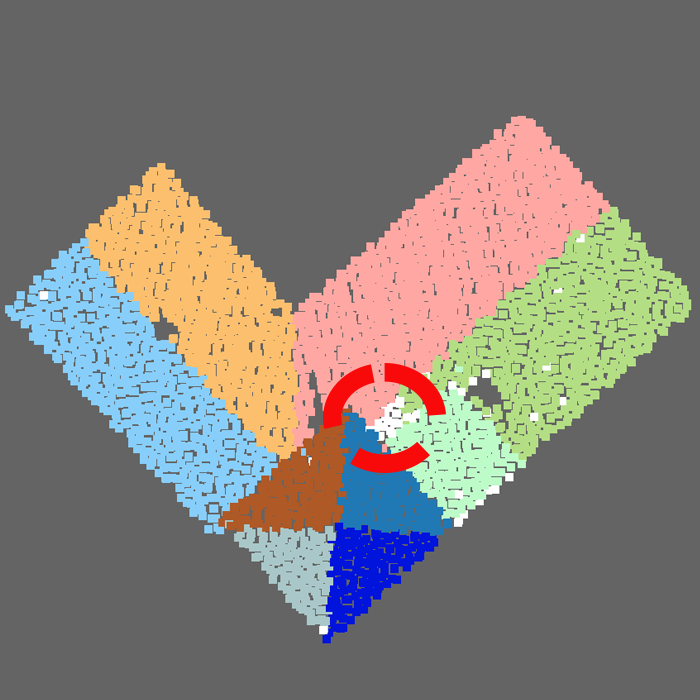} &
        \includegraphics[width=2.5cm]{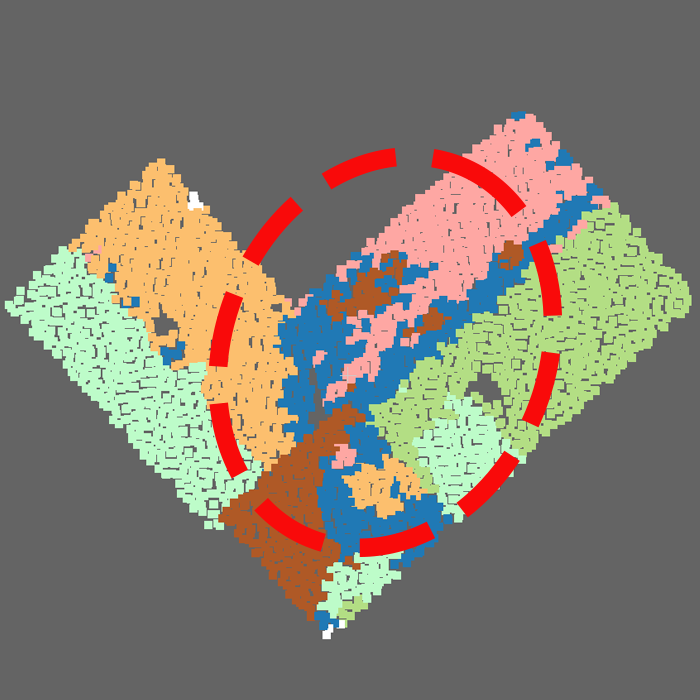} &
        \includegraphics[width=2.5cm]{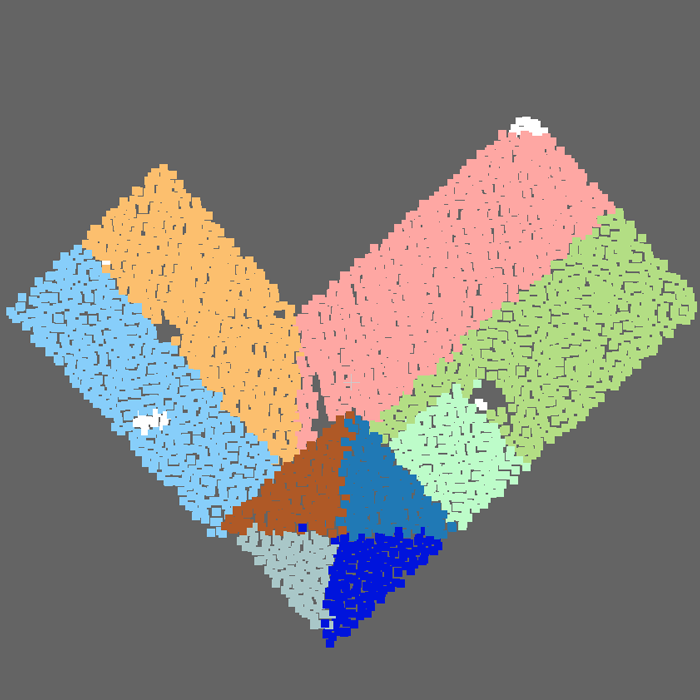} \\
        
        \multirow{6}{*}[6.8em]{\rotatebox[origin=c]{90}{\makecell{\textbf{PC-5}}}} &
        \includegraphics[width=2.5cm]{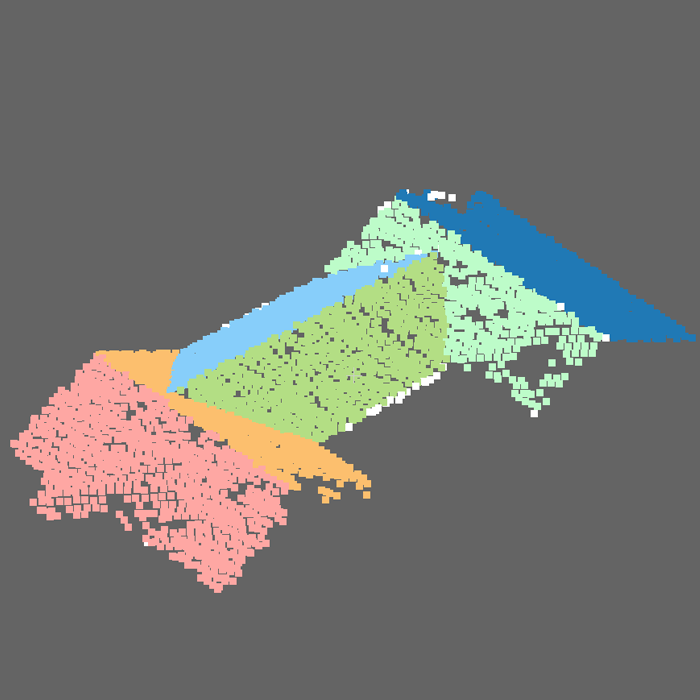} &
        \includegraphics[width=2.5cm]{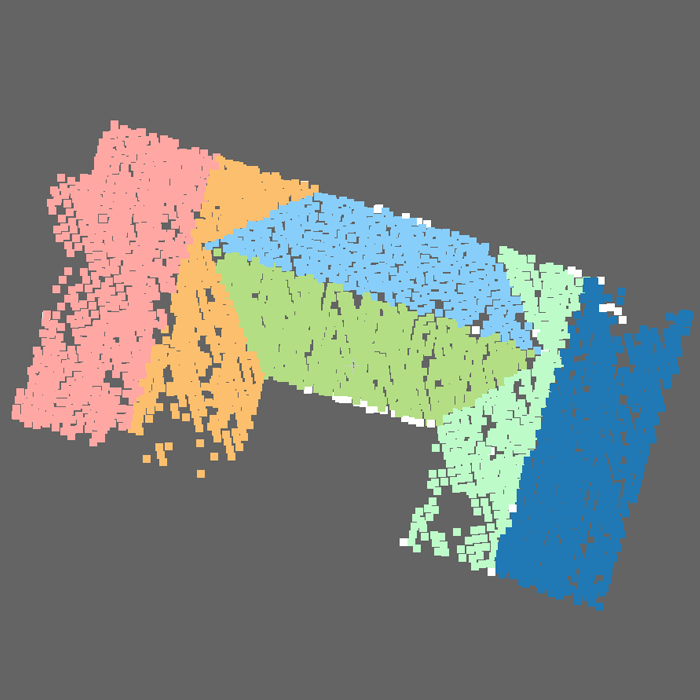} &
        \includegraphics[width=2.5cm]{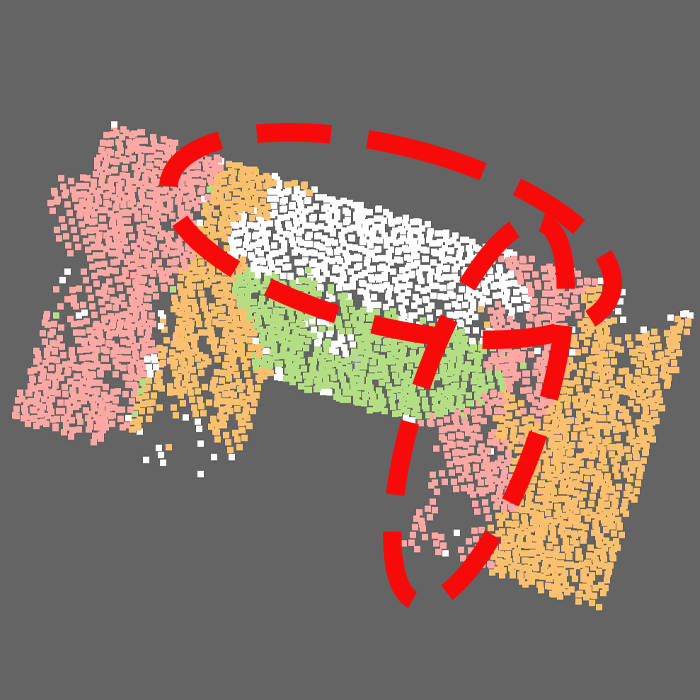} &
        \includegraphics[width=2.5cm]{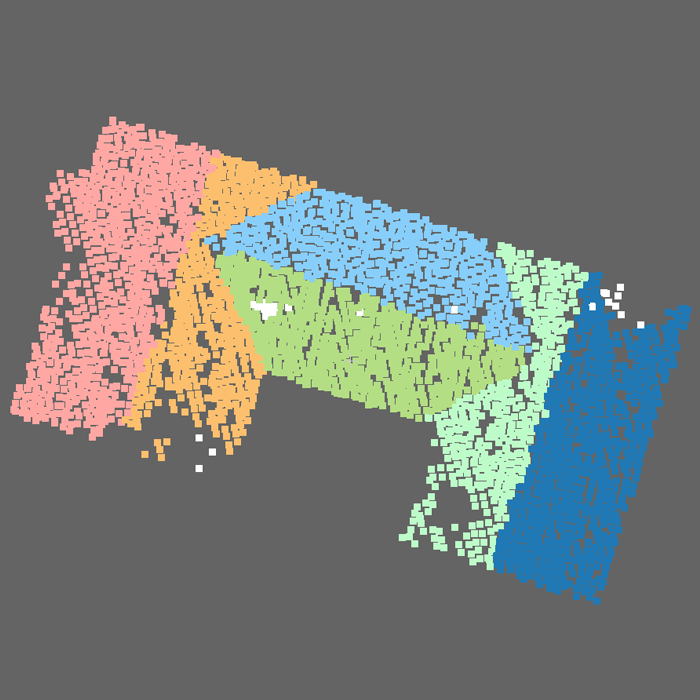} &
        \includegraphics[width=2.5cm]{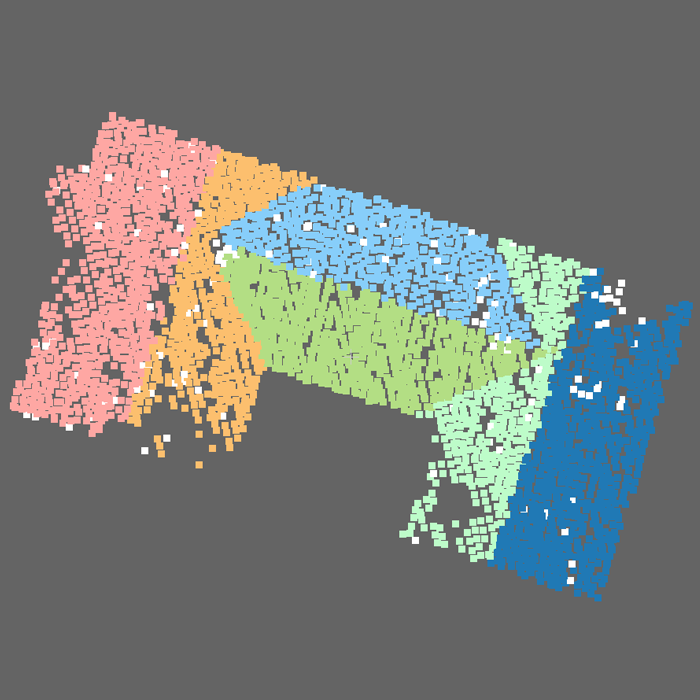} &
        \includegraphics[width=2.5cm]{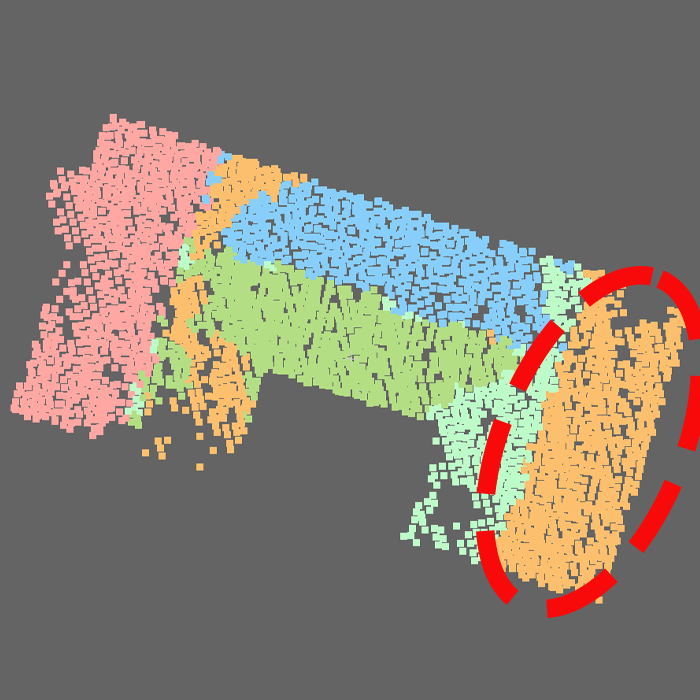} &
        \includegraphics[width=2.5cm]{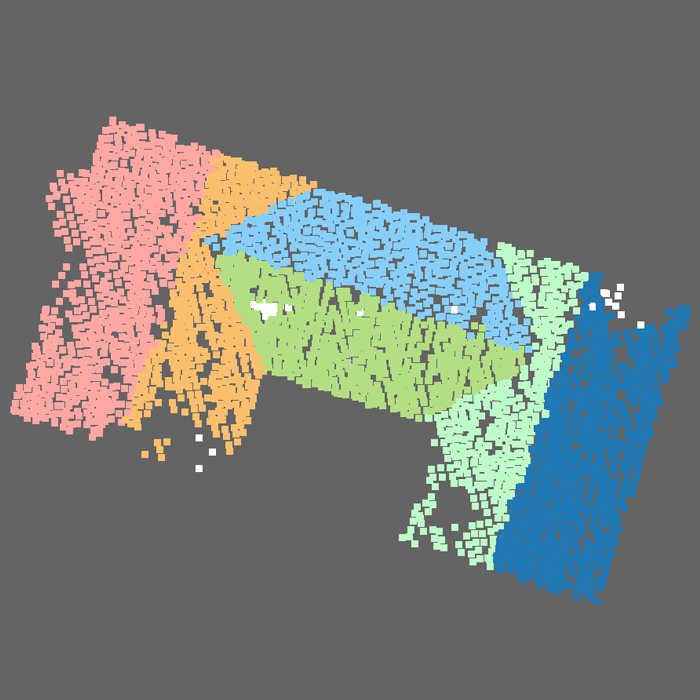} \\
        
        \multirow{6}{*}[6.8em]{\rotatebox[origin=c]{90}{\makecell{\textbf{PC-6}}}} &
        \includegraphics[width=2.5cm]{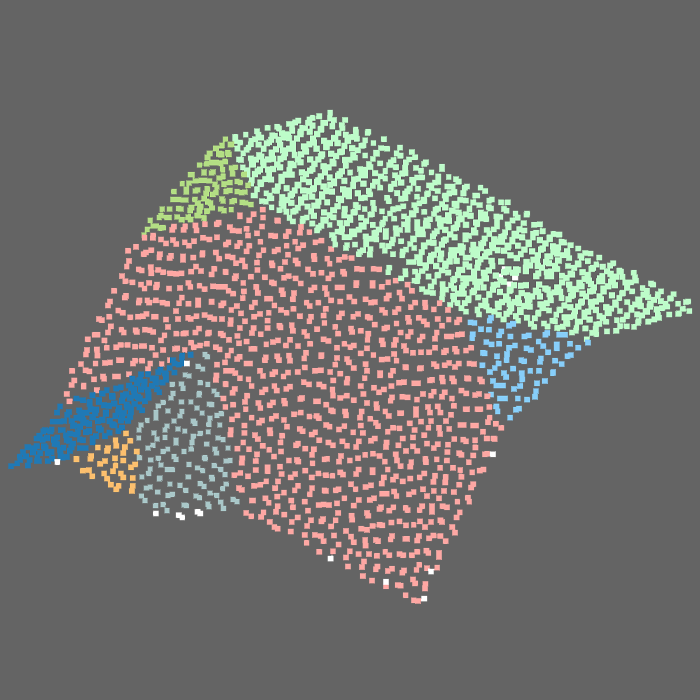} &
        \includegraphics[width=2.5cm]{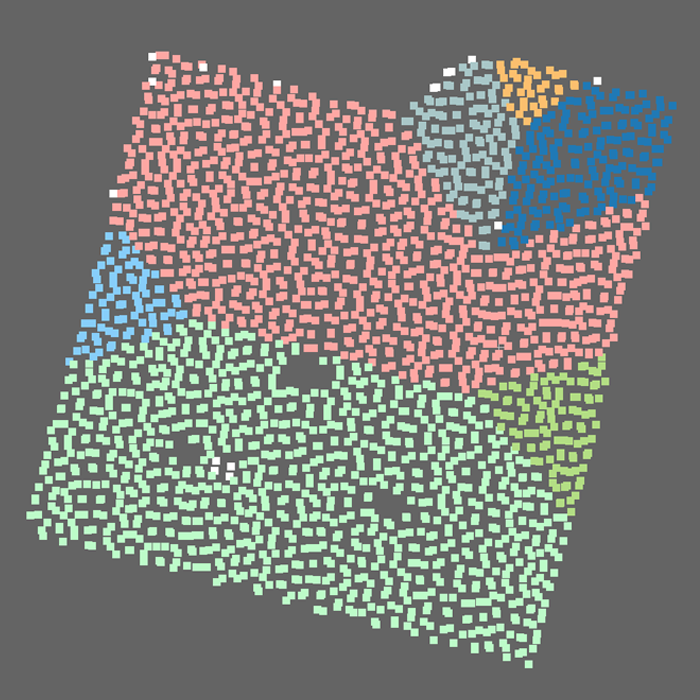} &
        \includegraphics[width=2.5cm]{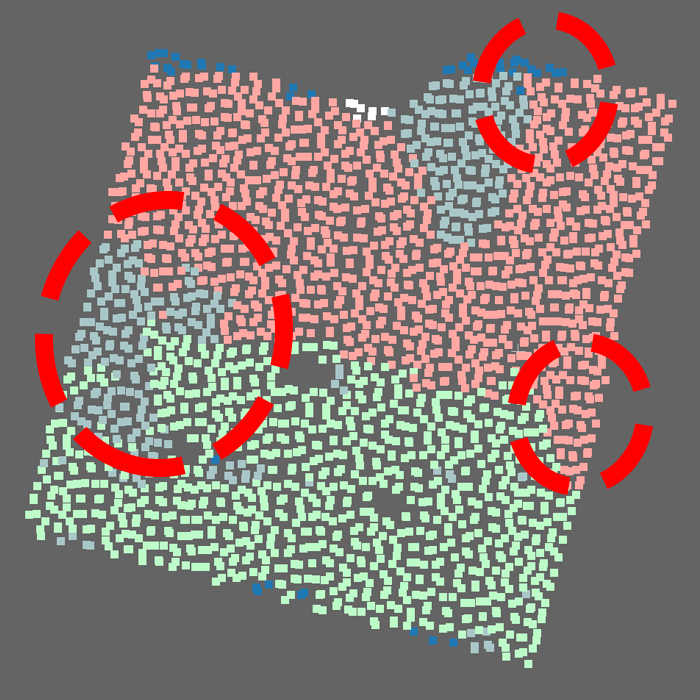} &
        \includegraphics[width=2.5cm]{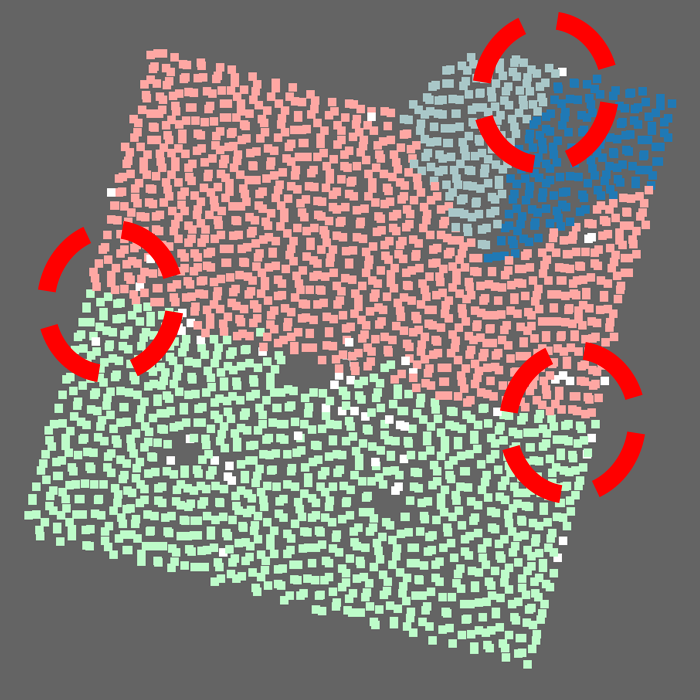} &
        \includegraphics[width=2.5cm]{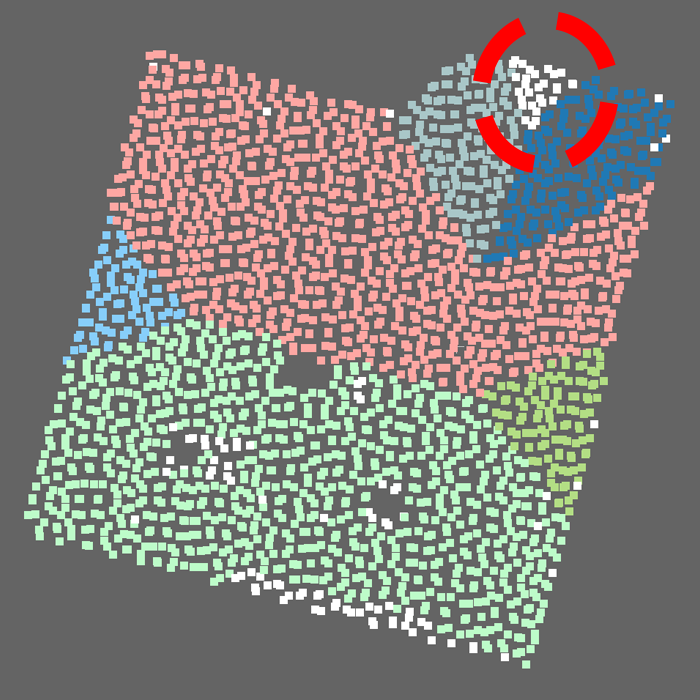} &
        \includegraphics[width=2.5cm]{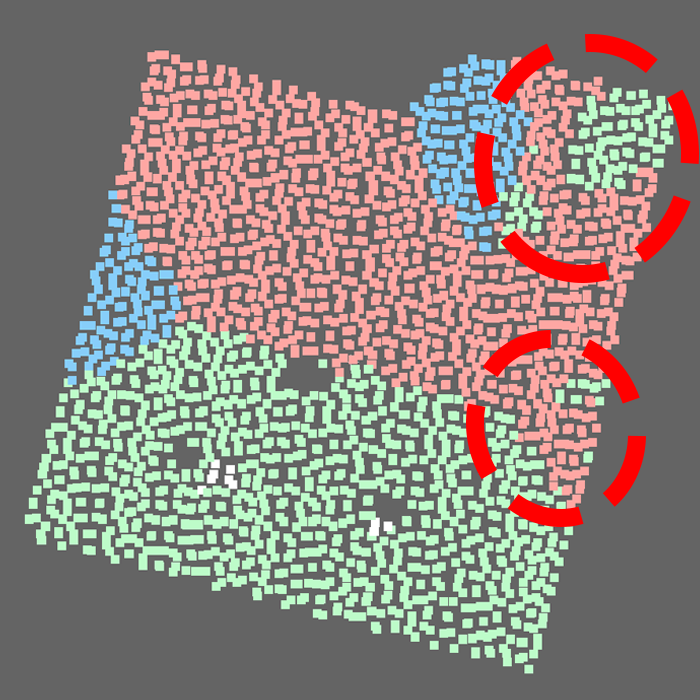} &
        \includegraphics[width=2.5cm]{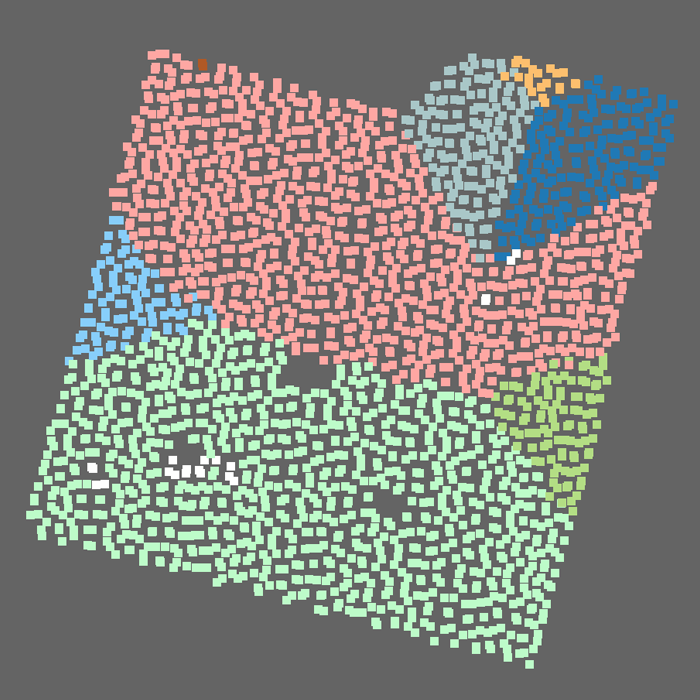} \\
    \end{tabularx}
    \caption{Visual comparison of segmentation results produced by different methods on the RoofN3D and Building3D datasets. 
        PC-1 and PC-2 are samples from RoofN3D, while PC-3 to PC-6 are from Building3D. 
        Nonplane points are shown in white, and different plane instances are indicated by distinct colors. 
        Red dotted lines highlight representative problem areas.}
    \label{Fig:6}
\end{figure*}

A key highlight of our work is the development of a lightweight model architecture, SPPSFormer nano, achieved by reducing the number of decoder layers, their dimensions, and the size of hidden layers. 
This optimization reduces the model complexity by 90\%, bringing the number of parameters down to just 1.8M. 
Despite this significant reduction in complexity, SPPSFormer nano still performs exceptionally well, as shown in Table~\ref{Tab:1}. 
This demonstrates the model’s ability to maintain strong performance while drastically reducing the computational requirements. 
Such advancements open the door for deploying effective deep learning models in resource-constrained environments.

Table~\ref{Tab:2} presents the quantitative evaluation results on the Building3D dataset. 
As illustrated, our approach outperforms other methods, showing an improvement of 1.77\% in Cov and 1.17\% in WCov compared to the second-best method, DeepRoofPlane \cite{Li_ISPRS_2024}. 
Notably, the QTPS algorithm only worked on 901 out of 1,000 test samples, and the values in Table~\ref{Tab:2} are averaged over these 901 samples. 
The qualitative comparison results shown in Fig.~\ref{Fig:6} further validate the effectiveness of our approach. 
Additionally, SPPSFormer nano also demonstrates strong performance on the Building3D dataset.

\begin{table}[!htb]
    \renewcommand{\tabcolsep}{4.0 pt}
    \scriptsize
    \renewcommand{\arraystretch}{1.5}
    \begin{center}
        \caption{Quantitative evaluation results on the Building3D dataset.}
        \label{Tab:2}
        \begin{tabular}{c|cccccc} 
            \hline
            \multirow{2}{*}{Different approaches} & \multicolumn{6}{c}{Building3D test set} \\ 
            \cline{2-7}
            & Cov & WCov & Precision & Recall & F1 score & Accuracy \\
            \hline 
            SPFormer           & 0.7895 & 0.8359 & 0.8877          & 0.9846 & 0.9270 & 0.8938 \\
            QTPS               & 0.8593 & 0.8981 & 0.9139          & 0.9703 & 0.9337 & 0.9470 \\
            HCBR               & 0.8927 & 0.9263 & 0.9352          & 0.9893 & 0.9561 & 0.9349 \\
            DeepRoofPlane      & 0.8835 & 0.9230 & \textbf{0.9562} & 0.9499 & 0.9480 & 0.9214 \\
            Our SPPSFormer     & \textbf{0.9010} & \textbf{0.9347} & 0.9465 & \textbf{0.9860} & \textbf{0.9623} & \textbf{0.9417} \\
            SPPSFormer nano    & 0.8559 & 0.8960 & 0.9190          & 0.9770 & 0.9405 & 0.9059 \\
            \hline 
        \end{tabular}
    \end{center}
\end{table}

\begin{figure}[!h]
    \centering
    \begin{tabular}{@{\hspace{0em}}c@{\hspace{0.4em}}c@{\hspace{0.4em}}c@{\hspace{0em}}}
        \includegraphics[width=0.32\linewidth]{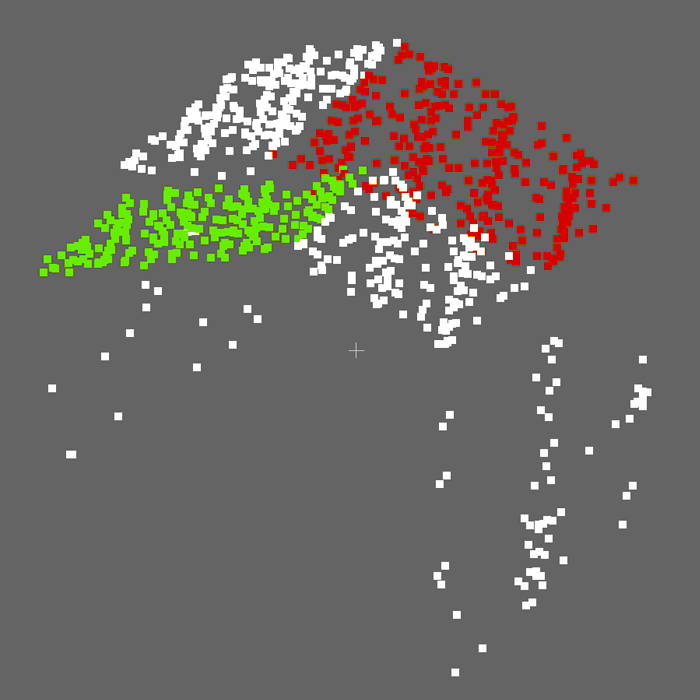} &
        \includegraphics[width=0.32\linewidth]{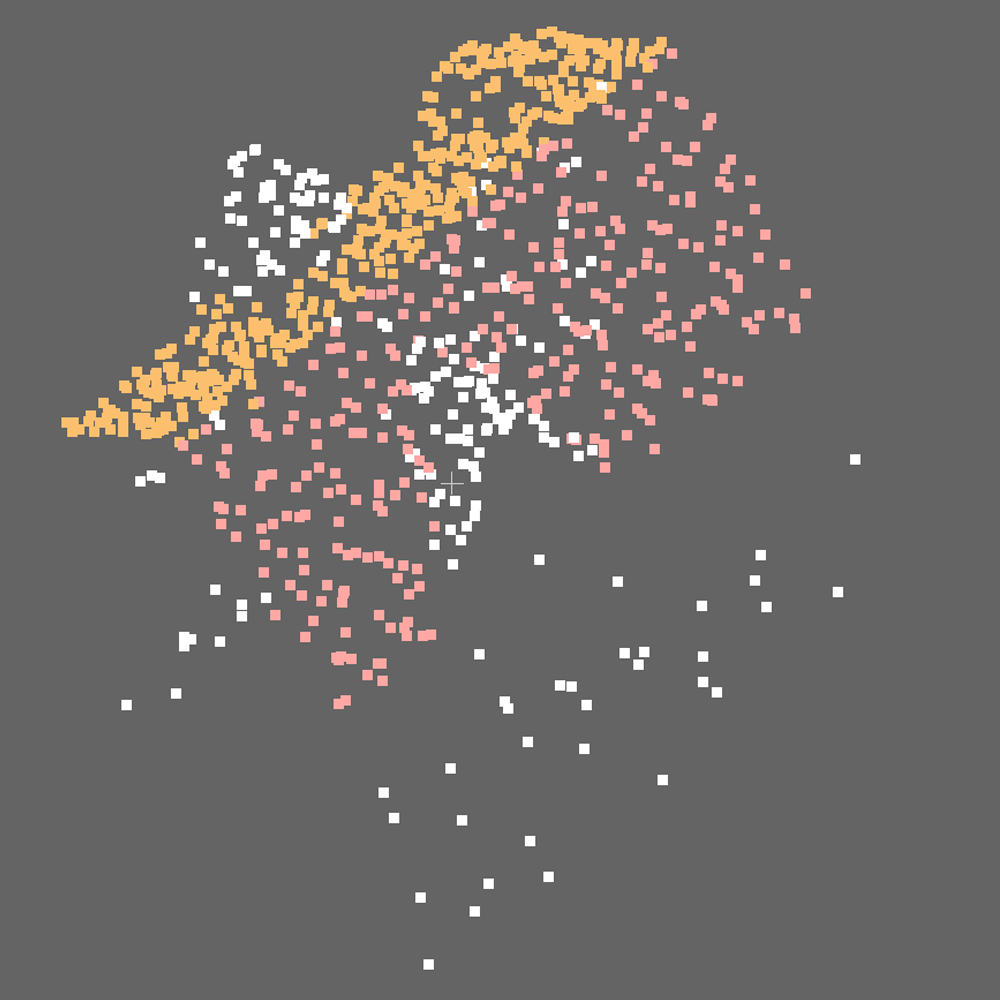} &
        \includegraphics[width=0.32\linewidth]{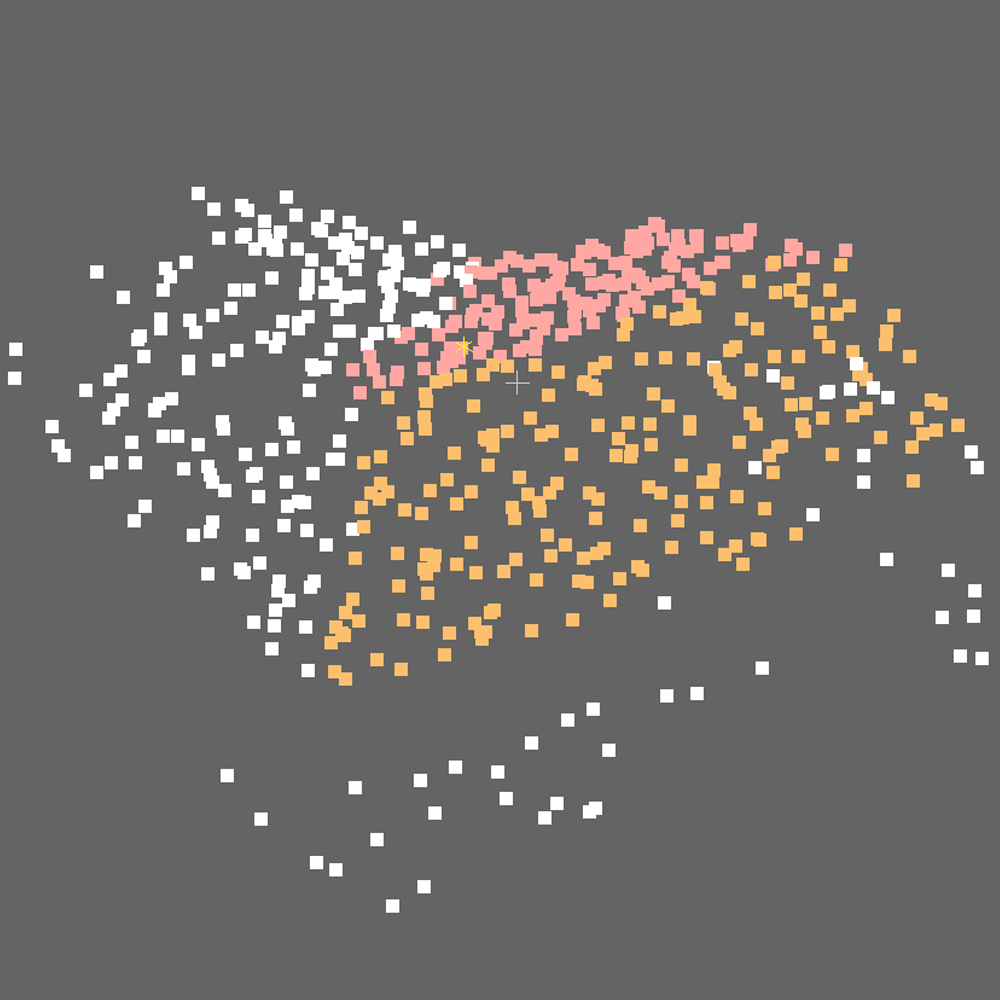} \\
        (a) & (b) & (c) \\[1ex]
        \includegraphics[width=0.32\linewidth]{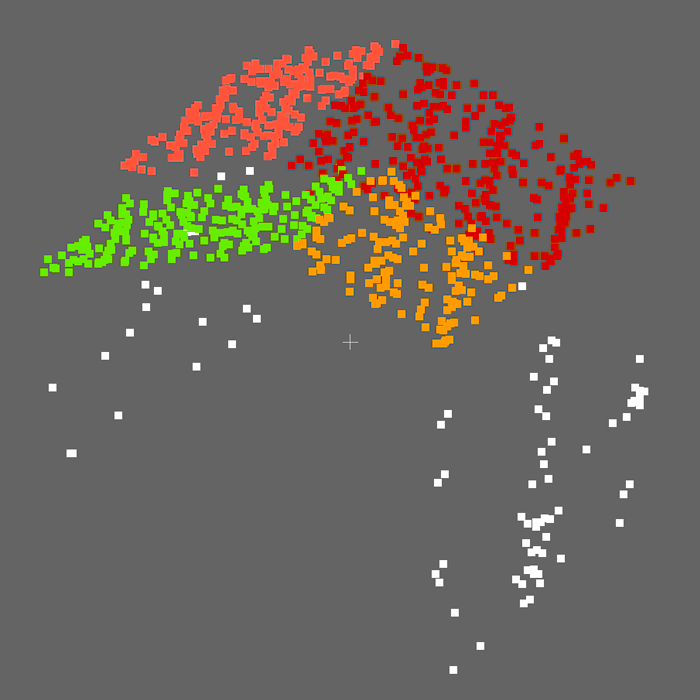} &
        \includegraphics[width=0.32\linewidth]{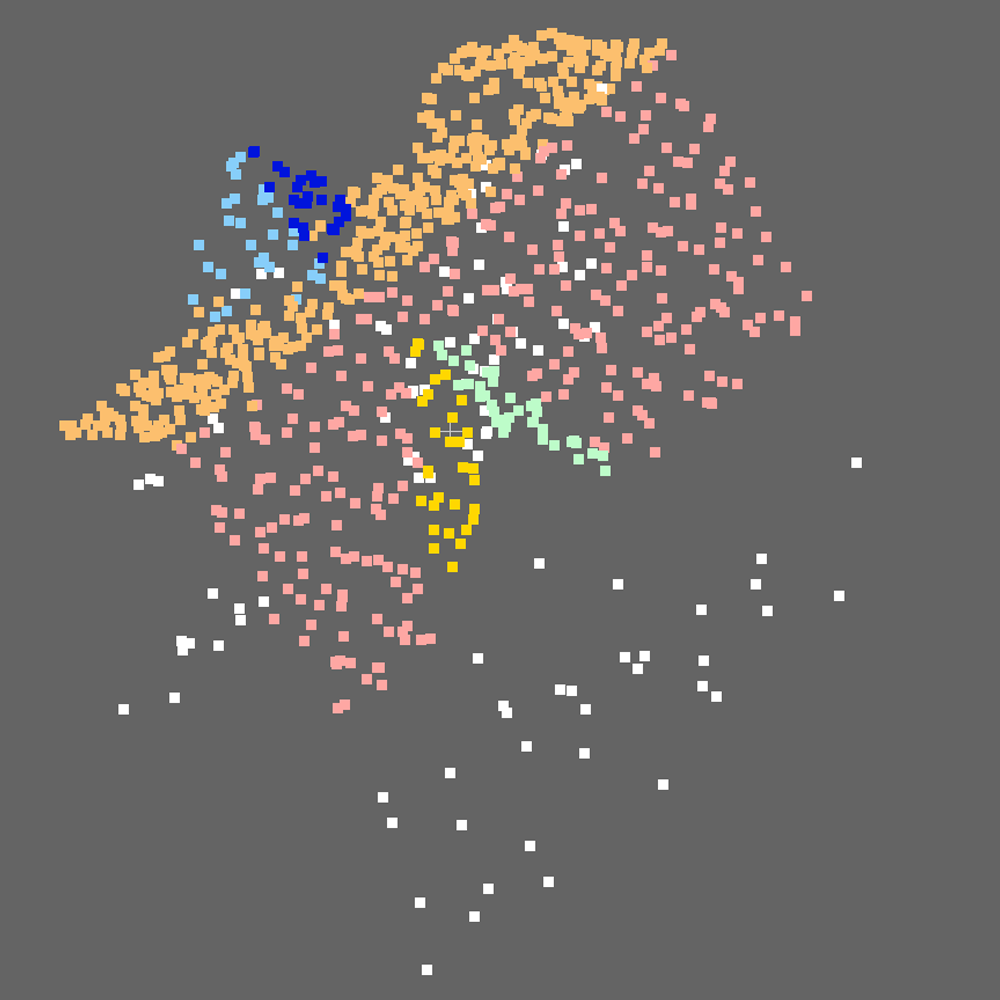} &
        \includegraphics[width=0.32\linewidth]{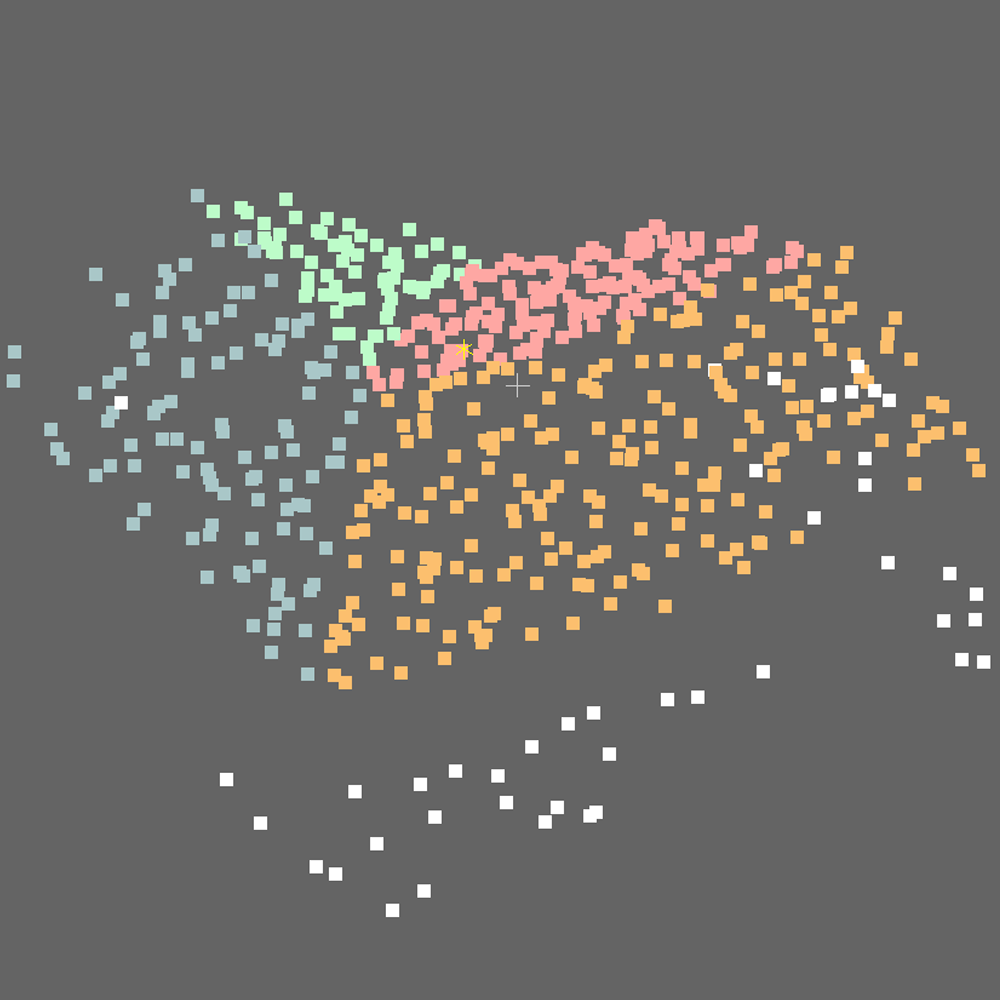} \\
        (d) & (e) & (f)
    \end{tabular}
    \caption{Reannotated examples from the RoofN3D test set. 
	(a), (b), and (c) show the original labels containing annotation errors, while (d), (e), and (f) display the corresponding corrected labels after reannotation for (a), (b), and (c), respectively. }

    \label{Fig:7}
\end{figure}

Because the RoofN3D dataset was automatically annotated using a traditional algorithm, some of the ground-truth annotations contain errors. 
To address this, we manually reannotated 49 samples in the test set that exhibit severe annotation errors. 
Examples of the corrected samples are shown in Fig.~\ref{Fig:7}. 
With this reannotated test set, we conducted a more rigorous evaluation. 
Table~\ref{Tab:3} presents the quantitative evaluation results, which once again demonstrate the significant advantages of our SPPSFormer. 
Notably, our SPPSFormer nano also performs well on the revised RoofN3D test set.

\begin{table} [!htb]
\renewcommand{\tabcolsep}{4.0 pt}
\scriptsize
\renewcommand{\arraystretch}{1.5}
    \begin{center}
        \caption{Quantitative evaluation results on the reannotated RoofN3D test set.}
        \label{Tab:3}
        \begin{tabular}{c|cccccc}
            \hline
            \multirow{2}{*}{Different approaches} & \multicolumn{6}{c}{Reannotated RoofN3D test set} \\
            \cline{2-7}
            & Cov & WCov & Precision & Recall & F1 score & Accuracy \\
            \hline
            SPFormer & 0.6576 & 0.7075 & 0.7809 & 0.5013 & 0.6108 & 0.4985 \\
            QTPS & 0.6337 & 0.6613 & 0.6994 & 0.6192 & 0.6571 & 0.6871 \\
            HCBR & 0.6767 & 0.7139 & 0.7027 & 0.9628 & 0.7912 & 0.7922 \\
            DeepRoofPlane & 0.9014 & 0.9211 & 0.8554 & \textbf{0.9969} & 0.9175 & 0.9516 \\
            Our SPPSFormer & \textbf{0.9271} & \textbf{0.9385} & \textbf{0.9078} & 0.9571 & \textbf{0.9304} & \textbf{0.9571} \\
            SPPSFormer nano & 0.9055 & 0.9120 & 0.8690 & 0.9800 & 0.9212 & 0.9392 \\
            \hline
        \end{tabular}
    \end{center}
\end{table}

\subsection{Ablation experiments} 
\label{sec:Ablation_study}
In this section, we conduct ablation studies to evaluate the impact of five improvements proposed in this paper. 

\textbf{Ablation experiments on our three modifications to the network.}
We first performed a detailed quantitative assessment of the performance improvements resulting from three key modifications to the network architecture, using the RoofN3D and Building3D datasets. 
The ablation results are shown in Table~\ref{Tab:4}, where not selecting SP indicates the use of the superpoints employed by SPFormer \cite{Sun_AAAI_2023} and not selecting KAN indicates that the MLP in the decoder is not replaced by KAN. 
The comparative results highlight that introducing our superpoint generation method alone leads to substantial improvements. 
Specifically, on the RoofN3D dataset, Cov increases by 15.45\% and WCov by 15.33\%. 
On the Building3D dataset, Cov improves by 8.75\% and WCov by 8.42\%. 
These results validate the significance of our superpoint generation criteria in deep learning tasks and confirm the effectiveness of our proposed superpoint generation approach. 
Additionally, ablation experiments on both datasets show that incorporating our handcrafted features enhances the model’s performance, indicating that our handcrafted features play a crucial role, especially when the training set is limited in size. 
Additionally, replacing the traditional MLP layer with the FourierKAN layer in the decoder yields further performance improvements under most testing conditions, confirming the positive role of KAN in model optimization.

\begin{table*} [!htbp] 
	\renewcommand{\tabcolsep}{12 pt}
	\scriptsize
	\renewcommand{\arraystretch}{1.5}
	\newcommand{\tabincell}[2]{\begin{tabular}{@{}#1@{}}#2\end{tabular}}
	\begin{center}
		\caption{Quantitative ablation results evaluating the impact of different components on network performance.}
		\label{Tab:4}
		\begin{tabular}{cccc|cccc|cccc} 
			\hline
			\multicolumn{4}{c}{Different components} & \multicolumn{4}{c}{RoofN3D dataset} &  \multicolumn{4}{c}{Building3D dataset}   \\ 
			\cline{1-4} \cline{5-8} \cline{9-12} 
			& HF    & KAN   & SP
	     	& Cov     & WCov  & F1 score   & Accuracy
	     	& Cov      & WCov  & F1 score   & Accuracy        \\
			\hline 
            \tabincell{c}{(a)} 
               &  &   &   & 
               0.6525 & 0.7023 & 0.6044 & 0.4966 &
               0.7895 & 0.8359 & 0.9270 & 0.8938 \\ 
            			\tabincell{c}{(b)} 
               & \checkmark &   &   & 
               0.6813 & 0.7145 & 0.6036 & 0.5192 &
               0.7891 & 0.8462 & 0.9315 & 0.8838 \\ 
                            \tabincell{c}{(c)}  
               & \checkmark & \checkmark &  & 
               0.6982 & 0.7256 & 0.6429 & 0.5424 &
               0.8205 & 0.8756 & 0.9484 & 0.9137 \\ 
            			\tabincell{c}{(d)} 
               &   &   &  \checkmark  & 
               0.8034 & 0.8540 & 0.8856 & 0.9391 &
               0.8770 & 0.9201 & 0.9573 & 0.9391 \\ 
                            \tabincell{c}{(e)} 
               &   & \checkmark  &  \checkmark  & 
               0.8070 & 0.8556 & 0.8847 & 0.9386 &
               0.8771 & 0.9238 & 0.9562 & 0.9393 \\ 
                            \tabincell{c}{(f)} 
               & \checkmark  &   &  \checkmark  & 
               0.8199 & 0.8541 & \textbf{0.8893} & 0.9264 &
               0.8819 & 0.9268 & 0.9576 & 0.9417 \\ 
            			\tabincell{c}{(g)} 
               & \checkmark & \checkmark & \checkmark & 
               \textbf{0.8594} & \textbf{0.8856} & 0.8816 & \textbf{0.9435} &
               \textbf{0.8862} & \textbf{0.9280} & \textbf{0.9593} & \textbf{0.9421} \\ 
			\hline 
		\end{tabular}
	\end{center}
	\begin{center}
		\text{\#HF denotes handcrafted features, SP denotes our superpoints.}
	\end{center}
\end{table*}

When all three improvements are simultaneously introduced to the network model, as shown in Table~\ref{Tab:4}, we observe significant performance gains. 
On the RoofN3D dataset, Cov increases by 20.69\% and WCov by 18.33\%; on the Building3D dataset, Cov improves by 9.67\% and WCov by 9.21\%. 
These experimental results collectively demonstrate the effectiveness and importance of our proposed strategies in enhancing model performance.

\begin{figure}[!htb]
    \centering
    \begin{tabular}{@{\hspace{0em}}c@{\hspace{0.4em}}c@{\hspace{0.4em}}c@{\hspace{0em}}}
        \includegraphics[width=0.32\linewidth]{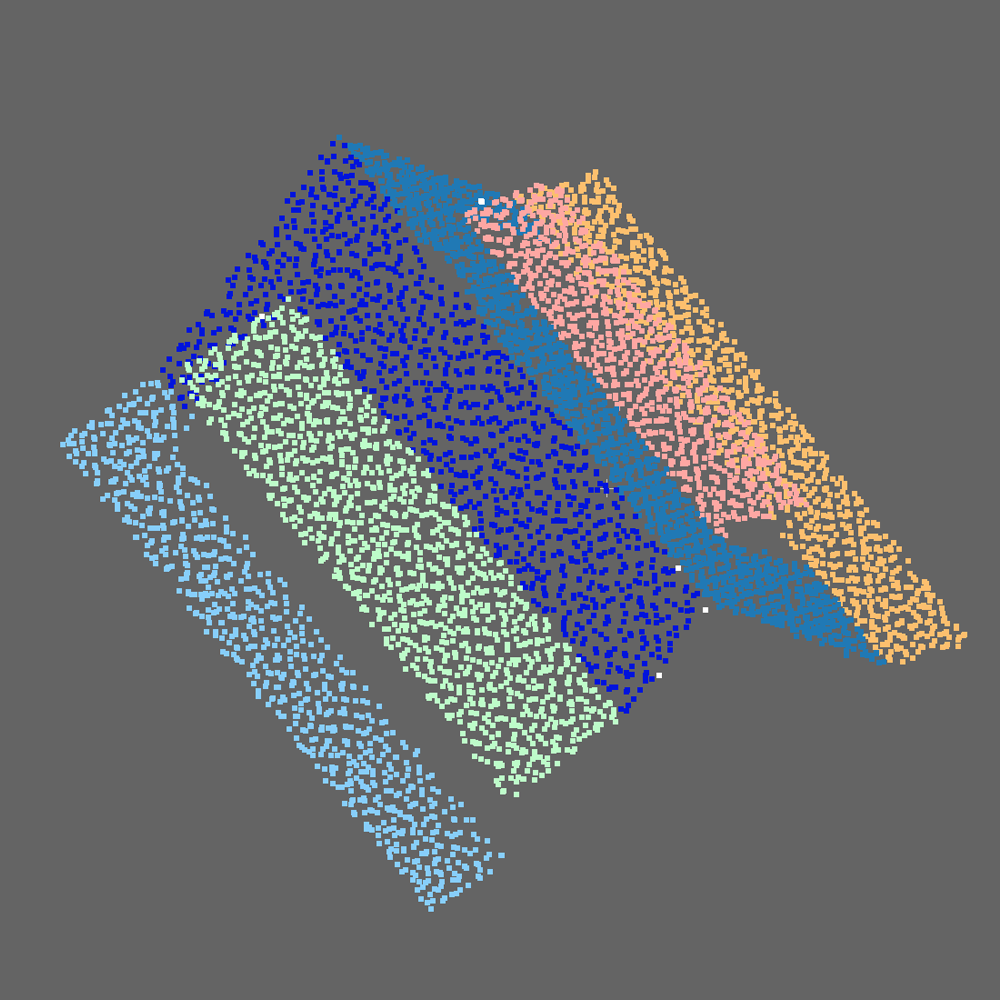} &
        \includegraphics[width=0.32\linewidth]{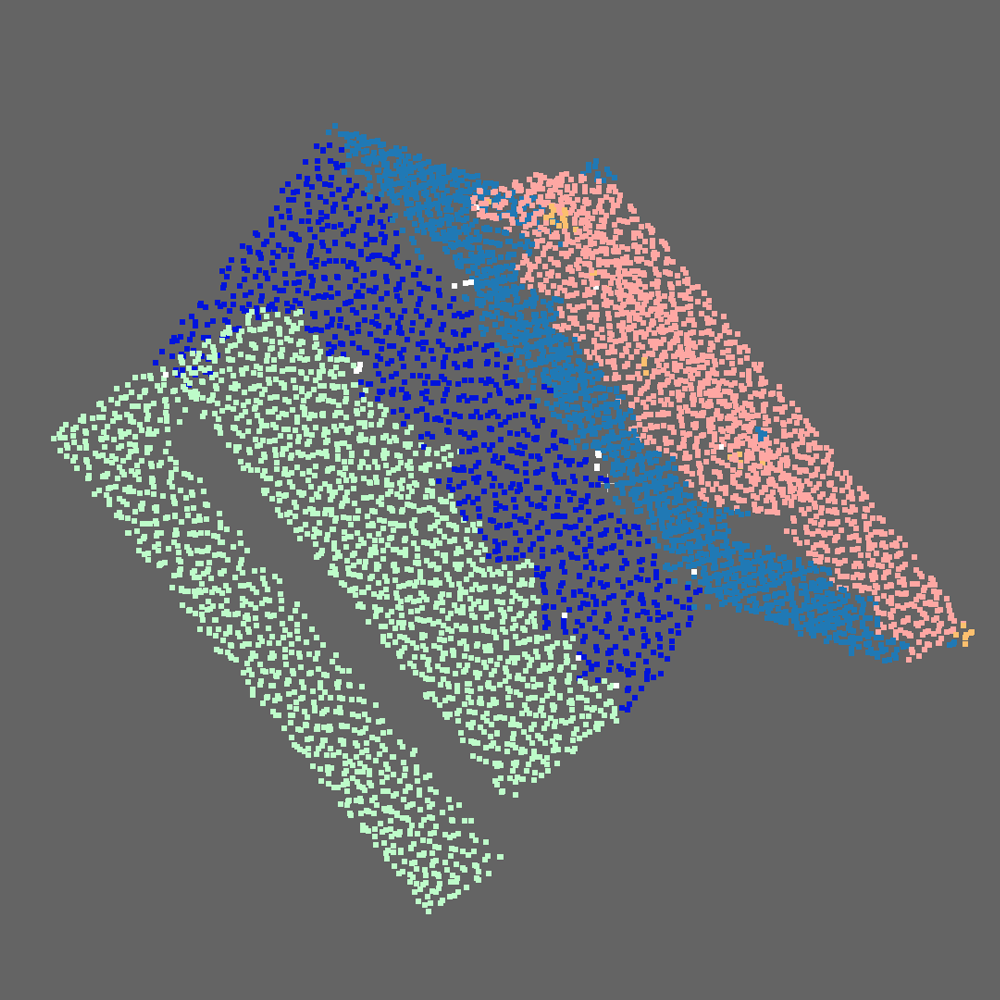} &
        \includegraphics[width=0.32\linewidth]{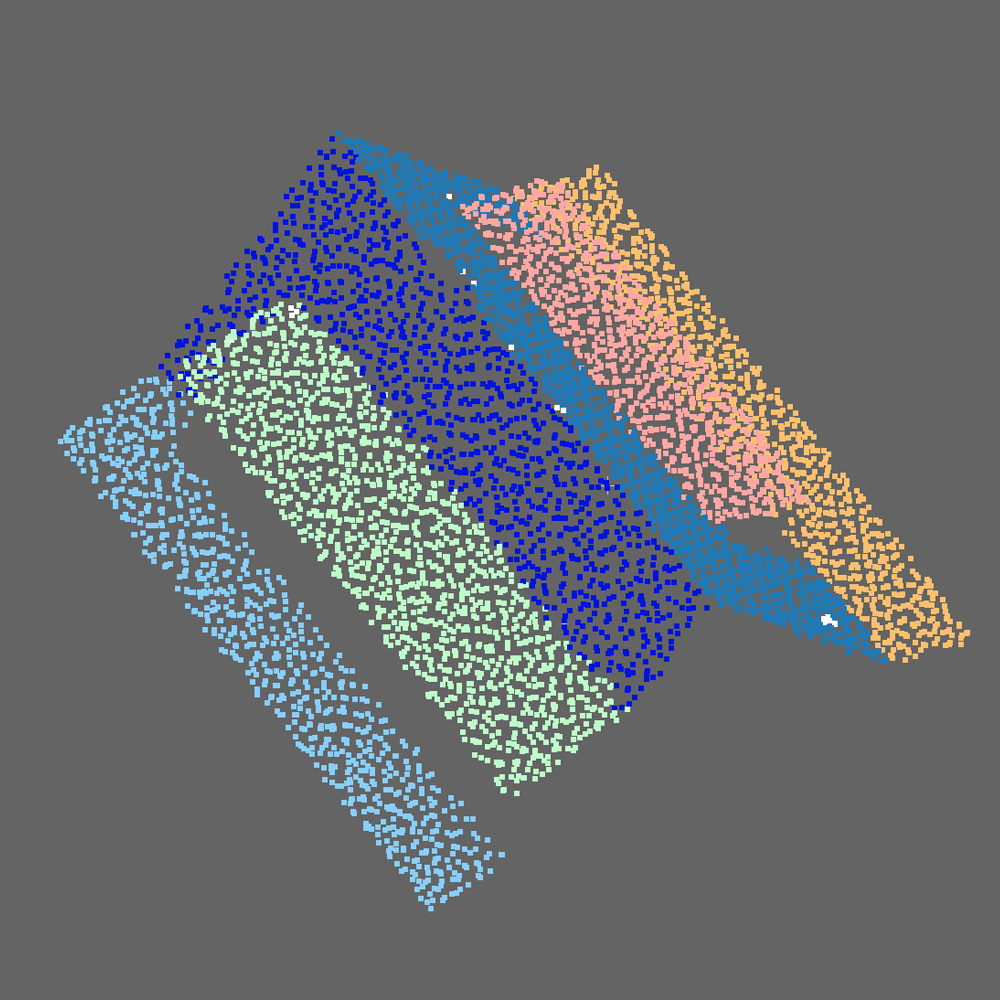} \\
        \scriptsize (a) Ground-truth &
        \scriptsize (b) Using basic superpoints &
        \scriptsize (c) Using our superpoints
    \end{tabular}
    \caption{Qualitative ablation results demonstrating the impact of our superpoint generation method on the Building3D dataset.}
    \label{Fig:8}
\end{figure}

Fig.~\ref{Fig:8} presents a qualitative comparison between the network predictions of our SPPSFormer and the SPPSFormer model using superpoints from SPFormer \cite{Sun_AAAI_2023} (referred to as basic superpoints) on the Building3D dataset. 
As shown in Fig.~\ref{Fig:8}, plane instance segmentation using basic superpoints results in inaccurate boundaries and severe under-segmentation. 
In contrast, using our proposed superpoints leads to significant improvements in segmentation performance. 
This further validates the crucial role of our superpoint generation method in enhancing the network’s ability to perform accurate plane instance segmentation.

\textbf{Ablation experiments on our two postprocessing modules.}
Table~\ref{Tab:5} presents the quantitative ablation results for the two proposed postprocessing modules on both datasets. 
From this table, we can see that on the RoofN3D dataset, applying only the plane completion module results in an increase of Cov by 1.98\% and WCov by 0.50\%. However, no significant changes in accuracy are observed on the Building3D dataset. 
This is because the Building3D dataset does not suffer from incomplete plane annotations, so networks trained on it are not impacted by such training set noise.
In contrast, the RoofN3D dataset contains this noise, which affects model training; however, our proposed plane completion algorithm effectively mitigates this issue and improves performance.

\begin{table} [!htb]  
	\renewcommand{\tabcolsep}{1.4 pt} 
	\scriptsize 
	\renewcommand{\arraystretch}{1.5} 
	\begin{center} 
		\caption{Quantitative ablation results evaluating the impact of our two postprocessing methods.} 
		\label{Tab:5} 
		\begin{tabular}{cc|*{4}{>{\centering\arraybackslash}p{0.9cm}}|*{4}{>{\centering\arraybackslash}p{0.9cm}}} 
 			\hline 
 			\multicolumn{2}{c|}{Setting} & \multicolumn{4}{c|}{RoofN3D dataset} &  \multicolumn{4}{c}{Building3D dataset}   \\ 
 			\cline{1-2} \cline{3-6} \cline{7-10} 
 			 PC    & BR 	     	& Cov & WCov & F1 score & Accuracy 	     	& Cov & WCov & F1 score & Accuracy \\ 
			\hline              
            &    &              0.8594 & 0.8856 & 0.8816 & 0.9435 &             0.8862 & 0.9280 & 0.9593 & 0.9421 \\              
            \checkmark &   &               0.8792 & 0.8906 & 0.9071 & 0.9330 &             0.8953 & 0.9297 & 0.9592 & 0.9384 \\              
            & \checkmark &               0.8428 & 0.8926 & 0.8964 & \textbf{0.9592} &             0.8937 & 0.9332 & 0.9615 & \textbf{0.9431} \\              
            \checkmark & \checkmark  &              \textbf{0.9321} & \textbf{0.9398} & \textbf{0.9238} & 0.9564 &             \textbf{0.9010} & \textbf{0.9347} & \textbf{0.9623} & 0.9417 \\  			
            \hline  		
        \end{tabular} 
	\end{center}     
    \begin{center} 
        \text{\#PC denotes plane completion, and BR denotes boundary refinement.} 
    \end{center} 
\end{table}

Building upon the plane completion module, the addition of the boundary refinement module further enhances segmentation performance on both datasets. 
The improvement is modest on the Building3D dataset but substantial on the RoofN3D dataset. 
This discrepancy arises because the Building3D dataset features relatively high point precision, density, and uniformity, allowing our network to already perform well, with limited room for enhancement. 
Conversely, the RoofN3D dataset suffers from low point cloud precision and density, and significant density variations, resulting in poorer initial performance and greater potential for improvement, and our boundary refinement module can effectively address the segmentation of 3D points at boundaries. 
When both plane completion and boundary refinement modules are applied, Cov and WCov increase by 7.29\% and 5.42\% on the RoofN3D dataset, respectively. 
On the Building3D dataset, the increases are more modest at 1.49\% and 0.67\%, respectively. 
These results demonstrate that our postprocessing algorithms adapt well to varying dataset characteristics and significantly refine network predictions.

\subsection{Discussion} 
\label{sec:Discussions}

Although the roof structures in the RoofN3D dataset are much simpler than those in the Building3D dataset, none of the algorithms tested in Section~\ref{sec:Comparative_experiments} achieved significantly better quantitative evaluation metrics on RoofN3D compared to Building3D. 
In particular, SPFormer \cite{Sun_AAAI_2023}, HCBR \cite{Li_RS_2020}, and QTPS \cite{Zhu_STAEORS_2021} show notably lower performance on RoofN3D. 
Regarding the performance of our method, as shown in Table~\ref{Tab:4}, without the proposed plane completion and boundary refinement modules, our network also performs significantly worse on RoofN3D than on Building3D. 
Even the results from DeepRoofPlane~\cite{Li_ISPRS_2024} suggest that RoofN3D is difficult to handle---despite the simpler roof shapes, its performance on RoofN3D is only comparable to that on Building3D, even with a slight decline in terms of WCov.

There are two primary challenges in the RoofN3D dataset that hinder high-quality plane instance segmentation: (1) the data quality itself is poor (low density, significant density variation, and low point precision), and (2) the annotation quality in the training set is suboptimal. 
The poor performance of traditional algorithms on the RoofN3D dataset is clearly attributed to the first factor.

To investigate the cause of the suboptimal performance of deep learning algorithms on the RoofN3D dataset, we further corrected the annotation errors for 408 samples in the RoofN3D training and validation sets (having already corrected the test set annotations, as previously described). 
We then retrained our SPPSFormer and DeepRoofPlane~\cite{Li_ISPRS_2024} using the corrected RoofN3D training set. 
As shown in the quantitative results in Tables~\ref{Tab:3} and~\ref{Tab:6}, the segmentation performance of both our SPPSFormer and DeepRoofPlane significantly improves after the training set annotations are corrected. 
This confirms that the suboptimal performance of these models in Table~\ref{Tab:3} can, at least in part, be attributed to the poor annotation quality of the RoofN3D training set. Therefore, for deep learning–based methods to perform optimally, it is crucial that the dataset annotations are sufficiently accurate. 
Our subsequent experiments further reveal that our SPPSFormer primarily relies on accurate annotations of the plane main bodies, while DeepRoofPlane depends heavily on both plane main body and boundary annotations.

\begin{table} [!htb] 
	\renewcommand{\tabcolsep}{4.0 pt}
	\scriptsize
	\renewcommand{\arraystretch}{1.5}
	\newcommand{\tabincell}[2]{\begin{tabular}{@{}#1@{}}#2\end{tabular}}
	\begin{center}
		\caption{Quantitative evaluation results of models retrained on the re-annotated RoofN3D training set.}
		\label{Tab:6}
		\begin{tabular}{c|cccccccc} 
			\hline
			\multirow{2}*{Different approaches} & \multicolumn{6}{c}{Reannotated RoofN3D test set}  \\  
			\cline{2-7}  
			& Cov     & WCov    & Precision & Recall   & F1 score & Accuracy  \\
			\hline 
			DeepRoofPlane    & 0.9237  & 0.9355  & 0.8775    & \textbf{0.9914}   & 0.9298   & 0.9606    \\
			Our SPPSFormer   & \textbf{0.9523} & \textbf{0.9593} & \textbf{0.9079} &0.9688 & \textbf{0.9360} & \textbf{0.9688} \\
			\hline 
		\end{tabular}
	\end{center}
\end{table}

Regarding the impact of data quality in the RoofN3D dataset on the plane segmentation performance of our SPPSFormer and DeepRoofPlane, it is challenging to evaluate experimentally. 
This is because we currently lack effective methods to improve the data quality of the RoofN3D dataset for a controlled comparison. 
Therefore, we conducted degradation experiments on the higher-quality Building3D dataset to investigate how point density, density variation, and geometric precision of points affect plane instance segmentation performance. 
We began by reducing the density of the Building3D dataset (including the test set) and retrained and tested both our SPPSFormer and DeepRoofPlane. 
The downsampling process involved randomly removing 50\% of the 3D points. The new quantitative results are presented in Table~\ref{Tab:7}. 
By comparing the results in Tables~\ref{Tab:2} and~\ref{Tab:7}, we observe that after reducing the density of the Building3D dataset, the plane instance segmentation performance of both our SPPSFormer and DeepRoofPlane decreases significantly.

\begin{table} [!htb] 
	\renewcommand{\tabcolsep}{4.0 pt}
	\scriptsize
	\renewcommand{\arraystretch}{1.5}
	\newcommand{\tabincell}[2]{\begin{tabular}{@{}#1@{}}#2\end{tabular}}
	\begin{center}
		\caption{Quantitative evaluation results of models retrained on the downsampled Building3D training set.}
		\label{Tab:7}
		\begin{tabular}{c|cccccccc} 
			\hline
			\multirow{2}*{Different approaches} & \multicolumn{6}{c}{Downsampled Building3D test set}  \\ 
			\cline{2-7}  
			& Cov & WCov & Precision & Recall & F1 score & Accuracy \\
			\hline 
			DeepRoofPlane & 0.8144 & 0.8572 & 0.9076 & 0.8797 & 0.8864 & 0.8475 \\
			Our SPPSFormer & \textbf{0.8418} & \textbf{0.9222} & \textbf{0.9884} & \textbf{0.9431} & \textbf{0.9632} & \textbf{0.9431} \\
			\hline 
		\end{tabular}
	\end{center}
\end{table}

Similarly, the corresponding quantitative evaluation results after introducing density variation and reducing the point precision of the Building3D dataset are presented in Tables~\ref{Tab:8} and~\ref{Tab:9}, respectively.
To increase density variation, a series of evenly spaced center planes parallel to the YZ plane are generated based on the spatial extent of the point cloud along the X-axis. 
Then, all 3D points are randomly shifted toward their nearest center plane by a small distance to simulate the uneven density characteristics observed in the RoofN3D dataset. 
To preserve the original point precision, each plane point is moved only along the direction parallel to its associated real plane during the shifting process.
For the reduction in point precision, we added a random offset ranging from 0 m to 0.5 m to the XYZ coordinates of each point.
By comparing the quantitative results in Table~\ref{Tab:2} with those in Tables~\ref{Tab:8} and~\ref{Tab:9}, we observe that when point cloud density variation increases or point precision degrades, the plane instance segmentation performance of both our SPPSFormer and DeepRoofPlane drops significantly.

\begin{table} [!htb] 
	\renewcommand{\tabcolsep}{4.0 pt}
	\scriptsize
	\renewcommand{\arraystretch}{1.5}
	\newcommand{\tabincell}[2]{\begin{tabular}{@{}#1@{}}#2\end{tabular}}
	\begin{center}
		\caption{Quantitative evaluation results of models retrained on the Building3D dataset with increased point density variation.}
		\label{Tab:8}
		\begin{tabular}{c|cccccccc} 
			\hline
			\multirow{2}*{Different approaches} & \multicolumn{6}{c}{Building3D test set with increased point density variation}  \\  
			\cline{2-7}  
			& Cov & WCov & Precision & Recall & F1 score & Accuracy \\
			\hline 
			DeepRoofPlane & 0.7714 & 0.8042 & 0.8871 & 0.8357 & 0.8536 & 0.8036 \\
			Our SPPSFormer & \textbf{0.8453} & \textbf{0.9168} & \textbf{0.9931} & \textbf{0.9339} & \textbf{0.9601} & \textbf{0.9339} \\
			\hline 
		\end{tabular}
	\end{center}
\end{table}

\begin{table} [!htb] 
	\renewcommand{\tabcolsep}{4.0 pt}
	\scriptsize
	\renewcommand{\arraystretch}{1.5}
	\newcommand{\tabincell}[2]{\begin{tabular}{@{}#1@{}}#2\end{tabular}}
	\begin{center}
		\caption{Quantitative evaluation results of models retrained on the Building3D dataset with reduced 3D point precision.}
		\label{Tab:9}
		\begin{tabular}{c|cccccccc} 
			\hline
			\multirow{2}*{Different approaches} & \multicolumn{6}{c}{Building3D test set with reduced 3D point precision}  \\  
			\cline{2-7}  
			& Cov & WCov & Precision & Recall & F1 score & Accuracy \\
			\hline 
			DeepRoofPlane & 0.3638   & 0.3836   & 0.6650    & 0.4520    & 0.5193    & 0.4131    \\
            Our SPPSFormer & \textbf{0.7843} & \textbf{0.8674} & \textbf{0.9962} & \textbf{0.9012} & \textbf{0.9425} & \textbf{0.9012} \\
			\hline 
		\end{tabular}
	\end{center}
\end{table}

Therefore, we conclude that for roof plane instance segmentation algorithms, in addition to the complexity of the roof structure, the quality of the 3D point cloud has a significant impact on segmentation performance. 
This suggests that future research on point cloud roof plane segmentation should focus more on the negative effects of low-quality point cloud data. 
For instance, during network training, data augmentation strategies that target point cloud quality---such as variations in density, density uniformity, and point precision---are essential to ensure robustness in handling complex and unseen real-world data. 
Notably, in all the quantitative results presented in Tables~\ref{Tab:7}, \ref{Tab:8}, and~\ref{Tab:9}, our SPPSFormer consistently outperforms DeepRoofPlane, with a particularly pronounced advantage observed in Table~\ref{Tab:9}. 
This demonstrates that our method is more effective at handling lower-quality point cloud data.

Many current deep learning researchers prioritize end-to-end models, whereas our method incorporates traditional algorithms for postprocessing. 
At first glance, this might seem like a limitation of our approach. 
However, it is precisely this architecture---deep learning–based prediction followed by traditional postprocessing---that reduces our model’s reliance on highly accurate plane boundary annotations in the training set. 
To validate this, we intentionally degraded the plane boundary annotations in the reannotated RoofN3D and Building3D training sets while keeping the plane main body annotations intact. 
We then retrained both our SPPSFormer and DeepRoofPlane models using these datasets with degraded boundary annotations. 
Specifically, we perturbed the boundary annotations by randomly swapping labels of points belonging to different planes if the distance between them is less than 0.5 m. 
Table~\ref{Tab:10} presents the quantitative results of the retrained model evaluated on the reannotated RoofN3D test set (the same as in Table~\ref{Tab:3}), while Table~\ref{Tab:11} shows the results evaluated on the original Building3D test set.

By comparing Table~\ref{Tab:2} with Table~\ref{Tab:11} and Table~\ref{Tab:3} with Table~\ref{Tab:10}, we observe that when plane boundary annotations in the RoofN3D and Building3D training sets are degraded, the segmentation performance of DeepRoofPlane drops significantly. 
In contrast, the performance of our SPPSFormer (including the postprocessing steps) declines only slightly. 
This is because our method requires only accurate segmentation of plane main bodies in the network prediction stage, which does not heavily rely on accurate boundary annotations in the training set. 
Furthermore, the traditional boundary refinement module in our method can effectively refine the segmentation of boundary points as long as the plane main bodies are correctly segmented. 
Although DeepRoofPlane also includes a boundary refinement module, it depends on deep learning at every stage of its processing pipeline. 
As a result, its performance is more sensitive to the quality of training annotations. 
When boundary annotations are inaccurate, DeepRoofPlane’s generalization ability degrades accordingly.

\begin{table} [!htb] 
	\renewcommand{\tabcolsep}{4.0 pt}
	\scriptsize
	\renewcommand{\arraystretch}{1.5}
	\newcommand{\tabincell}[2]{\begin{tabular}{@{}#1@{}}#2\end{tabular}}
	\begin{center}
		\caption{Quantitative evaluation results of models retrained on the boundary-corrupted RoofN3D training set.}
		\label{Tab:10}
		\begin{tabular}{c|cccccccc} 
			\hline
			\multirow{2}*{Different approaches} & \multicolumn{6}{c}{Reannotated RoofN3D test set}  \\  
			\cline{2-7}  
			& Cov     & WCov    & Precision & Recall   & F1 score & Accuracy  \\
			\hline 
			DeepRoofPlane    & 0.6962  & 0.7117  & 0.8212    & 0.7587   & 0.7784   & 0.7288    \\
			Our SPPSFormer   & \textbf{0.9159} & \textbf{0.9451} & \textbf{0.8928} & \textbf{0.9946} & \textbf{0.9381} & \textbf{0.9594} \\
			\hline 
		\end{tabular}
	\end{center}
\end{table}

\begin{table} [!htb] 
	\renewcommand{\tabcolsep}{4 pt}
	\scriptsize
	\renewcommand{\arraystretch}{1.5}
	\newcommand{\tabincell}[2]{\begin{tabular}{@{}#1@{}}#2\end{tabular}}
	\begin{center}
		\caption{Quantitative evaluation results of models retrained on the boundary-corrupted Building3D training set.}
		\label{Tab:11}
		\begin{tabular}{c|cccccccc} 
			\hline
			\multirow{2}*{Different approaches} & \multicolumn{6}{c}{Original Building3D test set}  \\  
			\cline{2-7}  
			& Cov     & WCov    & Precision & Recall   & F1 score & Accuracy  \\
			\hline 
			DeepRoofPlane    & 0.7655  & 0.8030  & 0.8905    & 0.8558   & 0.8659   & 0.8121    \\
			Our SPPSFormer   & \textbf{0.8792} & \textbf{0.9320} & \textbf{0.9456} & \textbf{0.9914} & \textbf{0.9654} & \textbf{0.9451} \\
			\hline 
		\end{tabular}
	\end{center}
\end{table}

Thus, while our use of traditional algorithms in postprocessing may seem at odds with the common end-to-end approach favored by many researchers, it allows our method to function effectively without relying heavily on highly accurate plane boundary annotations. 
In fact, accurately labeling boundary points or points near plane intersections is often the most challenging part of annotating point cloud plane segmentation datasets. 
In contrast, labeling plane main bodies is much easier. 
As a result, the reduced dependence on accurate boundary annotations in our method significantly reduces the difficulty of creating training datasets for roof plane instance segmentation in 3D point clouds.

\section{Conclusion}
\label{sec:conclusion}
In this study, we propose a superpoint Transformer for 3D point cloud roof plane instance segmentation. 
By leveraging superpoints as processing units, we significantly reduce the number of tokens the Transformer needs to process. 
However, existing superpoint generation methods are not optimized for effective Transformer-based feature learning. 
To address this, we establish superpoint generation criteria specifically designed for Transformers and introduce a corresponding generation approach, which substantially enhances the Transformer’s feature learning capability.  
To further improve segmentation performance, we integrate handcrafted features with spatial position features at the input stage of our network and employ a decoder that combines KAN with a Transformer. 
Additionally, we design a self-supervised plane completion method that infers segmentation parameters from segmented plane instances, alongside an efficient boundary refinement method.

We annotated a real-world dataset based on wireframe models in the Building3D dataset and corrected annotation errors in the RoofN3D dataset to enable a quantitative evaluation of our method. 
Experimental results demonstrate that our approach achieves SOTA performance on both our annotated dataset and the original and reannotated versions of the RoofN3D dataset. 
Moreover, our model relies predominantly on the accuracy of plane main body annotations in the training set and is far less dependent on accurate plane boundary annotations. 
This reduces the annotation effort required for training datasets significantly. 
We also identify key factors influencing point cloud roof plane instance segmentation performance: beyond roof structure complexity, point cloud quality, including factors such as density, density variation, and point precision, greatly impacts segmentation quality. 
This insight highlights the importance of incorporating data augmentation strategies related to point cloud quality during training to enhance robustness, enabling the network to handle complex and unseen real-world data more effectively.

In addition to roof plane instance segmentation, other instance segmentation tasks hold significant importance. 
We believe the superpoint generation criteria proposed here are broadly applicable to various instance segmentation tasks. 
Therefore, extending our superpoint generation approach to other instance segmentation problems will be a key direction for future work.

\section*{Acknowledgements}
We would like to thank the Wuhan Supercomputing Center for providing part of the computational resources.
The original RoofN3D training data \cite{Wichmann_RS_2019} was provided by the chair Methods of Geoinformation Science of Technische Universität Berlin and is available at \href{https://roofn3d.gis.tu-berlin.de}{https://roofn3d.gis.tu-berlin.de}.

\bibliographystyle{IEEEtran}
\bibliography{refs}

\end{document}